\documentclass[11pt, onecolumn, journal,compsoc]{IEEEtran}

\usepackage{amsmath, amssymb,amsthm}
\usepackage{stmaryrd}
\usepackage{epsfig,calc, graphicx}
\usepackage{wasysym}
\usepackage[usenames]{color}
\usepackage{graphicx}
\usepackage{subcaption}
\usepackage{multirow}
\usepackage{booktabs}
\usepackage{hyperref}
\usepackage{todonotes}
\usepackage{array}
\usepackage{tabulary}
\newcolumntype{C}[1]{>{\centering\arraybackslash}p{#1}}  %%% to control column widths for the PSNR table

\title{Dictionary Learning from Incomplete Data}

\author{Valeriya Naumova and Karin Schnass
\thanks{V.~Naumova is with Simula Research Laboratory AS, Martin Linges 17, Fornebu, Norway, valeriya@simula.no.}
\thanks{K.~Schnass is with the Department of Mathematics, University of Innsbruck, Technikerstra\ss e 13, 6020 Innsbruck, Austria, karin.schnass@uibk.ac.at.}
}

\newcommand\ip[2]{\langle #1, #2\rangle}

\newcommand\sparsity{S}

\newcommand\ie{{i.e. }}

\newcommand\dico{\Phi}

\newcommand\atom{\phi}

\newcommand\pdico{\Psi}
\newcommand\ppdico{\bar\Psi}
\newcommand\patom{\psi}
\newcommand\patomn{\bar\psi}

\newcommand\lrcomp{\Gamma}
\newcommand\lratom{\gamma}
\newcommand\plratom{\bar \gamma}

\newcommand\lrc{v}
\newcommand\scale{s}

\newcommand\noise{r}

\newcommand\nsigma{\rho}

\newcommand\signop{\operatorname{sign}}

\newcommand{\R}{{\mathbb{R}}}

\newcommand{\E}{{\mathbb{E}}}
\newcommand{\I}{{\mathbb{I}}}

\theoremstyle{plain}
\newtheorem{theorem}{Theorem}[section]

\newtheorem{algorithm}[theorem]{Algorithm}

\theoremstyle{remark}

\begin{document}		

\maketitle

\begin{abstract}
%In many real-life applications, the amount of high-quality training data available for model's training is extremely limited. Motivated by a particular example in diabetes therapy management, this paper presents a new method for efficient dictionary learning from incomplete and limited data. 

This paper extends the recently proposed and theoretically justified iterative thresholding and $K$ residual means algorithm ITKrM to learning dicionaries from incomplete/masked training data (ITKrMM). It further adapts the algorithm to the presence of a low rank component in the data and provides a strategy for recovering this low rank component again from incomplete data. Several synthetic experiments show the advantages of incorporating information about the corruption into the algorithm. Finally, image inpainting is considered as application example, which demonstrates the superior performance of ITKrMM in terms of speed at similar or better reconstruction quality compared to its closest dictionary learning counterpart.
\end{abstract}


%\input{sec_intro}
%%%%%%%%%%%%%%
% !TEX root = maskdl.tex
%%%%%%%%%%%%%%

%%%%%%%%%%%
\section{Introduction}\label{sec:intro}
%%%%%%%%%%%

Many notable advances in modern signal processing are based on the fact that even high-dimensional data follows a low complexity model.  
One such model, which has become an important prior for many signal processing tasks ranging from denoising and compressed sensing, to super-resolution, inpainting and classification is sparsity in a dictionary, \cite{doelte06, bestfa13, carota06, do06cs, yawrhuma10, elstqudo05, wrma09, rubrel10}. In the sparse model each datum (signal) can be approximated by the linear combination of a small (sparse) number of elementary signals,
called atoms, from a pre-specified basis or frame, called dictionary.
In mathematical terms, if we represent each signal by a vector $y_n \in \mathbb R^d$ and collect the entire dataset in the matrix $Y = (y_1,\ldots, y_N) \in \mathbb \R^{d \times N},$ the sparse model can be formalised as 
%represent $Y$ as a linear combination of the dictionary  columns  
\begin{equation}
\label{main_model}
Y = \Phi X \text{ where } X \text{ is columnwise sparse}.
\end{equation}
Here the dictionary matrix $\Phi$ contains $K$ normalised vectors (atoms) $\phi_k,$ stored as columns in $\Phi = (\phi_1, \ldots, \phi_K) \in \mathbb \R^{d \times K},$ and 
each vector-column $x_n \in \mathbb \R^{K}$ of the matrix $X = (x_1,\ldots, x_N) \in \mathbb \R^{K \times N}$ contains only few nonzero entries.
Since the model expressed in Eq.~\eqref{main_model} has proven to be very useful in signal processing, the natural next question is, how to automatically learn a dictionary $\Phi$, providing sparse representations for a given data class. This problem is also known as dictionary learning, sparse coding or sparse component analysis. By now there exist not only a multitude of dictionary learning algorithms to choose from, \cite{ahelbr06, enaahu99, olsfield96, kreutz03, lese00, mabaposa10, sken10, mabapo12}, but also theoretical results have started to accumulate, \cite{grsc10, spwawr12, argemo13, aganjaneta13, sc14, sc14b, bagrje14, bakest14, suquwr15, argemamo15}. As our reference list is necessarily incomplete we also point to the surveys \cite{rubrel10, sc15imn} as trailheads for algorithms and theory respectively.\\
 One common assumption on which all algorithms and associated theories are based is that large numbers 
 of clean signals are available for learning the dictionary. However, this assumption is often not valid in actual applications. Motivated by a real-life prediction task in diabetes therapy management, in this paper we will consider the following problem: How do we learn a dictionary when there are only a few or no clean training signals available?\\
Diabetes is currently considered one of the global healthcare challenges of the century, with more than 380 million people affected worldwide.
The biggest challenge in diabetes management is the prediction of blood glucose levels from past and current measurements. The most recent advances in the field are based on the observation that purely data-driven algorithms lead to more clinically accurate results than the ones based on physiological models or a combination of both \cite{nape12}. Due to recent technological advances,  
blood glucose measurements can be provided on a close-to continuous basis, every 5 to 10 minutes, by a Continuous Glucose Monitoring (CGM) device, which is inserted under the skin. 
 \begin{figure}
  \begin{center}
  \includegraphics[width=.7\linewidth]{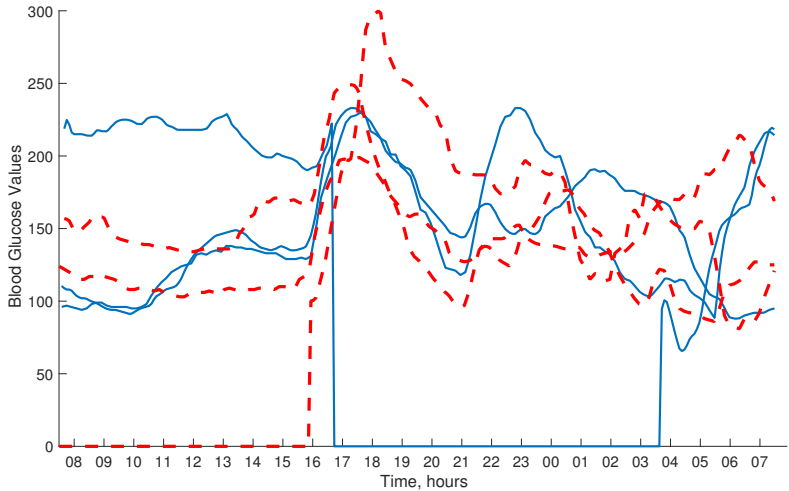}
  \end{center}
\caption{Examples of blood glucose profile of two patients (solid and dashed lines, respectively) during inpatient stay for three days. Each curve represents blood glucose profile for a 24 hour-period from 08:00 till 07:59 next day. Notice signal dropouts for several hours for at two out of six glucose traces.}
\label{fig:blood}
\end{figure}
However, in addition to mandatory unpleasant calibration procedures of the device several times a day, CGM quite often returns obviously wrong, e.g. rapidly oscillating or negative, estimations of the blood glucose level and suffers from frequent signal dropouts, Figure~\ref{fig:blood}, \cite{schpa16}. The latter are especially common during the night, and can be particularly dangerous since there is no warning for low glucose levels, which in extreme cases can  lead to coma or even death. One task is therefore prediction of glucose levels even with signal dropouts. This can be interpreted as inpainting (into the future) and a data-driven approach could be to learn a dictionary for the class of CGM signals and to use dictionary based inpainting. The obvious problem is that the CGM signals for learning are quite difficult to obtain and suffer from dropouts themselves. Therefore, any dictionary learning algorithm, or any other data-driven approach, needs to make use of all possible information and include the corrupted signals by properly modelling the corruption and accounting for it in the learning.\\
To solve the problem of learning from incomplete data, we propose an algorithm called {\it Iterative Thresholding and $K$ residual Means for Masked data} (ITKrMM). As the name suggests, it is built upon the inclusion of a signal corruption model into the theoretically-justified and numerically efficient {\it Iterative Thresholding and $K$ residual Means} (ITKrM) algorithm, \cite{sc15}.\\
In order to model the data corruption/loss process, we will adapt the concept of the binary erasure channel. In this model, the measurement device sends a value and the receiver either receives the value or it receives a message that the value was not received ('erased'). The model is used frequently in information theory due to its simplicity and its abstraction towards modelling various types of data losses. At the same time, this setting provides information on the location of the erasures and, thus, we can employ the concept of a mask $M$ to describe the corrupted data as $My$. Without loss of generality, we will think of a mask $M$ as orthogonal projection onto the linear span of vectors from the standard basis $(e_j)_j$ or simply as diagonal matrix with $M(j,j) \in \{0,1\}.$ 
We further extend the algorithms to account for the presence of low-rank component in the data, which appears in many real-life signals and, as we will illustrate below, should be treated cautiously in the considered context. \\
To evaluate the accuracy and efficiency of the algorithm, we perform various numerical tests on synthetic data. We also demonstrate the practical usefulness of the algorithms to inverse problems, by considering an image inpainting task.\\
The dictionary learning community does not directly address the problem under consideration. However, several recent works by Elad and co-authors \cite{maelsa08, masael08} introduced a weighted algorithm for dictionary learning, called weighted K-SVD (wK-SVD), for handling non-homogenous noise in signals. 
The proposed construction is also applicable in cases with missing values, such as in colour image demosaicing and inpainting. We will show that our algorithm not only performs on par with the weighted K-SVD algorithm %(or rather a version further adapted to the problem) 
but also requires much less computational resources.\\
{\bf Contribution:} This paper provides an efficient and simple algorithm for dictionary learning from incomplete data and the recovery of the low-rank component also from incomplete data. For the sake of brevity and different interests across communities, 
we here focus on the methodological description and extensive numerical justification and aim to provide a theoretical analysis in a follow-up paper.\\
{\bf Outline:} The paper is organised as follows: Section~\ref{sec:setup} contains the complete problem set-up, explaining the combined low-rank and sparse model and as well as the corruption model. %as well as summarises notational conventions. 
The ITKrMM algorithm for dictionary recovery is introduced in Section~\ref{sec:itkrm}. An adaptation of this algorithm for recovery of the low-rank component from incomplete data together with a short discussion of related works in the field of matrix completion and dimensionality reduction is provided in Section~\ref{sec:lr}. 
Section~\ref{sec:tests} contains extensive simulations on synthetic data, while Section~\ref{sec:applications} exemplifies the usefulness of the algorithm to problems in image processing by applying it to image inpainting. Finally Section~\ref{sec:discussion} offers a snapshot of the main contributions and points out open questions and directions for future work.\\
{\bf Notation:} Before finally lifting the anchor, we provide a short reminder of the standard notations used in this paper. For a matrix $A$, we denote its (conjugate) transpose by $A^\star$ and its Moore-Penrose pseudo inverse by $A^\dagger$. 
%Note that in the special case where the columns of $A$ are linearly independent we have $A^\dagger = (A^\star A)^{-1} A^\star$. 
By $P(A)$ we denote the orthogonal projection onto the column span of $A$, \ie $P(A)=AA^\dagger$ and by $Q(A)$ the orthogonal projection onto the orthogonal complement of the column span of $A$, that is $Q(A) = \I_d - P(A)$, where $\I_d$ is the identity operator (matrix) in $\R^d$.\\%We denote its operator norm by $\|A\|_{2,2}=\max_{\|x\|_2=1}\|Ax\|_2$ and its Frobenius norm by $\|A\|_F= \tr(A^\star A)^{1/2}$, remember that we have $\|A\|_{2,2}\leq \|A\|_F$.\\
%We consider a {\bf dictionary} $\dico$ a collection of $K$ unit norm vectors $\atom_k\in \R^d$, $\|\atom_k\|_2=1$. By abuse of notation we will also refer to the $d \times K$ matrix collecting the atoms as its columns as the dictionary, i.e. $\dico=(\atom_1, \ldots \atom_K)$. 
The restriction of the dictionary $\dico$ to the atoms indexed by $I$ is denoted by $\dico_I$, i.e. $\dico_I=(\atom_{i_1},\ldots, \atom_{i_\sparsity} )$, $i_j\in I$. The maximal absolute inner product between two different atoms is called the coherence $\mu$ of a dictionary, $\mu=\max_{k \neq j}|\ip{\atom_k}{\atom_j}|$ and encapsulates information about the local dictionary geometry.\\

%%%%%%%%%%%%%%
% !TEX root = maskdl.tex
%%%%%%%%%%%%%%

%%%%%%%%%%%
\section{Problem set-up}\label{sec:setup}
%%%%%%%%%%%

Our goal is to learn a dictionary $\dico$ from corrupted signals $M_n y_n$, under the assumption that the signals $y_n$ are sparse in the dictionary $\dico$. There are some notable differences in this problem setting compared to the uncorrupted situation. First, we cannot without loss of generality assume that the corrupted signals are normalised, since the action of the mask distorts the signal energy, $\|My\|_2 \leq \| y\|_2$, which makes simple renormalisation impossible.\\
Another issue in modelling a natural phenomenon is that the signals might not be perfectly sparse but can only be modelled as the orthogonal sum of a low-rank and a sparse component. An example for such signals are images, where one usually subtracts the foreground or in other words the signal mean before learning the dictionary, which consequently will consist of atoms with zero mean \cite{ahelbr06}. Without taking into account the existence of  the low-rank component one would likely end up with a very ill-conditioned and coherent dictionary, where most atoms are distorted towards the low-rank component.\\
Similarly, in our motivating example of the blood glucose data (see Figure \ref{fig:blood}), we can see at first glance that the signals vary around a baseline signal and that imposing a sparse structure in a dictionary makes sense only after subtracting this common component. As before, the atoms in this dictionary should then be orthogonal to the baseline signal.\\
In the case of uncorrupted signals one can simply determine the common low-rank component $\lrcomp=(\lratom_1\ldots \lratom_L)$ using one's preferred method such as a singular value decomposition and subtract its contribution from the signals via $\tilde{y}_n= Q(\lrcomp)y_n$. Then in a second separate step one can run the dictionary learning algorithm on the modified signals $\tilde{y}_n$ and the resulting atoms will automatically be orthogonal to the low-rank component $\Gamma$. However, in the case of corrupted signals the action of the masks destroys the structure. So, while the dictionary is orthogonal to the low-rank component, $\dico^\star \lrcomp=0$, this orthogonality is not preserved by the action of the mask, that is $\dico^\star M \lrcomp \neq 0$. As we will see later, the consequence of this effect is that we have to take the presence of the low-rank component into account when learning the sparsifying dictionary. Moreover, before even going to the dictionary learning phase, we have to find a strategy to recover the low-rank component from the corrupted signals.\\ 
The third difference is that we get additional constraints on the dictionaries in order for them to be recoverable.
In the case of uncorrupted signals the main criterion for a dictionary to be recoverable is that its coherence scales well with the average sparsity level $S$ of the signals ($S \mu^2 \lesssim 1$) and that all atoms are somewhat equally and independently used. In our scenario, where we want to learn a dictionary from corrupted data, we impose another criterion for the recoverability of the dictionary, which is the robustness of the dictionary to corruption. 
For instance, we will not have a chance to recover an atom $\atom_k$ if its presence in a signal always triggers the same corruption pattern $M_0$ which distorts the atom, $M_0\atom_k \neq \atom_k$. This means that we have to assume some sort of independence between the signals $y_n$ and the corruption, represented by the masks $M_n$. Similarly, it will be very hard to recover a dictionary, whose incoherence is not robust towards corruption. To avoid this complication, we assume that the dictionary and the low-rank component consist of 'flat' atoms, where $\| \atom_k\|_\infty \ll 1$ resp. $\|\lratom_\ell \|_\infty \ll1$. A more detailed discussion why this is a suitable assumption can be found in Section~\ref{sec:itkrm}. 
For the moment we just want to point out that this is in line with the potential application of the learned dictionaries to signal reconstruction tasks such as inpainting. There the information in the corrupted part of an image needs to be encoded by the rest of the image, which is the case if the image is sparsely represented by flat atoms.\\
Summarising these considerations, we arrive at the following signal model, which we will use as inspiration for the development of the algorithms, and
as basis for the planned theoretical analysis.%, \cite{nasc16b}.

\noindent{\bf Signal model:}\\
Given a $d\times L$ low-rank component $\lrcomp$ with $\lrcomp^\star \lrcomp = \I_L$ and a $d\times K$ dictionary $\dico$, where $\lrcomp^\star \dico = 0$ and $L\ll K$ the signals are generated as, 
\begin{align}\label{noisymodel1}
y= \scale \cdot \frac{ \lrcomp \lrc  + \dico x +\noise}{\sqrt{1+\|\noise \|_2^2}} \approx \scale ( \lrcomp \lrc  + \dico_I x_I ),
\end{align}
where $\|\lrc\|_2^2+ \|x\|^2_2 = 1$ and $|I|=S$. \\The scaling parameter $\scale$ is distributed between
$\scale_{\min}$ and $\scale_{\max}$ and accounts for signals with different energy levels.\\
 The low-rank component is assumed to be present in every (most signals) and irreducible, meaning the coefficients $\lrc$ are dense and $\E(\lrc\lrc^\star)$ is a diagonal matrix. Also the average contribution of a low-rank atom should be larger than that of a sparse atom, $\E(|\lrc(\ell)|)\gg \E (|x(k)|)$. \\
The sparse coefficients $x$ should be distributed in a way that for every single signal only $S$ entries in $x$ are effectively non-zero. All atoms $\atom_k$ should be irreducible and on average contribute equally to the signals $y_n$. Specifically, no two atoms should always be used together, since in this case they could be replaced by any other two atoms with the same span. For a more detailed discussion of admissible coefficient models we refer to \cite{sc15}. \\% and again the accompanying paper \cite{nasc16b}. \\
For our derivations we will keep in mind the following simple model: with constant scale and without noise. The low-rank component is one-dimensional, $L=1$, and the low-rank coefficient is equally Bernoulli distributed on $\pm c_\lrc$. The sparse coefficients are constructed by choosing a support $I$ of size $S$ uniformly at random and setting $x(k)=\pm c$, iid equally Bernoulli distributed, for $k\in I$ and $x(k) =0$ else. In other words the coefficients restricted to the support are a scaled Rademacher sequence. Following the above considerations concerning the scalings, we have $c_\lrc^2 + S\cdot c^2 =1$ and $c_\lrc \gg c S/K $.\\
Similar to the signal model we also summarise our considerations concerning the corruption in a model.\\

\noindent{\bf Corruption model:}\\
As mentioned above, the corruption of a signal $y$ is modelled by applying a mask $M$, where we assume that the distribution of the mask is independent of the signal distribution. By receiving a corrupted signal, we understand that we have access both to the corrupted signal $My$, and the location of the corruption in form of the mask $M$, meaning we receive the pair $(My,M)$. \\
For the development of the algorithms we will keep two types of corruption in mind. The first type are random erasures, where the $j$-th coordinate is received with probability $\eta_j$ independently of the reception of the other coordinates, meaning $M(j,j)\sim B(\eta_j)$ are independent Bernoulli variables.\\
The second type are burst errors or sensor malfunctions, as can be observed in the blood glucose example. We model them by choosing a burst-length $\tau$ and a burst-start $t$, according to a distribution $\nu_{\tau,t}$. Based on $\tau$ and $t$ we then set $M(j,j)=0$ for $t \leq j < t+\tau$ and $M(j,j)=1$ else. One simple realisation of such a distribution would be to have no burst, $\tau=0$, with probability $\theta$ and a burst of fixed size, $\tau=T$, which corresponds, for instance, to the time the sensor needs to be reset, with probability $1-\theta$. The burst-start could be uniformly distributed, if the sensor is equally likely to malfunction throughout the measurement period, or for instance with a higher weight on part of the coordinates, if the sensor is more likely to malfunction during part of the measurement period, for instance, during the night.\\
Having defined our problem set-up we are now ready to address the recovery of the dictionary from corrupted data.

%%%%%%%%%%%%%%
% !TEX root = maskdl.tex
%%%%%%%%%%%%%%

%%%%%%%%%%%%%%%
\section{Dictionary recovery} \label{sec:itkrm}
%%%%%%%%%%%%%

We will use the iterative thresholding and $K$ residual means algorithm (ITKrM), \cite{sc15}, as base for recovering the dictionary.
It belongs to the class of alternating projection algorithms, which alternate between sparsely approximating the signals in the current version of the dictionary and updating the dictionary based on the sparse approximations. As the name suggests, ITKrM uses thresholding as sparse approximation algorithm and residual averages for the dictionary update and as such has the advantage of being computationally light and sequential. Further, there are theoretical results concerning its local convergence and good experimental results concerning its global convergence. Together with our expectation that it will be much easier to incorporate the information about corruption into a dictionary update scheme that uses averages than into one that uses higher order statistics such as singular vectors, this makes ITKrM a promising starting point.
%%%%
\begin{algorithm}[ITKrM - one iteration] 
Given an input dictionary $\pdico$, a sparsity level $S$ and $N$ training signals $y_n$ do:
\begin{itemize}
\item For all $n$ find $I_n^t= \arg\max_{I: | I |=S} \| \pdico_I^\star y_n\|_1$.
\item For all $k$ calculate%\footnote{For a matrix M $P(M)$ denotes the orthogonal projection onto the rowspan of $M$}. 
\begin{align}
\patomn_k= \sum_{n: k\in I_n^t} \big[\I_d - P(\pdico_{I_n^t}) + P(\patom_k)\big] y_n \cdot \signop(\ip{\patom_k}{y_n}).
%\patomn_k= \sum_{n: k\in I_{n}^t} \big[\I - P&(\pdico_{I_{n}^t})  + P(\patom_k) \big] y_n \cdot \signop(\ip{\patom_k}{y_n}),\notag %\cdot  \chi(I_n^t, k),
\end{align}
\item Output $\ppdico=( \patomn_1/\| \patomn_1\|_2, \ldots, \patomn_K/\|\patomn_K\|_2)$.
\end{itemize}
\end{algorithm}
%%%%%%%
To see how we have to modify the algorithm to deal with corrupted data, it will be helpful to understand how ITKrM works. ITKrM can be understood as fixed point iteration, meaning the generating dictionary $\dico$ is a fixed point and locally, around the generating dictionary, one iteration of ITKrM is a contraction, $\|\atom_k - \frac{\patomn_k}{\| \patomn_k\|_2}\|_2 < \kappa \|\atom_k -\patom_k\|_2$ for all $k$ and some $\kappa<1$.
We refer to \cite{sc15} for details but for the sake of completeness we will provide some perhaps intuitive background for both the fixed point and the contraction property. \\
Assume for a moment that the signals follow the simplest sparse model, that is, they are perfectly $S$-sparse in a generating dictionary $\dico$, meaning $y_n = \dico_{I_n} x_n(I_n)$ for some $|I_n|=S$ and $x_n(i) \approx \pm c$ for $i\in I_n$, compare to the model presented in Section~\ref{sec:setup}.
In particular, they all have the same scaling and contain neither a low-rank component nor are they contaminated by noise. If we are given the generating dictionary as input dictionary, $\pdico=\dico$, then as long as the dictionary is not too coherent compared to the sparsity level, $\mu^2S\lesssim 1$, thresholding will recover the generating support, meaning $I_n^t = I_n$. Provided that the generating support was always recovered, we have $P(\pdico_{I_n^t})y_n= P(\dico_{I_n})y_n = y_n$ and before normalisation the updated atom takes the form
\begin{align}
\patomn_k= \sum_{n: k\in I_n}  P(\atom_k) y_n \cdot \signop(\ip{\atom_k}{y_n}) = \sum_{n: k\in I_n}  |\ip{\atom_k}{y_n}| \cdot \atom_k.
%\patomn_k= \sum_{n: k\in I_{n}^t} \big[\I - P&(\pdico_{I_{n}^t})  + P(\patom_k) \big] y_n \cdot \signop(\ip{\patom_k}{y_n}),\notag %\cdot  \chi(I_n^t, k),
\end{align}
This means that the output dictionary is again the generating dictionary $\ppdico =\dico$ or, in other words, that the generating dictionary is a fixed point of ITKrM. Note also that before normalisation the updated atom consists of roughly $N_k =\sharp \{n: k\in I_n\}$ scaled copies of itself because 
$|\ip{\atom_k}{y_n}| \approx |x_n(k)| \approx c$ and therefore 
\begin{align} \label{nkcopies}
\patomn_k \approx \sum_{n: k\in I_n}  c \atom_k = c N_k \atom_k.
\end{align}
\\
To provide insight why one iteration of ITKrM acts as contraction, assume again that we know all generating supports $I_n$ and that our current estimate for the dictionary consists of all generating atoms except for the first one, $\patom_k = \atom_k$ for $k\geq2$. For the first atom we only have some (poor) approximation, which is, however, still incoherent with all other atoms, $1 > |\ip{\patom_1}{\atom_1}| \gg |\ip{\patom_1}{\atom_k}| \approx d^{-1/2}$ for $k\geq2$, or, in other words, the current estimate $\patom_1$ contains more of the first than of any other generating atom. As before, one iteration of ITKrM will preserve all atoms $\patom_k = \atom_k$ for $k\geq2$ and on top of that contract $\patom_1$ towards $\atom_1$.
To see this, observe that as long as the current estimate contains more of the first than of any other generating atoms, $|\ip{\patom_1}{\atom_1}| \gg |\ip{\patom_1}{\atom_k}|$, whenever $1\in I$ for $y = \dico_{I} x(I)$ we have 
\begin{align}
P(\patom_1)y =  P(\patom_1) \dico_{I} x(I) \approx  x(1) P(\patom_1) \atom_1.
\end{align}
and, similarly,
\begin{align}
y - P(\pdico_{I}) y = x(1) \left[\atom_1 - P(\pdico_{I}) \atom_1\right] \approx x(1)\left[\atom_1 - P(\patom_1) \atom_1\right].
\end{align}
Combining the two estimates we get 
\begin{align}
\patomn_1&= \sum_{n: 1\in I_n} \big[\I_d - P(\pdico_{I_n}) + P(\patom_1)\big] y_n \cdot \signop(\ip{\patom_1}{y_n}) \approx \sum_{n: 1\in I_n}  x_n(1)\signop(\ip{\patom_k}{y_n}) \cdot \atom_1,
\end{align}
which shows that also a poor approximation $\patom_1$ is quickly contracted towards the generating atom $\atom_1$. \\
In summary, for our modifications we have to ensure to preserve both the fixed point and the contraction property. For the start, we again assume that the corrupted signals have equal scale, contain no low-rank component, and are not contaminated by noise, but are perfectly $S$-sparse, that is $M_n y_n =  M_n \dico_{I_n} x_n(I_n)$. First, observe that a corrupted signal $M_n y_n$ is not sparse in the generating dictionary $\dico$ but in its corrupted version $M_n\dico$,
\begin{align} 
M_n y_n =  M_n\dico_{I_n} x_n(I_n) =  \sum_{i\in I_n} x_i M_n \atom_i. 
\end{align}
Still, we can recover the support $I_n$ via thresholding using the corrupted dictionary $M_n \dico$ since we have access to the mask $M_n$. However, we have to take into account that, strictly speaking, the corrupted dictionary is not actually a dictionary in the sense that its columns are not normalised. Depending on the shape of the atoms, flat or spiky, and the amount of corruption, $\|M_n\|_F^2$, the norm of the corrupted atoms $\|M_n \atom_k\|_2$ can vary between 0 and 1 corresponding to the extreme cases of being completely destroyed, $M_n \atom_k = 0$, or perfectly preserved, $M_n \atom_k=\atom_k$. Therefore the proper dictionary representation of the corrupted signal is 
\begin{align}\label{eq:corr_rep}
M_n y_n =   \sum_{i\in I_n:\atop M_n \atom_i \neq 0} x_i \|M_n \atom_i\|_2 \cdot \frac{M_n \atom_i}{\|M_n \atom_i\|_2}
\end{align}
and, in order to recover the support $I_n$ via thresholding, we have to look at the inner products between the corrupted signal and the renormalised non-vanishing corrupted atoms,%. Setting $D_n(k,k) = \|M_n \atom_k\|_2^{-1}$ if $M_n \atom_k \neq 0$ and $D_n(k,k) = 0$ else we can write concisely
\begin{align} 
I_n^t =\arg\max_{I: |I| = S}  \sum_{i\in I: \atop M_n \atom_i \neq 0} \frac{| \ip{M_n \atom_i}{y_n}|}{\|M_n \atom_i\|_2} = \arg\max_{I: |I| = S}\sum_{i\in I} \| P(M_n \atom_i) M_n y_n\|_2.
\end{align}
Looking back at the representation of a corrupted signal in the properly scaled corrupted dictionary~\eqref{eq:corr_rep} we can also see why we assume flatness of the dictionary atoms, i.e. $\| \atom_k \|_\infty \ll 1$ for all $k$. In the ideal case where for all atoms $\atom_k$ we have $\|\atom_k\|_\infty=1/\sqrt{d}$ the energy
of the corrupted atoms will be constant $\|M_n \atom_k\|_2=\|M_n\|_F/\sqrt{d}$ so the dynamic range of the corrupted signal with respect to the corrupted normalised dictionary is the same as the original dynamic range,
\begin{align}
\frac{\max_{i \in I_n} |x_i| \|M_n \atom_i\|_2}{\min_{i \in I_n}|x_i| \|M_n \atom_i\|_2 } = \frac{\max_{i \in I_n} |x_i| }{\min_{i \in I_n}|x_i| } 
\end{align}
However, the less equally distributed over the coordinates the energy of the undamaged atoms is, the more the energy of the corrupted atoms varies. This leads to an increase of the dynamic range, which in turn makes it is harder for thresholding to recover the generating support.\\
The second reason for assuming flat atoms is the increase in coherence caused by the corruption. If the coherence of two flat atoms is small this means that their inner product is a sum of many small terms with different signs eventually almost cancelling each other out. Such a sum is quite robust under erasures, since both negative and positive terms are erased. On the other hand, if the energy of two atoms is less uniformly distributed, small coherence might be due to one larger entry in the sum being cancelled out by many small entries. Thus, the erasure of one large entry can cause a large increase in coherence, which again decreases the chances of thresholding recovering the generating support.\\
Finally, to see that the flatness-assumption is not merely necessary due to the imperfection of the thresholding algorithm for sparse recovery, assume that the atoms of the generating dictionary are combinations of two diracs $\atom_i=(\delta_i - \delta_{(i+1)})/\sqrt{2}$, the coefficients follow our simple sparse model and that the corruption takes the form of random erasures, i.e. $M_n(j,j)$ are iid Bernoulli variables with $P(M_n(j,j)=0)=\eta$. For large erasure probabilities, $\eta>1/2$, on average about half of the maximally $2S$ non zero entries of the signals will be erased and so the Dirac dictionary $\patom_i = \delta_i$ or rather its erased version will provide as plausible an $S$-sparse representation to the corrupted signals as the original dictionary $\dico$.\\ 
To see how to best modify the atom update rule, we first consider the case, where the corruption occurs always in the same locations, meaning $M_n =M$. Since we never observe the atoms on the coordinates where $M(k,k)=0$, we can only expect to learn the corrupted dictionary $M\dico = (M\atom_1\ldots M\atom_k)$ or rather its normalised version $(M\atom_k/ \|M \atom_k\|_2)$. On the other hand, the problem reduces to a simple dictionary learning problem for $M\dico$ instead of $\dico$ with update rule,
\begin{align}
M\patomn_k= \sum_{n: k\in I_n^t} \big[\I_d - P(M\pdico_{I_n^t}) + P(M\patom_k)\big] M y_n \cdot \signop(\ip{\patom_k}{M y_n}),
\end{align}
where we have used the fact that the projection onto a subdictionary is equal to the projection onto its normalised version and that $\signop(\ip{M\patom_k}{M y_n}/\|M \patom_k\|_2) =\signop(\ip{\patom_k}{M y_n})$.
Provided that thresholding always recovers the correct support $I_n$, we can conclude directly from above that the normalised corrupted dictionary will be a fixed point and that the update rule will contract towards it. Indeed, for any corruption pattern $M$ we know that before normalisation an updated atom $M\patomn_k$ will be contracted towards $N_k =\sharp \{n: k\in I_n\}$ scaled copies of the corrupted generating atom $M\atom_k$,
\begin{align}
\sum_{n: k\in I_n} \big[\I_d - P(M\pdico_{I_n}) + P(M\patom_k)\big] M y_n \cdot \signop(\ip{\patom_k}{M y_n})\quad  \rightsquigarrow \quad N_k \cdot c M\atom_k = c \cdot \sum_{n: k\in I_n} M \atom_k.
\end{align}
This suggests that for the case of different corruption patterns $M_n$ we can simply replace $M$ by $M_n$ and the updated atom will be contracted towards the sum of scaled copies of the generating atom, corrupted with the different patterns, 
\begin{align} 
 \sum_{n: k\in I_n} \big[\I_d - P(M_n\pdico_{I_n}) + P(M_n\patom_k)\big] M_n y_n \cdot \signop(\ip{\patom_k}{M y_n}) \quad \rightsquigarrow\quad c \cdot \sum_{n: k\in I_n} M_n \atom_k.
\end{align}
To then reconstruct the generating atom from the sum of its corrupted copies we just need to count how often we observe the atom on each coordinate. If each coordinate has been observed at least once,
we can then obtain the generating atom simply by rescaling according to the number of observations, meaning we calculate
\begin{align}
\patomn_k= \sum_{n: k\in I_n^t} \big[\I_d - P(M_n\pdico_{I_n^t}) + P(M_n \patom_k)\big] M_n  y_n \cdot \signop(\ip{\patom_k}{M_n y_n}) \notag\\\mbox{and}\qquad W_k=\sum_{n: k\in I_n^t}  M_n,\notag
%\patomn_k= \sum_{n: k\in I_{n}^t} \big[\I - P&(\pdico_{I_{n}^t})  + P(\patom_k) \big] y_n \cdot \signop(\ip{\patom_k}{y_n}),\notag %\cdot  \chi(I_n^t, k),
\end{align}
set $\bar \patomn_k = W_k^{\dagger} \patomn_k$
and output $\ppdico=(\bar \patomn_1/\|\bar \patomn_1\|_2, \ldots, \bar \patomn_K/\|\bar \patomn_K\|_2)$.\\
The last detail we need to account for is the possible existence of a low-rank component $\lrcomp$; other than noise or different signal scalings its contribution cannot be expected to average out once we have enough observations. Fortunately removing the low-rank component is quite straightforward, once we have a good estimate $\tilde \lrcomp$ with $P(\tilde \lrcomp)\lrcomp \approx \lrcomp$. If a signal contains a low-rank component then the corrupted signal will contain the corrupted component, 
$My = M\lrcomp \lrc + M\dico_{I} x(I)$ and we can remove its contribution by a simple projection $M\tilde y = Q(M\tilde \lrcomp) My$. However, since the mask destroys the orthogonality between the dictionary and the low-rank component, we do not get only the sparse contribution $M\dico_{I} x(I)$ but also a (small) contribution of the low-rank component, $Q(M\tilde \lrcomp) M\dico_{I} x(I) = M\dico_{I} x(I) - P(M\tilde \lrcomp) M\dico_{I} x(I)$. Thus, to stably estimate which part of an atom in the support has not been captured yet, we need to remove also the low-rank contribution and in our update rule replace the projection onto the current estimate of the corrupted atoms in the support with the projection onto these and the (estimated) corrupted low-rank component, $P(M_n\pdico_{I_n^t}) \rightarrow P(M_n(\tilde \lrcomp,\pdico_{I_n^t}))$. Further, to ensure that the output dictionary is again orthogonal to the low-rank component, we project the updated atoms onto the orthogonal complement of the (estimated) low-rank component.
Putting it all together, we arrive at the following modified algorithm.
\begin{algorithm}[ITKrM for corrupted data - one iteration] 
Given an estimate of the low-rank component $\tilde \lrcomp$, an input dictionary $\pdico$ with $\pdico^\star \tilde \lrcomp = 0$, a sparsity level $S$ and $N$ corrupted training signals $y^M_n=(M_ny_n,M_n)$ do:
\begin{itemize}
\item For all $n$ set $M_n \tilde y_n = Q(M_n\tilde \lrcomp) M_n y_n$.
\item For all $n$ find $I_n^t= \arg\max_{I: |I| = S}  \sum_{i\in I: M_n \atom_i \neq 0} \frac{| \ip{M_n \atom_i}{M_n\tilde y_n}|}{\|M_n \atom_i\|_2}$.
\item For all $k$ calculate%\footnote{For a matrix M $P(M)$ denotes the orthogonal projection onto the rowspan of $M$}. 
\begin{align}
\patomn_k= \sum_{n: k\in I_n^t} \big[\I_d - P(M_n(\tilde \lrcomp,\pdico_{I_n^t})) + P(M_n \patom_k)\big] M_n \tilde y_n \cdot \signop(\ip{\patom_k}{M_n \tilde y_n})  \\
\mbox{and}\qquad W_k=\sum_{n: k\in I_n^t}  M_n .
%\patomn_k= \sum_{n: k\in I_{n}^t} \big[\I - P&(\pdico_{I_{n}^t})  + P(\patom_k) \big] y_n \cdot \signop(\ip{\patom_k}{y_n}),\notag %\cdot  \chi(I_n^t, k),
\end{align}
\item Set $\bar \patomn_k = Q(\tilde \lrcomp) W_k^{\dagger} \patomn_k$
and output $\ppdico=(\bar \patomn_1/\|\bar \patomn_1\|_2, \ldots, \bar \patomn_K/\|\bar \patomn_K\|_2)$.
\end{itemize}
\end{algorithm}
Before we can start testing the modified algorithm, we still need to develop a method for actual recovery of the low-rank component from corrupted data, which we will do in the next section. In a follow up paper we hope to provide a theoretical analysis, which confirms that - as planned - the modified algorithm retains both the fixed point and the contraction property and thus is locally convergent.

%%%%%%%%%%%%%%
%!TEX root = maskdl.tex
%%%%%%%%%%%%%%%

%%%%%%%%%%%
\section{Recovery of the low-rank component}\label{sec:lr}
%%%%%%%%%%%
As already mentioned, in the case of uncorrupted signals the low-rank component can be straightforwardly removed, since $\Gamma$ will correspond to the $L$ left singular vectors associated to the largest $L$ singular values of the data matrix. In the case of corrupted signals this is no longer possible since the action of the corruption will distort the left singular vectors in the direction of the more frequently observed coordinates. 
To counter this effect, one would have to include the mask information in the singular value decomposition. This is, for instance, done by Robust PCA which was developed for the related problem of low-rank matrix completion \cite{calimawr11}. Unfortunately, one of the main assumptions therein is that the corruption is homogeneously spread among the coordinates, which might not be the case in our setup. To recover the low rank component, we will, therefore, pursue a different strategy.\\
Let us assume for a moment that we are looking for only one low-rank atom, $L=1$. One interpretation of all (masked) signals having a good part of their energy captured by the (masked) low-rank atom is to say that all (masked) signals are 1-sparse in a dictionary of one (masked) atom. Since we already have an algorithm to learn dictionaries from corrupted signals, we can also employ it to learn the low-rank atom. Moreover, since we have an algorithm to learn dictionaries from corrupted signals that contain a low-rank component, we can iteratively learn the low-rank component atom by atom. Adapting the algorithm also leads to some simplifications. After all, we do not need to find the sparse support, since (almost) all signals are expected to contain the one new atom. Summarising these considerations, we arrive at the following algorithm.
%%%% LR Component Algorithm
\begin{algorithm}[low-rank atom recovery from corrupted data - one iteration] 
Given an estimate of the previously recovered low-rank component $\tilde \lrcomp = (\tilde \lratom_1 \ldots \tilde \lratom_{\ell-1}),$ an input low-rank atom $\hat \lratom_\ell$ and $N$ corrupted training signals $y^M_n=(M_ny_n,M_n)$ do:
\begin{itemize}
\item For all $n$ set $M_n \tilde y_n = Q(M_n\tilde \lrcomp) M_n y_n$.
\item Calculate 
\begin{align}
 \plratom_\ell = \sum_{n} \big[\I_d - P(M_n(\tilde \lrcomp,\hat \lratom_\ell)) + P(M_n \hat \lratom_\ell)\big] M_n \tilde y_n \cdot \signop(\ip{\hat \lratom_\ell}{M_n \tilde y_n})  \\
\mbox{and}\qquad W=\sum_{n}  M_n .
\end{align}
\item Set $\bar \plratom_\ell = Q(\tilde \lrcomp) W^{\dagger} \plratom_\ell$
and output $\bar \plratom_\ell/\|\bar \plratom_\ell \|_2$.
\end{itemize}
\end{algorithm}

Note that for the first low-rank atom in each iteration the update rule reduces to a summation of the signals aligned according to $\signop(\ip{\hat \lratom_\ell}{M_n y_n})$.  
We also want to point out that our iterative approach offers the possibility to automatically choose the size of the low-rank component by comparing the signal energy captured by the last low-rank atom $\sum_n \|P(M_n \lratom_\ell) M_n  y_n\|_2^2=\sum_n |\ip{\lratom_\ell}{M_n y_n}|^2/\| M_n\lratom_\ell\|^2_2$ to the signal energy expected to be captured by a dictionary atom $\frac{1}{K} \sum_n \| M_n  y_n \|_2^2$. A natural choice for determining $L$ is to stop adding low rank atoms once the ratio between the two energies drops below a prescribed value. Since this ratio depends very much on the problem at hand, we will not go into more details here but instead turn to finally testing our algorithms on synthetic data.

% !TEX root = maskdl.tex
%%%%%%%%%%%
\section{Numerical simulations: synthetic data}\label{sec:tests}
%%%%%%%%%%%

%%%%%%%%%%%%%%% Table Signal Distribution
\begin{table}[b]
\fbox{\parbox{\textwidth}{{\bf Signal Model}\\
Given the generating low-rank component $\lrcomp$ and dictionary $\dico$ our signal model further depends on 6 coefficient parameters, 
\vspace{0.5em}

\begin{tabular}{lcl}
$e_\lrcomp$ &-& the energy of the low-rank coefficients,\\
$b_\lrcomp$ &-& defining the decay factor of the low-rank coefficients,\\
$S$ &-& the sparsity level, \\
$b_S$ &-& defining the decay factor of the sparse coefficients,\\
$\nsigma$ &-& the noise level and\\
$\scale_{m}$ &-& the maximal signal scale. \\
\end{tabular}
\vspace{0.5em}

Given these parameters, we choose a low-rank decay factor $c_\lrcomp$ uniformly at random in the interval $[1-b_\lrcomp,1]$. We set $\lrc(\ell) =\sigma_\ell c_\lrcomp^\ell$ for $1\leq \ell \leq L$, where $\sigma_\ell$ are iid. uniform $\pm 1$ Bernoulli variables, and renormalise the sequence to have norm $\|v\|_2 = e_\lrcomp$. Similarly, we choose a decay factor $c_S$ for the sparse coefficients uniformly at random in the interval $[1-b_S,1]$. We set $x(k) = \sigma_k c_S^k$ for $1\leq k \leq S$, where $\sigma_\ell$ are iid. uniform $\pm 1$ Bernoulli variables, and renormalise the sequence to have norm $\|x\|_2 =1- e_\lrcomp$. Finally, we choose a support set $I=\{i_1 \ldots i_S\}$ uniformly at random as well as a scaling factor $\scale$ uniformly at random from the interval $[0,\scale_{m}]$ and according to our signal model in \eqref{noisymodel1} set
$$y= \scale \cdot \frac{ \lrcomp \lrc  + \dico_I x +\noise}{\sqrt{1+\|\noise \|_2^2}},$$
where $r$ is a Gaussian noise-vector with variance $\nsigma^2$ if $\nsigma>0$.
}}\caption{Signal Model \label{coeffmodel}}
\end{table}
%%%%%%%%%%%%%%% End Table Signal Distribution

We first explore the performance of the adapted algorithms in comparison to their original counterparts not using mask information, that is, singular value decomposition for low-rank recovery and for dictionary learning ITKrM on the signals projected onto the orthogonal complement of the low-rank component. To do this, we look at two representation pairs, consisting of a low-rank component and a dictionary, and test the recovery using 6-sparse signals with corruptions of two types, random erasures and burst errors. \\
{\bf Dictionary \& low-rank component:} The first representation pair corresponds to the Discrete Cosine Transform (DCT) basis in $\R^d$ for $d=256$. As low-rank component we choose the first two DCT atoms, that is the constant atom and the atom corresponding to an equidistant sampling of the cosine
on the interval $[0,\pi)$, while the remaining basis elements form the dictionary. For the second pair, we construct the low-rank component by choosing two vectors uniformly at random on the sphere in $\R^d$ for $d=256$ and setting $\Gamma$ the closest orthonormal basis as given by the singular value decomposition. To create the dictionary, we then choose another $1.5d$ random vectors uniformly on the sphere, project them onto the orthogonal complement of the span of $\Gamma$ and renormalise them. These two representation pairs exhibit different complexities. The first forms an orthonormal basis, thus is maximally incoherent, and every element has $\|\gamma_\ell\|_\infty = \|\atom_k\|_\infty = \sqrt{2/d}\approx 0.088$. The second dictionary is overcomplete with coherence $0.2788$ and the supremum norm of both the low-rank and the dictionary atoms varies between $0.1529$ and $0.2754$ and averages at $0.1897$. \\
{\bf Signals:} To create our signals, we use the signal model in \eqref{noisymodel1} with a particular choice of distributions for the sparse and low-rank coefficients, the scaling factor and the noise, described in Table~\ref{coeffmodel}. For the first experiment, we set the parameters to $e_\lrcomp = 1/3$, $b_\lrcomp=0.15$, $S=6$, $b_S=0.1$, $\nsigma=1/(4\sqrt{d})$ and $\scale_{m}=4$, resulting in 6-sparse signals with dynamic coefficient range between 1 and $0.9^{-6}\approx 1.88$ and the low-rank component containing a third of the energy. The signal-to-noise ratio is 16 and the scaling is uniformly distributed on [0,4].\\
{\bf Corruption:} We consider two types of corruptions, whose distributions are described in Table~\ref{maskmodels}. The random erasure patterns depend on 4 parameters determining (the difference in) the erasure probabilities of the first and second half of the coordinates ($p_1,p_2$) and one half and the other half of the signals ($q_1,q_2$). The expected average corruption corresponds to $1-\E(\sum_k M(k,k)) = 1- (p_1+p_2)(q_1+q_2)/4$ and in our experiments varies between 10\% and 90\%. \\
The burst error patterns also depend on 4 parameters determining the burstlength $T$, the probability of no burst, a burst of size $T$ or of size $2T$ occurring ($p_0,p_T, p_{2T}$ where $p_0 = 1- p_T - p_{2T}$), as well as the probability of the burst occurring among the first half of the coordinates ($q$). In our experiments, we consider burstlengths $T=64,96$ with varying burst location and occurrence probabilities, leading to an empirical average corruption varying between 10\% and 60\%.

%%%%%%%%%%%%%%% Table Mask Distribution
\begin{table}[tb]
\fbox{\parbox{\textwidth}{{\bf Erasure Model}\\
Our erasure model depends on 4 parameters, \vspace{0.5em}

\begin{tabular}{lcl}
$p_1$ &-& the relative signal corruption of the first half of coordinates,\\
$p_2$ &-& the relative signal corruption of the second half of coordinates,\\
$q_1$ &-& the corruption factor of one half of the signals, \\
$q_2$ &-& the corruption factor of the other half of the signals.\\
\end{tabular}
\vspace{0.5em}

Based on these parameters we generate a random erasure mask as follows. First we choose $q \in \{q_1,q_2\}$ uniformly at random and determine
for every entry the probability of being non-zero as $\eta_j = q p_1$ for $j\leq d/2$ and $\eta_j = q p_2$ for $j > d/2$. We then generate a mask as a realisation of the independent Bernoulli variables $M(j,j)\sim B(\eta_j)$, that is $P(M(j,j)=1)=\eta_j$.\\
\\
{\bf Burst Error Model}\\
Our burst error model depends on 4 parameters, \vspace{0.5em}

\begin{tabular}{lcl}
$p_T$ &-& the probability of a burst of length $T$,\\
$p_{2T}$ &-& the probability of a burst of length $2T$,\\
$T$ &-& the burst length,\\
$q$ &-& the probability of the burst starting in the first half of the coordinates.\\
\end{tabular}
\vspace{0.5em}

Based on these parameters, we generate a burst error mask as follows. First we choose a burstlength $\tau \in \{0, T, 2T\}$ according to the probability distribution prescribed by $\{ p_0, p_T, p_{2T}\}$, where $p_0=1-p_T-p_{2T}$. We then decide according to the probability $q$ whether the burst start $t$ occurs among the first half of coordinates, $t\leq d/2$, or the second half, $t> d/2$. Finally, we draw the burst start $t$ uniformly at random from the chosen half of coordinates and in a cyclic fashion set $M(j,j)=0$ whenever $t\leq j < t+ \tau$ or $ j < t+\tau-d$ and $M(j,j)=1$ else. 
}}\caption{Mask Models\label{maskmodels}}
\end{table}
%%%%%%%%%%%%%%% End Table Mask Distribution

\noindent{\bf Experimental setup:} We use two kinds of experimental setups. In the first, we learn the low-rank component and then the dictionary always using random initialisations. In particular, to learn the low-rank component with the adapted algorithm we use 10 iterations for every atom and 30,000 (new) signals per iteration. As initialisation, we use a vector drawn uniformly at random from the sphere in the orthogonal complement of the low-rank component recovered so far. For the unadapted low-rank recovery, we use a singular value decomposition, where the low-rank component corresponds to the first $L$ left singular vectors of the last 30,000 signals generated for the adapted algorithm. As measure for the final recovery error, we use the operator norm of the difference between the generating low-rank component $\Gamma$ and its projection onto the recovered component $\tilde \Gamma$, that is $\|\Gamma- P(\tilde \Gamma) \Gamma\|_2$. This corresponds to the worst case approximation error of a signal in the span of the generating low-rank component by the recovered one.\\
We then learn the dictionary using 100 iterations of ITKrM(M) and 100,000 (new) signals per iteration from a random initialisation, where the initial  atoms are drawn uniformly at random from the sphere in the orthogonal complement of the respective low-rank component. We measure the recovery success by the percentage of recovered or rather not recovered atoms, where we use the convention that a generating atom $\atom_k$ is recovered if there exists an atom $\tilde\patom_j$ in the output dictionary $\tilde\pdico$ for which $|\ip{\atom_k}{\tilde\patom_j}|\geq t$ for $t=0.99$. To provide a more complete picture, we also indicate the percentage of atoms, that are potentially recoverable using more training samples or iterations, that is $|\ip{\atom_k}{\tilde\patom_j}|\geq t$ for $t=0.90$. \\
In the second setup, we do not learn the low-rank component but provide the dictionary learning algorithms with the true low-rank component. We then learn the dictionary from a close-by initialisation, using 10 iterations with again 100,000 (new) signals per iteration. Here, the initial dictionary  atoms are 1:1 linear combinations of a generating atom and a random vector in the sphere orthogonal to it, which are then projected onto the orthogonal complement of the respective low-rank component. In case of the close-by initialisation, we measure the recovery success by the maximal error between a generating atom and its closest match in the output dictionary, 
\begin{align}
d_\infty(\dico,\tilde\pdico) := \max_k \min_j \| \atom_k \pm \tilde \patom_j\|_2 = \max_k \min_j \sqrt{2-2 |\ip{\atom_k}{\tilde\patom_j}|}.
\end{align}
Note that for an atom the recovery threshold $t=0.99$ corresponds to an error of $\sqrt{0.02} \approx 0.14$. Again to provide a more complete picture, we also indicate the mean error between all generating atoms and their closest matches in the output dictionary, $d_1(\dico,\tilde\pdico)=\frac{1}{K} \sum_k \min_j \| \atom_k \pm \tilde \patom_j\|_2$.

%%%%%%%f
\begin{figure}[htbp]
\begin{tabular}{cc}
%%%epsfigures - first time - slow
% \includegraphics[width=8cm]{plots/dct_lr.eps} & \includegraphics[width=8cm]{plots/rand_lr.eps} \\
%  (a) & (b)\\
%   \includegraphics[width=8cm]{plots/dct_rand.eps} & \includegraphics[width=8cm]{plots/rand_rand.eps}\\
%  (c) & (d)\\
%   \includegraphics[width=8cm]{plots/dct_close.eps} & \includegraphics[width=8cm]{plots/rand_close.eps}\\
%  (e) & (f)\\
%%%pdffigures 
 \includegraphics[width=8cm]{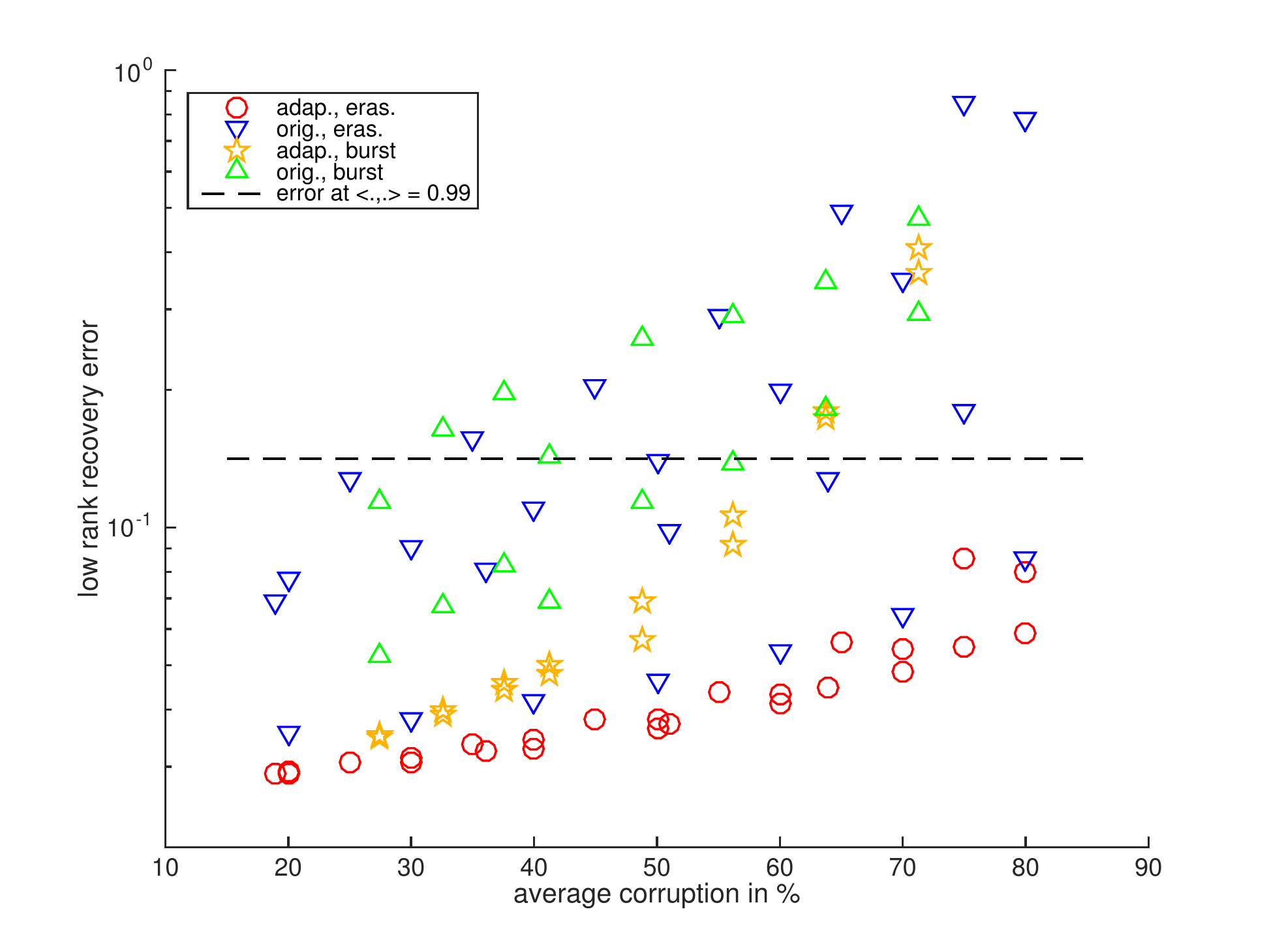} & \includegraphics[width=8cm]{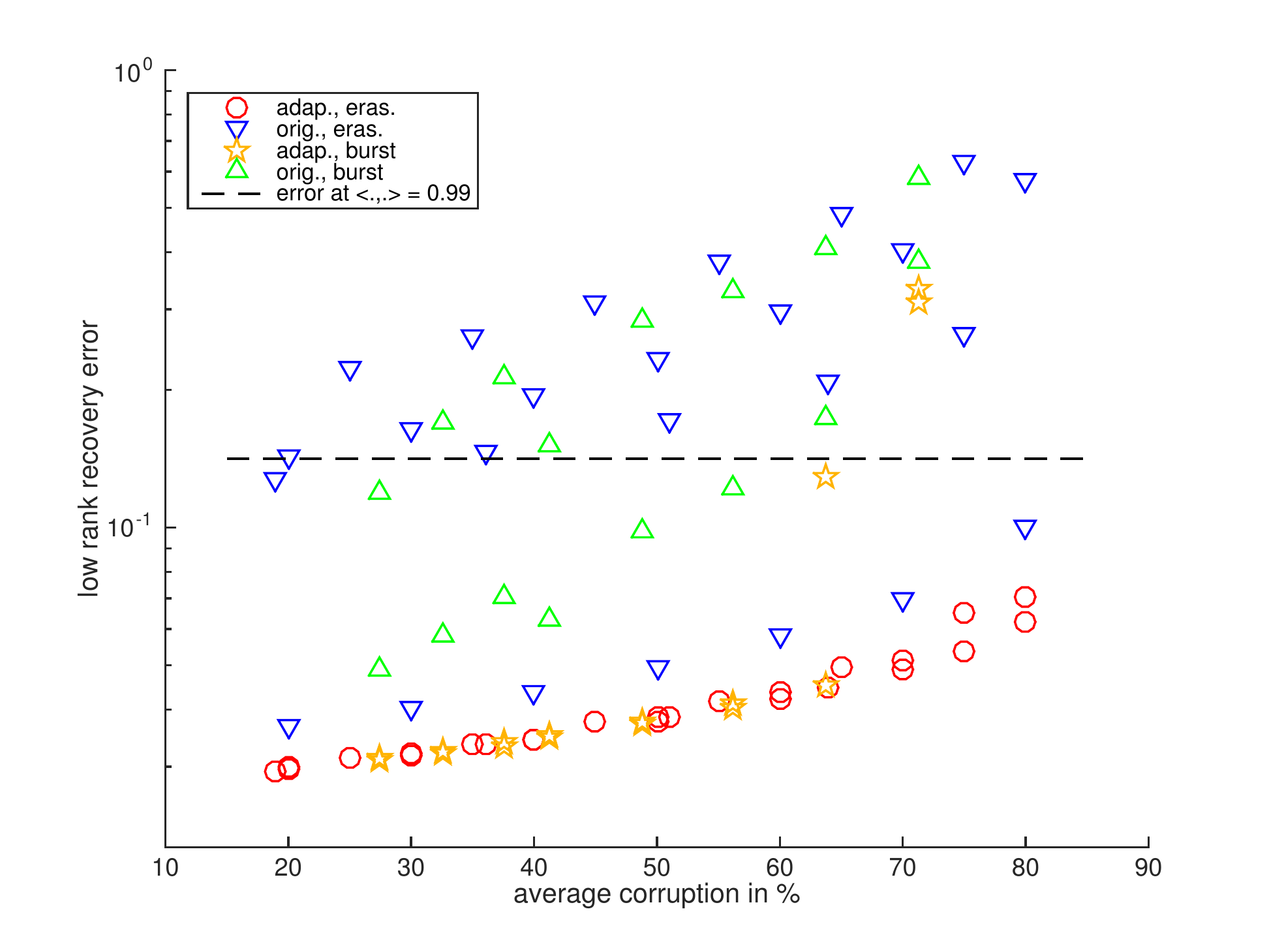} \\
  (a) & (b)\\
   \includegraphics[width=8cm]{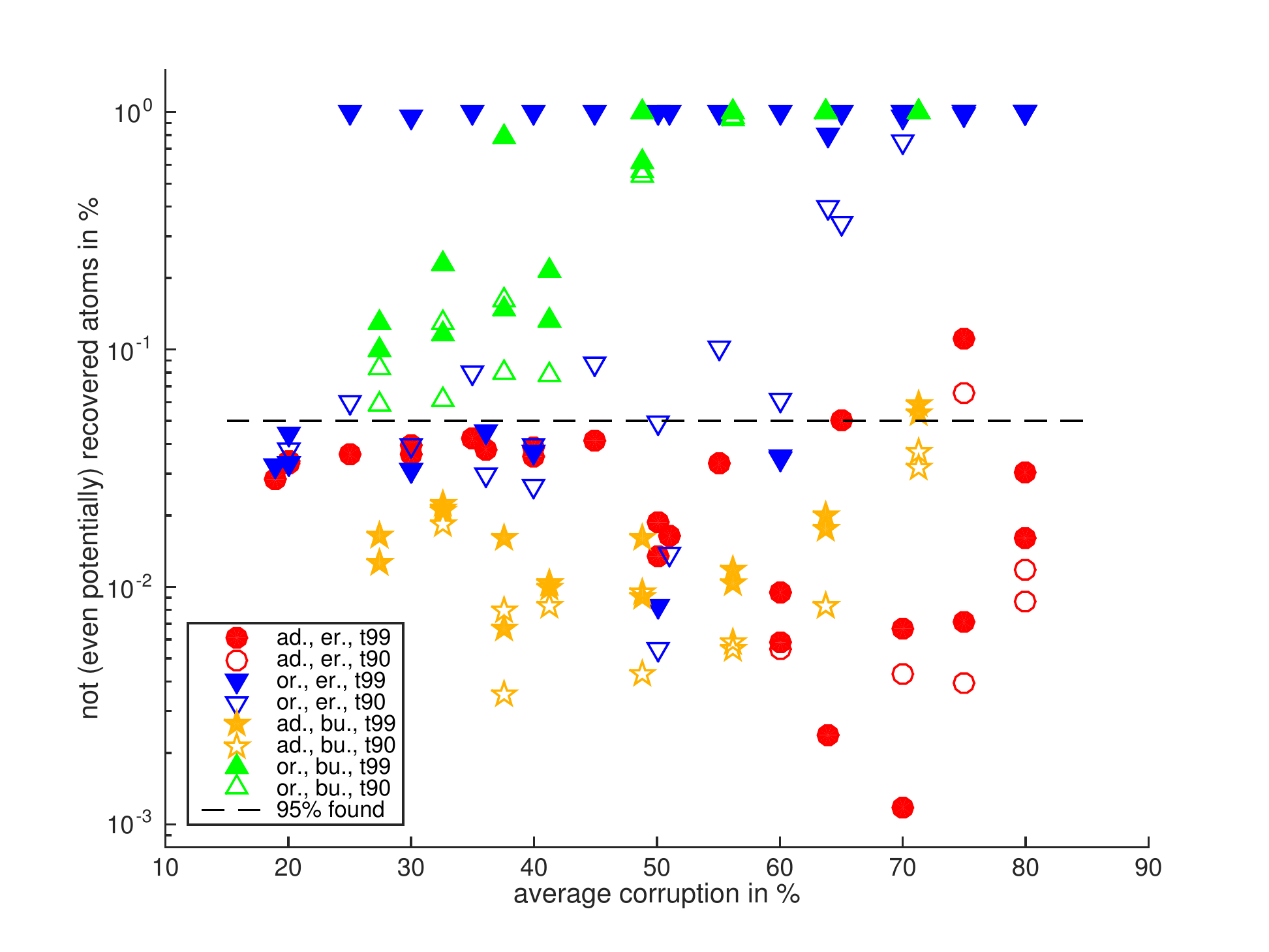} & \includegraphics[width=8cm]{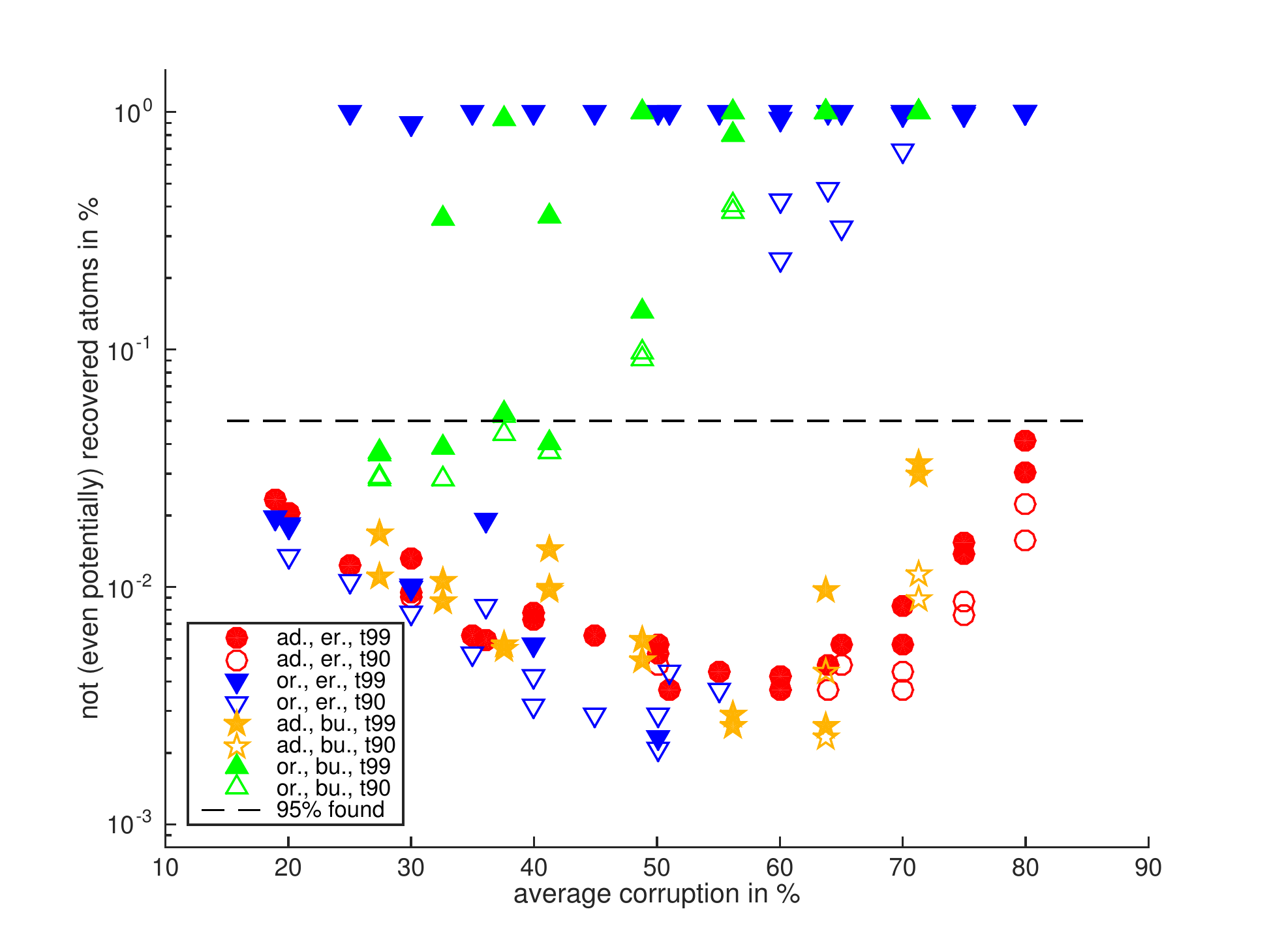}\\
  (c) & (d)\\
   \includegraphics[width=8cm]{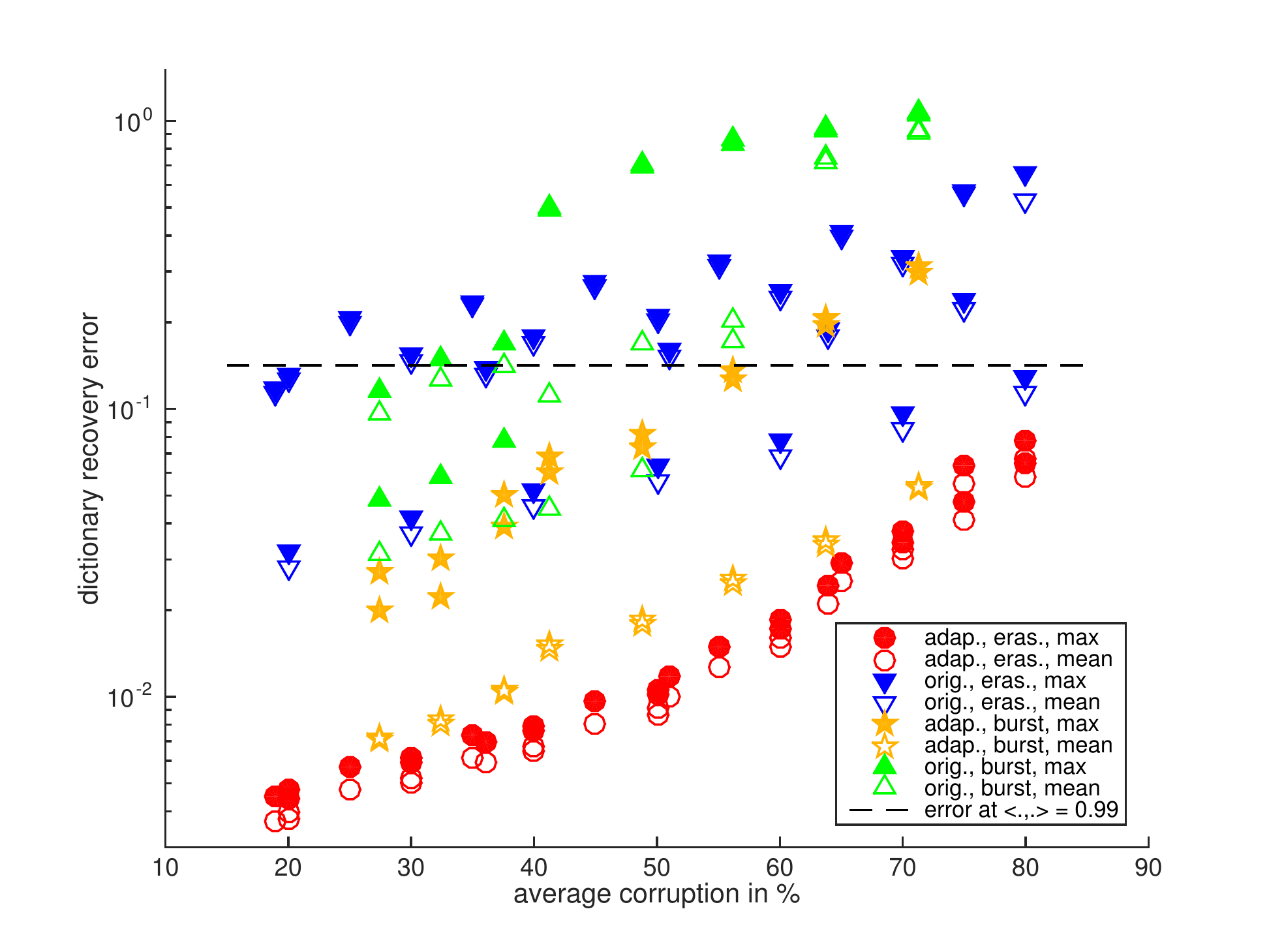} & \includegraphics[width=8cm]{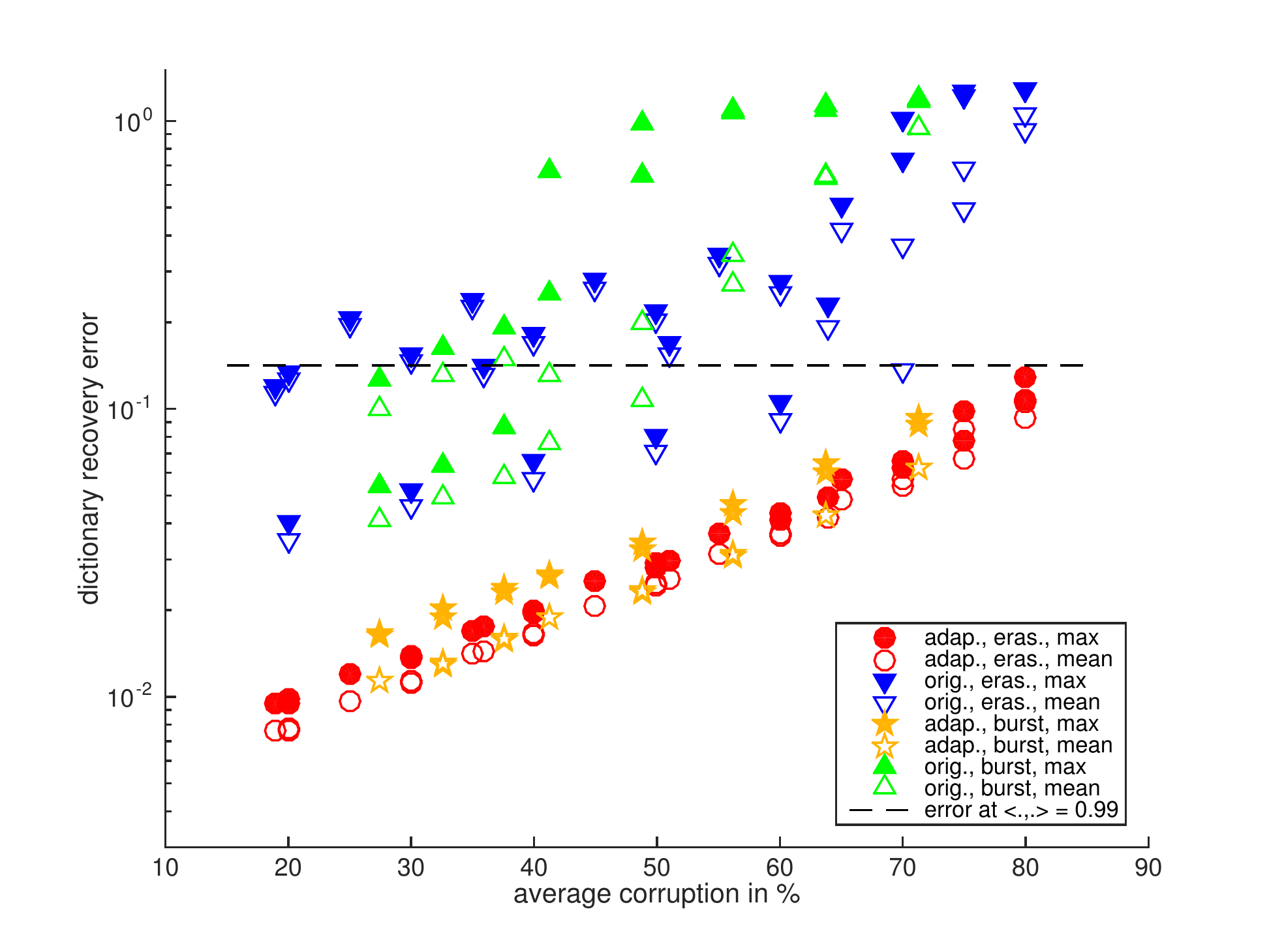}\\
  (e) & (f)\\
  \end{tabular}
    \caption{Recovery performance of the corruption adapted versus the unadapted learning algorithms for the DCT (a,c,e) and the random (b,d,f) representation pair in terms of low-rank recovery error (a,b), percentage of recovered dictionary atoms from a random initialisation (c,d) and dictionary recovery error from a close-by initialisation (e,f). \label{synth_fig1}}
\end{figure}
%%%%%%%%%%

Figure~\ref{synth_fig1} shows the recovery results for various corruption levels using the corruption adapted algorithms (ITKrMM) and their unadapted counterparts (ITKrM). 
We can see that for both representation pairs incorporating the corruption information into the learning algorithms clearly improves the performance. Another fact immediately visible is that for the adapted algorithms the success rates differ for the two erasure modalities and decrease with increasing corruption but do not depend much on the particular distribution of the erasures or bursts as long as they lead to the same average corruption level. In contrast, the success rates of the unmodified algorithms depend very much on the corruption distribution, and signals with similar average corruption can lead to very different error rates.\\
Distinguishing between the different error modalities, we note that, for the low-rank recovery and the dictionary recovery from a close-by initialisation, the corruption adapted algorithms outperform the unadapted ones in any setting. For the dictionary recovery from a random initialisation, we again see that overall the modified algorithm outperforms the unmodified one. So, for both dictionaries ITKrMM recovers more than 95\% of the atoms in all except two cases, while recovery, as well as potential recovery, via ITKrM completely breaks down at around 60\% corruption. \\
We also observe that corruption can improve the recovery rates of both the unmodified and the modified algorithms. A similar phenomenon has already been observed for ITKrM in connection with noise and a lower sparsity level, \cite{sc15}. So while one might expect the global recovery rates to decrease with increasing noise and with $S$, they actually increase. The reason for this is that a little bit of noise or lower sparsity, like a little bit of corruption, breaks symmetries and suppresses the following phenomenon. Two atoms converge to the same generating atom and, therefore, another atom has to do the job (is a 1:1 linear combination) of two generating atoms. For uncorrupted signals there are ongoing efforts to counter this phenomenon with replacement strategies, which have a straightforward extension to corrupted signals, \cite{sc17}. With this extension, we expect all (potential) recovery rates around 90\% to increase to 100\%, as happens in the uncorrupted case.\\
%%%%%%%f
\begin{figure}[htbp]
\begin{tabular}{cc}
%%%epsfigures - first time - slow
% \includegraphics[width=8cm]{plots/randlr_mdet.eps} & \includegraphics[width=8cm]{plots/randlr_tdet.eps} \\
%  (a) & (b)\\
%   \includegraphics[width=8cm]{plots/randdico_mdet.eps} & \includegraphics[width=8cm]{plots/randdico_tdet.eps}\\
%  (c) & (d)\\
%%%pdffigures 
 \includegraphics[width=8cm]{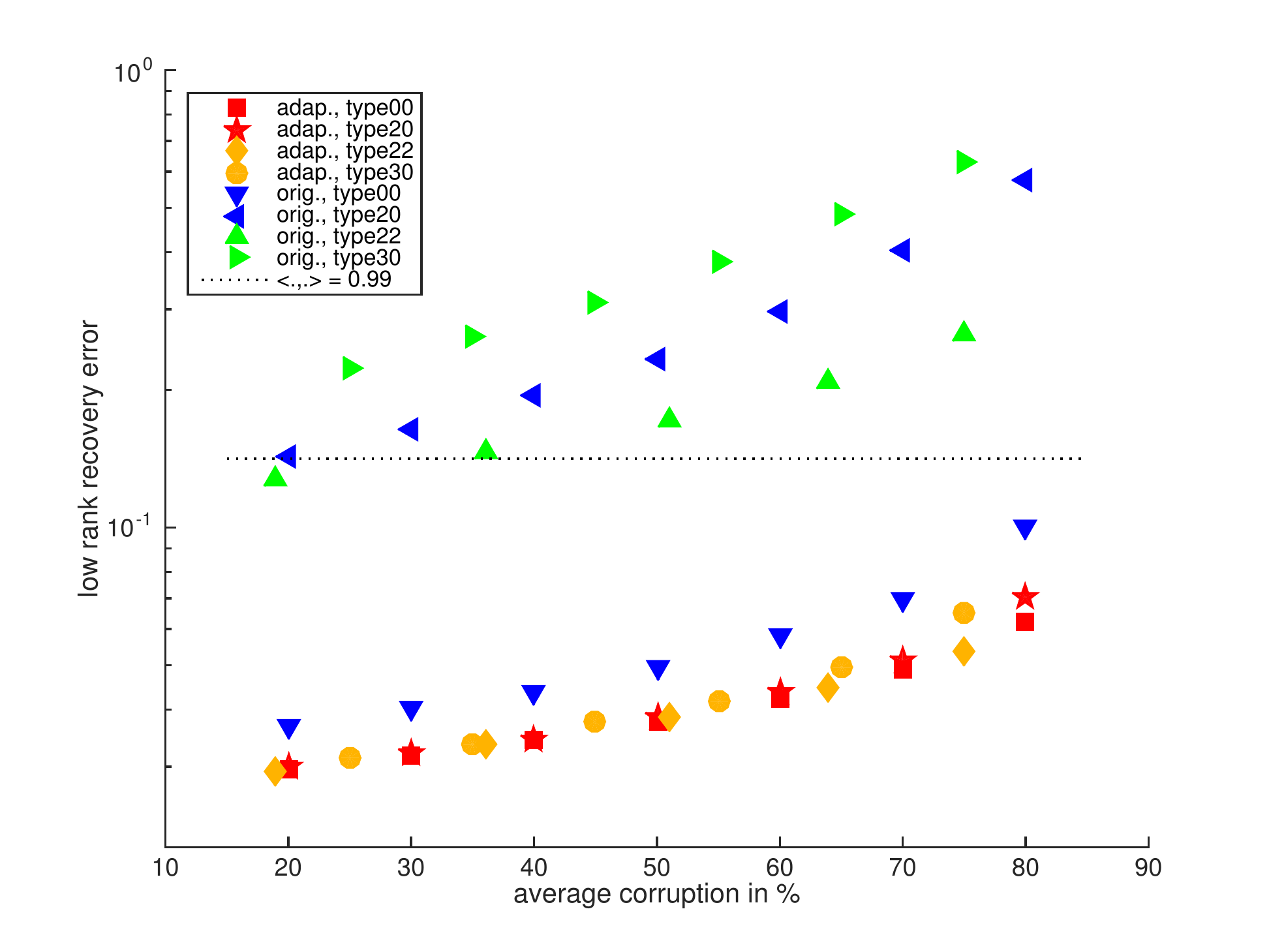} & \includegraphics[width=8cm]{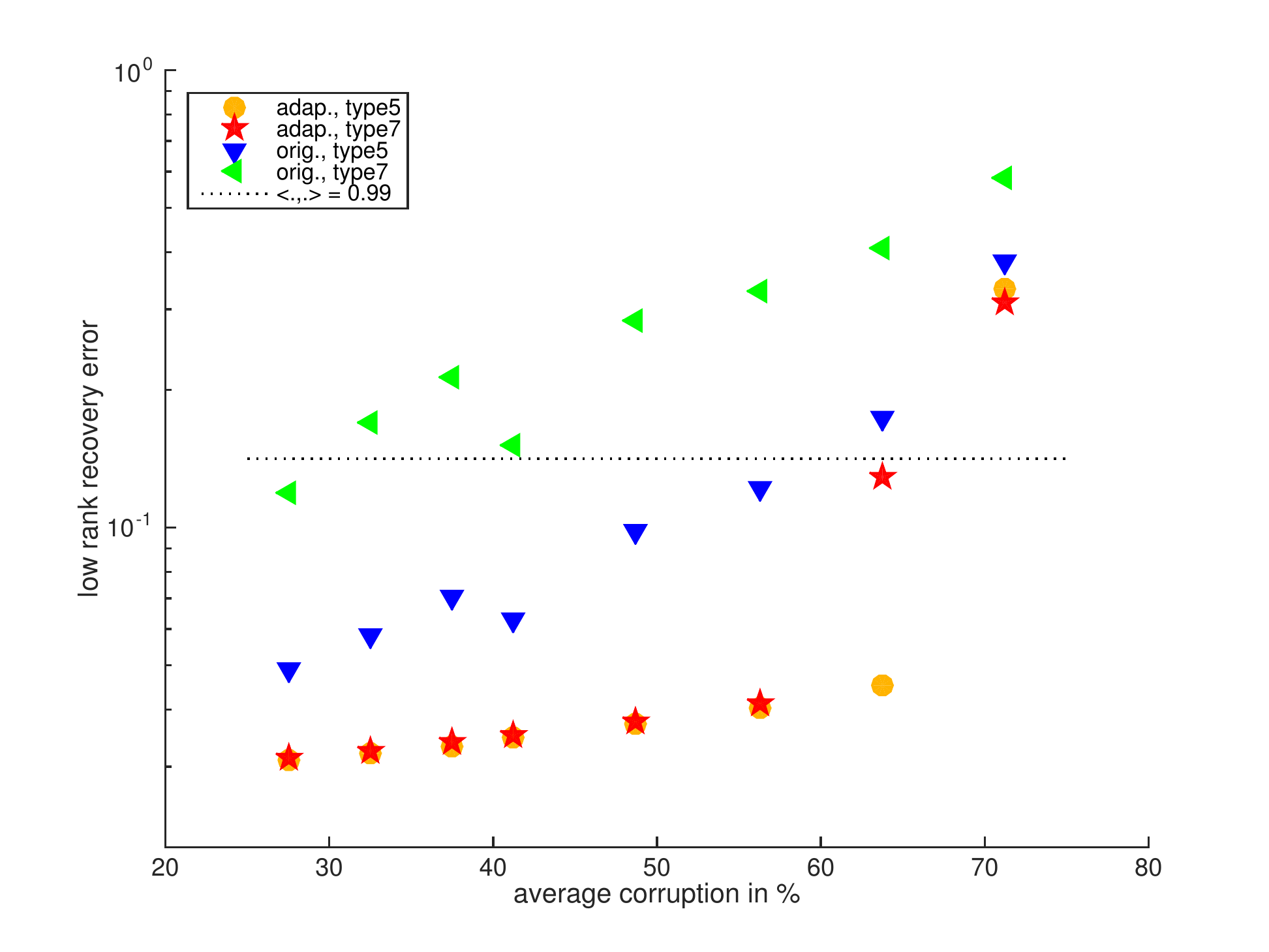} \\
  (a) & (b)\\
   \includegraphics[width=8cm]{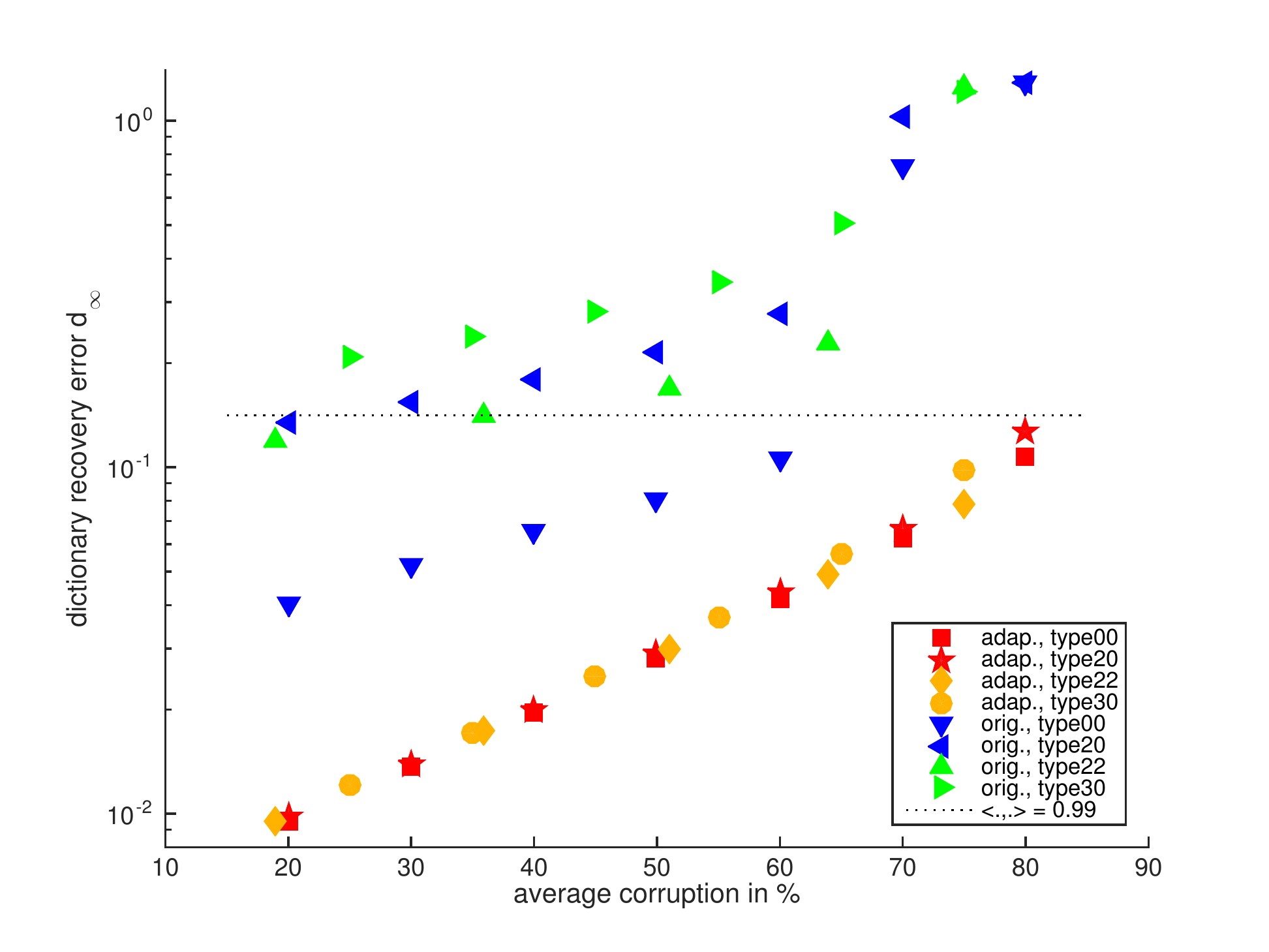} & \includegraphics[width=8cm]{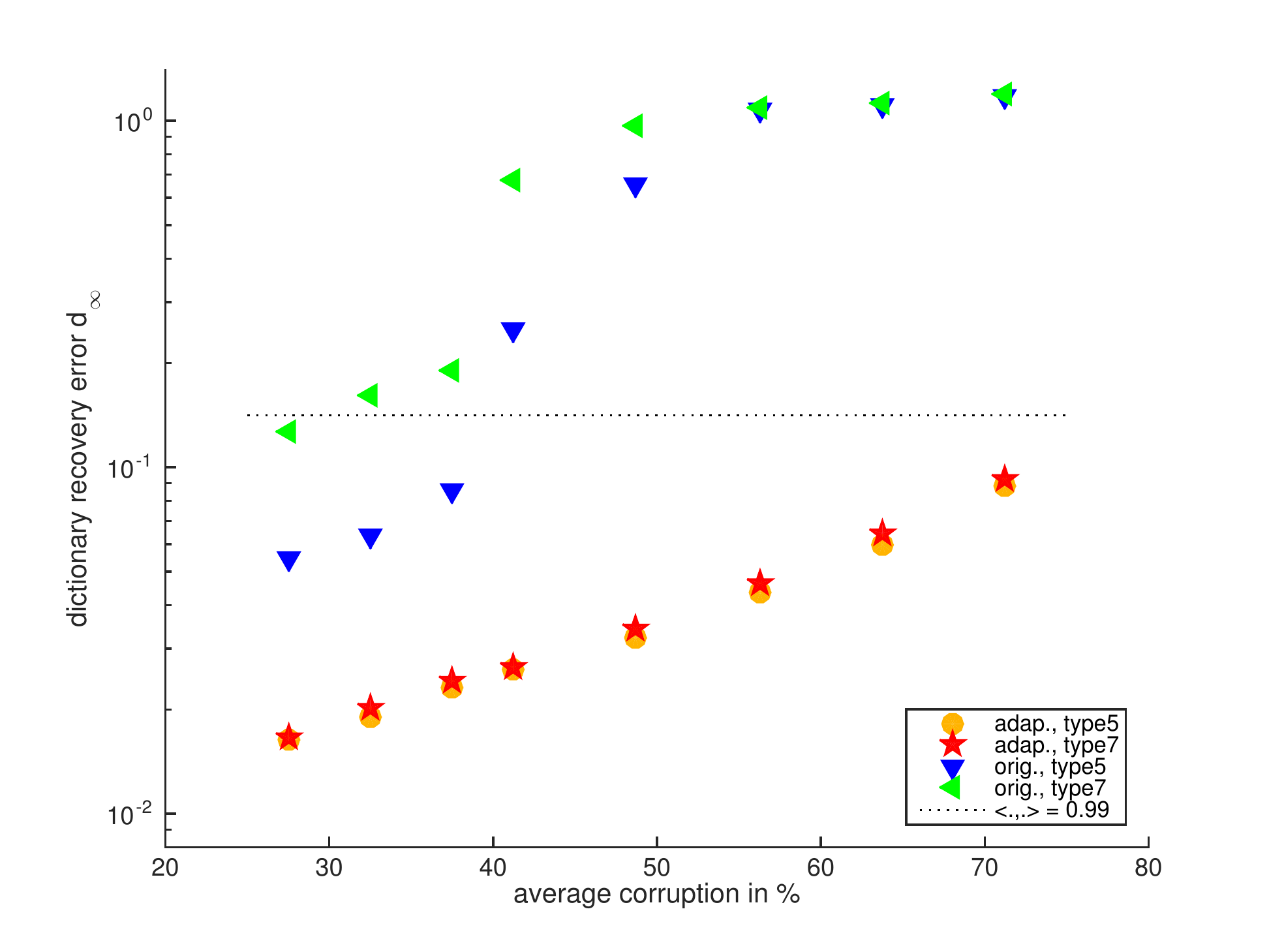}\\
  (c) & (d)\\
  \end{tabular}
    \caption{Detailed recovery performance of the corruption adapted versus the unadapted learning algorithms for the random representation pair for various types of random erasures (a,c) and burst errors (b,d). \label{synth_fig2}}
\end{figure}
%%%%%%%%
To find out when we gain most from incorporating the mask information, let us have a more detailed look at the recovery rates for different types of parameter settings. 
In case of the random erasures, we distinguish 4 types. 'type00' indicates that $p_1=p_2$ with $p_1$ varying between $0.2$ and $0.8$ and $q_1=q_2=1$, leading to a uniform erasure probability for all coordinates and all signals. 'type20(30)'  indicate that $p_2=p_1+0.2(0.3)$ with $p_1$ varying between $0.1$ and $0.7(0.6)$ and again $q_i=1$, leading to higher erasure probabilities for the first half of the coordinates, which are however uniform across signals. Finally, 'type22' indicates that $p_2=p_1+0.2$ and $q_i=p_i$ for $p_1$ varying between $0.4$ and $0.8$, leading to different erasure probabilities across coordinates and across signals.\\
In case of the burst errors we distinguish between 'type5' corresponding to a uniform burst distribution and 'type7' corresponding to a 0.7 probability of the burst occurring in first half of the coordinates. For each burst type, we consider the burstlength $T=64$ with probabilities $(p_T,p_{2T})\in\{ (0.5,0.3), (0.7,0.3), (0.5,0.5)\}$, leading to corruptions between 20\% and 40\%, and the burstlength $T=96$ with 
the same pairs and additionally $(p_0,p_T)\in\{(0.3, 0.7), (0.1,0.9)\}$ leading to corruptions between 40\% and 75\%. \\
For conciseness, we focus on the random low-rank component and dictionary and show only the error of the low-rank component and the maximal error of the dictionary from a close-by initialisation, which we take as indicator of global recovery rates with replacement strategies, 
Figure~\ref{synth_fig2}. Distinguishing between the different types, we can now see that incorporating the corruption information gives the highest benefits when the corruption is most unevenly distributed over the signal coordinates. So, for the evenly distributed random erasures and burst errors, 'type00' and 'type5', the low-rank component is still recovered by both the unadapted and the adapted algorithm, but as soon as there is intercoordinate variance in the corruption level, type20/22/30' and 'type7', the unadapted algorithm starts to lag behind. For the dictionary recovery, the adapted algorithm already shows advantages for the homogeneous corruption distributions, 'type00' and 'type5', which again become more and more pronounced with increasing intercoordinate variance of the corruption, 'type20/22/30' and 'type7'. \\
The second experiment explores the sensitivity of the algorithms to the flatness/spikyness of the representation pairs, measured by $\|\lratom_\ell\|_\infty$ and $\|\atom_k\|_\infty$. This is done by looking at the recovery of representation pairs, which form orthonormal bases and whose atoms have their energy concentrated on supports of size $m$ for $m=4,8,16, 32, 64, 128, 256$.\\
{\bf Dictionaries \& low-rank components:} For a given support size $m$ we choose $d$ vectors $z_k$ from the unit sphere in $\R^m$ and $d$ supports $I_k={i_1\ldots i_m}$ of size $m$ uniformly at random and set $B(I_k,k)=z_k$ and zero else. We then calculate the closest orthonormal basis to $B$ using the singular value decomposition. The first two elements of this orthonormal basis are chosen as the low-rank component, while the remaining elements form the dictionary. \\
{\bf Signals, corruptions \& setup:} For the signal generation we use the same parameters as in the last experiment and for the corruption we use the random erasure masks of 'type22' with $p_1=q_1=0.7/0.5$ and $p_1=q_1=0.7/0.9$ corresponding to $36\%$ and $64\%$ of corruption. The experimental set up for the recovery of each representation pair is again as in the last experiment. 

\begin{figure}[htbp]
\begin{tabular}{cc}
%%%epsfigures - first time - slow
% \includegraphics[width=8cm]{plots/supp_spike.eps}& \includegraphics[width=8cm]{plots/supp_lr.eps}\\ 
%   (a)& (b)\\
% \includegraphics[width=8cm]{plots/supp_close.eps} & \includegraphics[width=8cm]{plots/supp_rand.eps} \\
% (c)&(d)\\
%%% pdf figures
 \includegraphics[width=8cm]{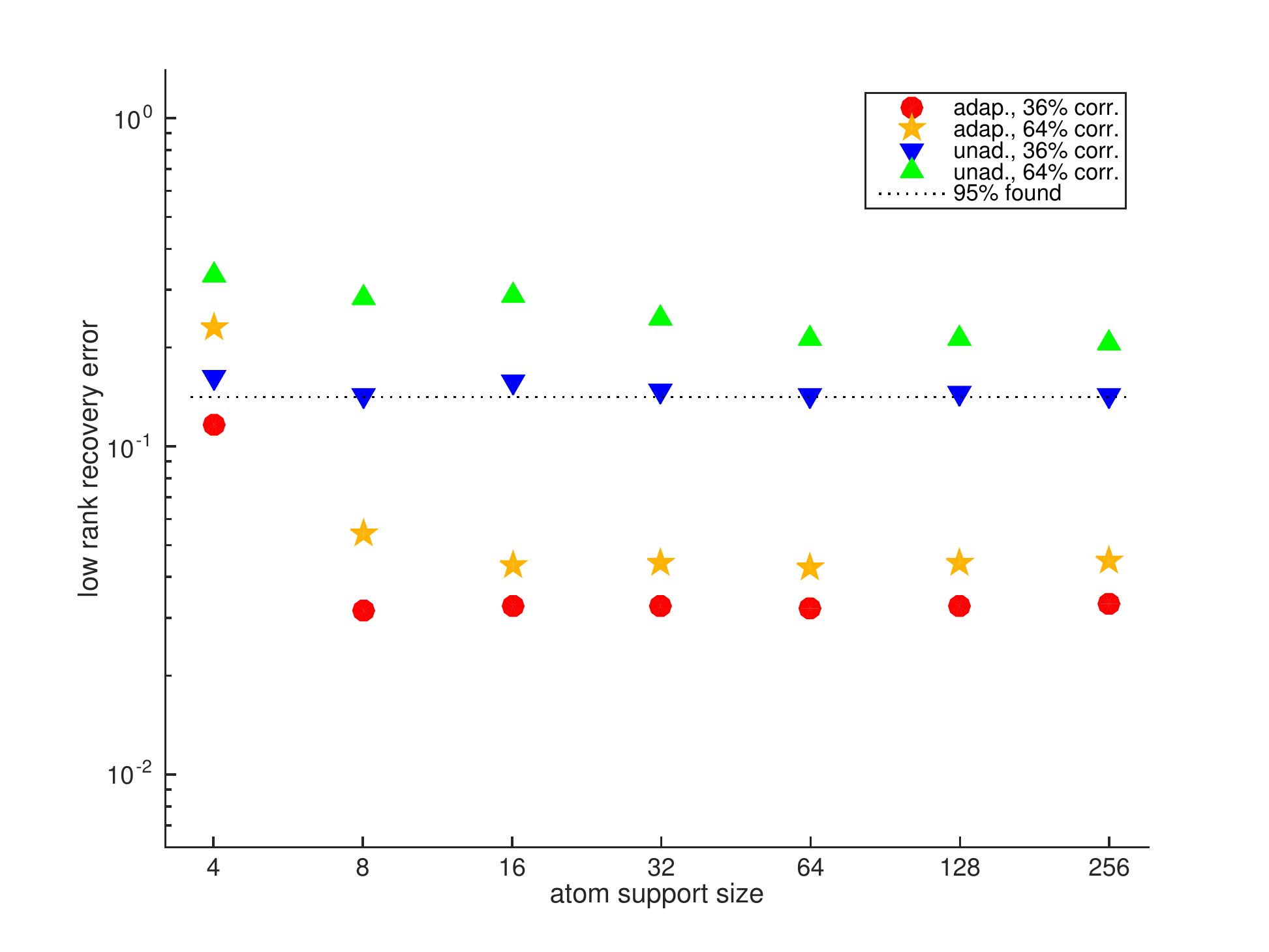}& \includegraphics[width=8cm]{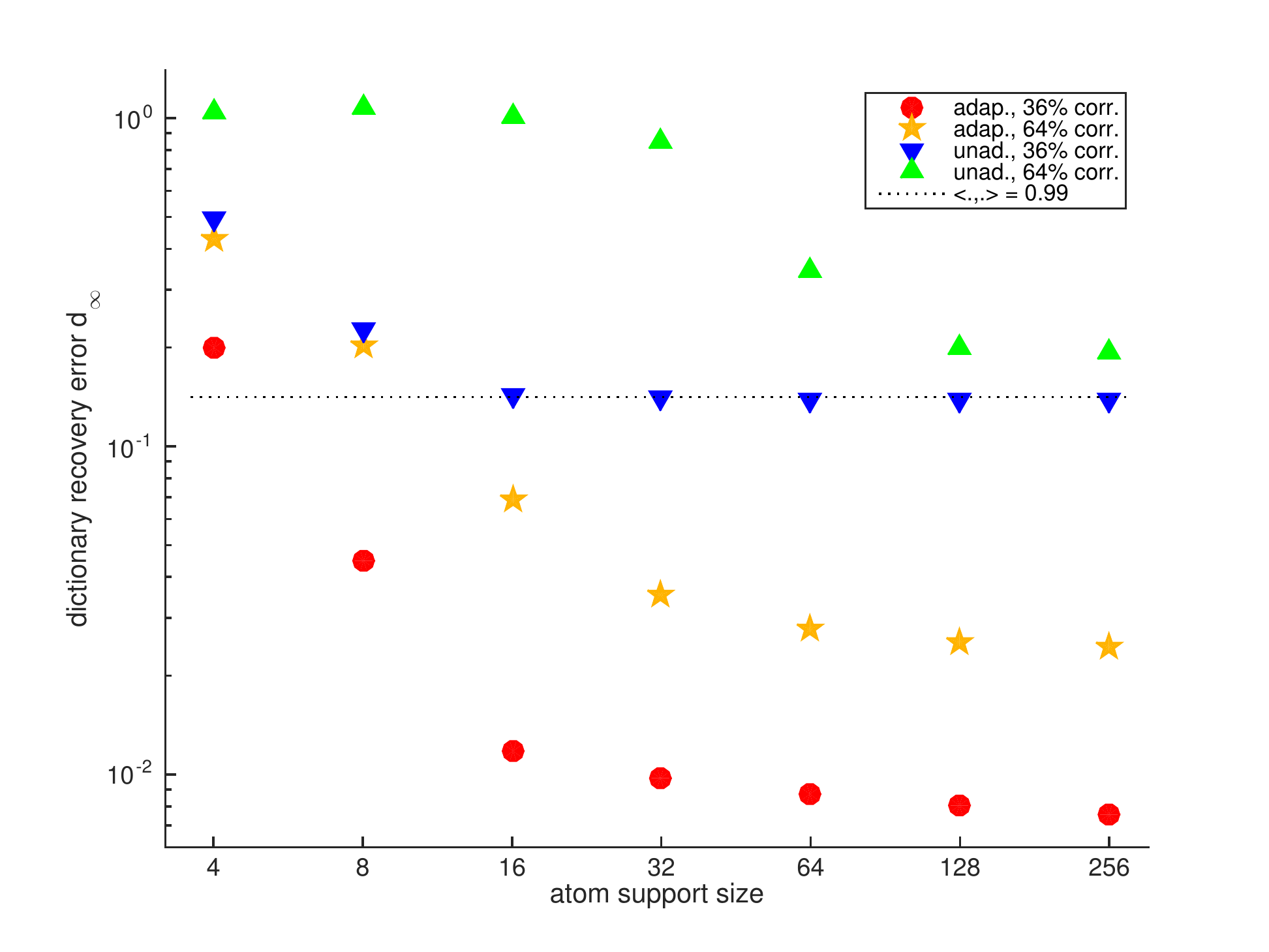}\\ 
   (a)& (b)\\
 \includegraphics[width=8cm]{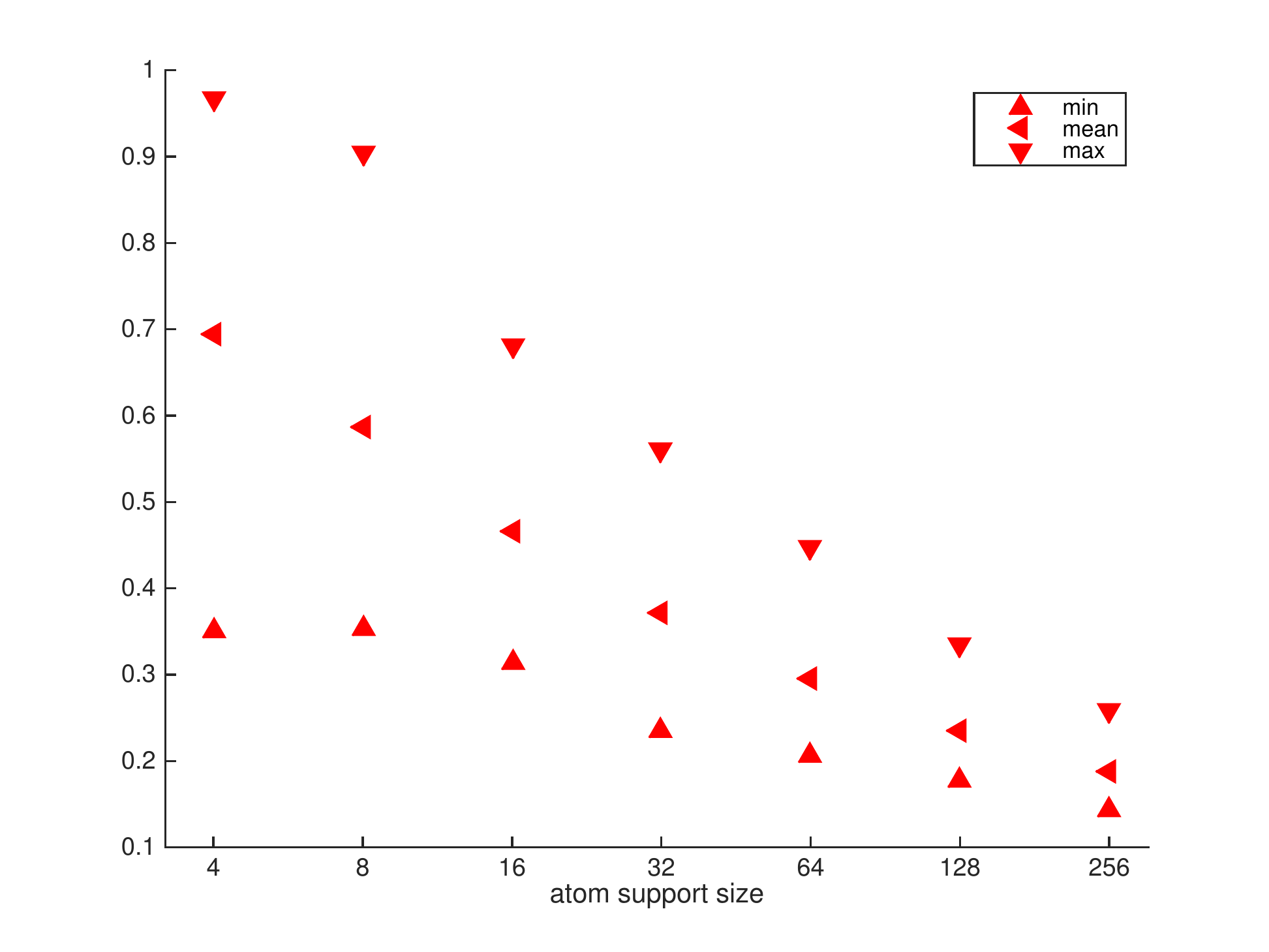} & \includegraphics[width=8cm]{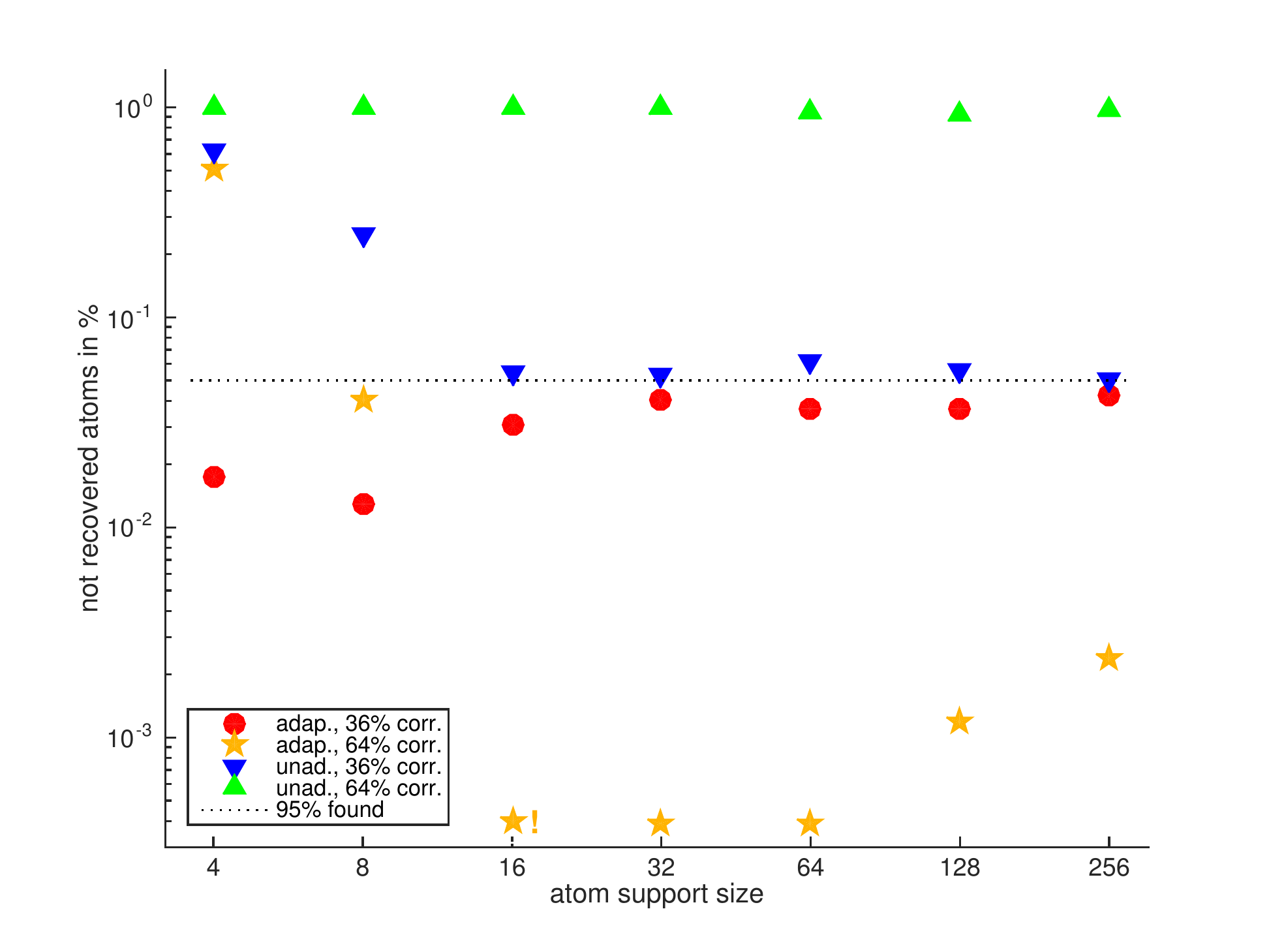} \\
 (c)&(d)\\
  \end{tabular}
    \caption{Recovery performance of the corruption adapted versus the unadapted learning algorithms for random representation pairs with varying atom support sizes and two corruption levels. \label{synth_fig3}}
\end{figure}

Figure~\ref{synth_fig3} shows the spikyness of the representation pairs for various support sizes as well as the corresponding recovery results for the two corruption types. Let us first point out that our construction based on decreasing atom support sizes indeed leads to representation pairs with increased spikyness (Figure~\ref{synth_fig3}c). As usual the recovery errors incurred by the modified algorithms are much lower than those of the unmodified ones. For the low-rank component (Figure~\ref{synth_fig3}a) the recovery error is very stable and only starts to deteriorate for $m=4$, when the low-rank atom carrying less energy is indeed almost a spike, $\|\lratom_2\|_\infty = 0.8997$, meaning $80\%$ of its energy are concentrated on one coordinate. Also for the dictionary recovery the robustness to spikyness of the adapted algorithm is quite surprising. So for the 1:1 initialisation the maximal atom recovery error stays below the critical error until $m=8$ for the low corruption level (36\%) and until $m=16$ for the higher corruption level (64\%). Even more interesting are the recovery rates for the random initialisation, which only break down and go below $94\%$ for $m=4$ at the higher corruption level. As in the last experiment, we observe the effect that spikyness like corruption can lead to better global recovery rates. The effect is more pronounced for the higher corruption level (64\%), where for $m=16$ we even have 100\% recovery. \\
We briefly investigated the effect of the signal scaling on the recovery rates of the modified algorithms for the DCT representation pair and the 'type22' erasure mask with 36\% corruption, with a same setup as in the first experiment, but found that there was no strong influence. That is, for $\scale_{m}$ varying between $2$ and $128$, the low-rank recovery error varies between $0.031$ and $0.036$, the atom recovery error from the close-by initialisation varies between $0.007$ and $0.012$ and the recovery rates from the random initialisation stay between 95\% and 96\%. \\
Similarly, exploring the effect of the sparsity level $S$, we do not gain much more insights over the experiments already conducted in the uncorrupted case, \cite{sc15}. So, fixing all mask and signal parameters except for the sparsity parameter $S$, which increases from $4$ to $16$, the low-rank recovery error stays constant, the dictionary recovery error from the close-by initialisation increases while the number of not recovered dictionary atoms from a random initialisation decreases.\\
In order not to overload the paper, we do not detail these experiments here but refer the interested reader to the ITKrMM MATLAB toolbox\footnote{available at \url{http://homepage.uibk.ac.at/~c7021041/code/ITKrMM.zip}}, which can be used to reproduce all the presented experiments and many more.
Instead we now turn to the application of dictionary learning from corrupted data to image inpainting.

%%%%%%%%%%%%%
%!TEX root = maskdl.tex
%%%%%%%%%%%%%

%%%%%%%%%%%
\section{Application: image inpainting}\label{sec:applications}
\begin{figure}[tbh]
\begin{tabular}{cc}
\includegraphics[width=9cm]{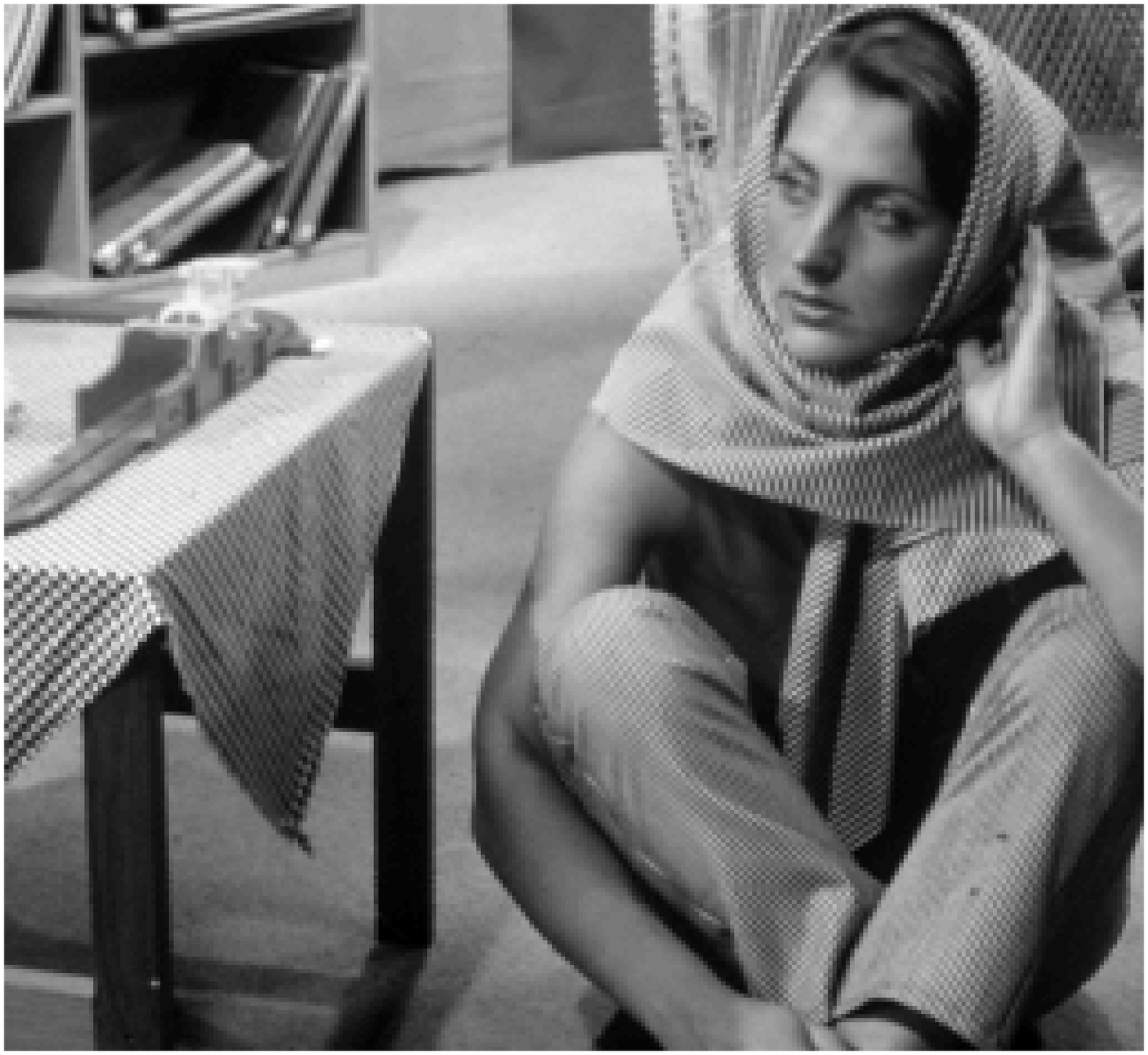}& \includegraphics[width=9cm]{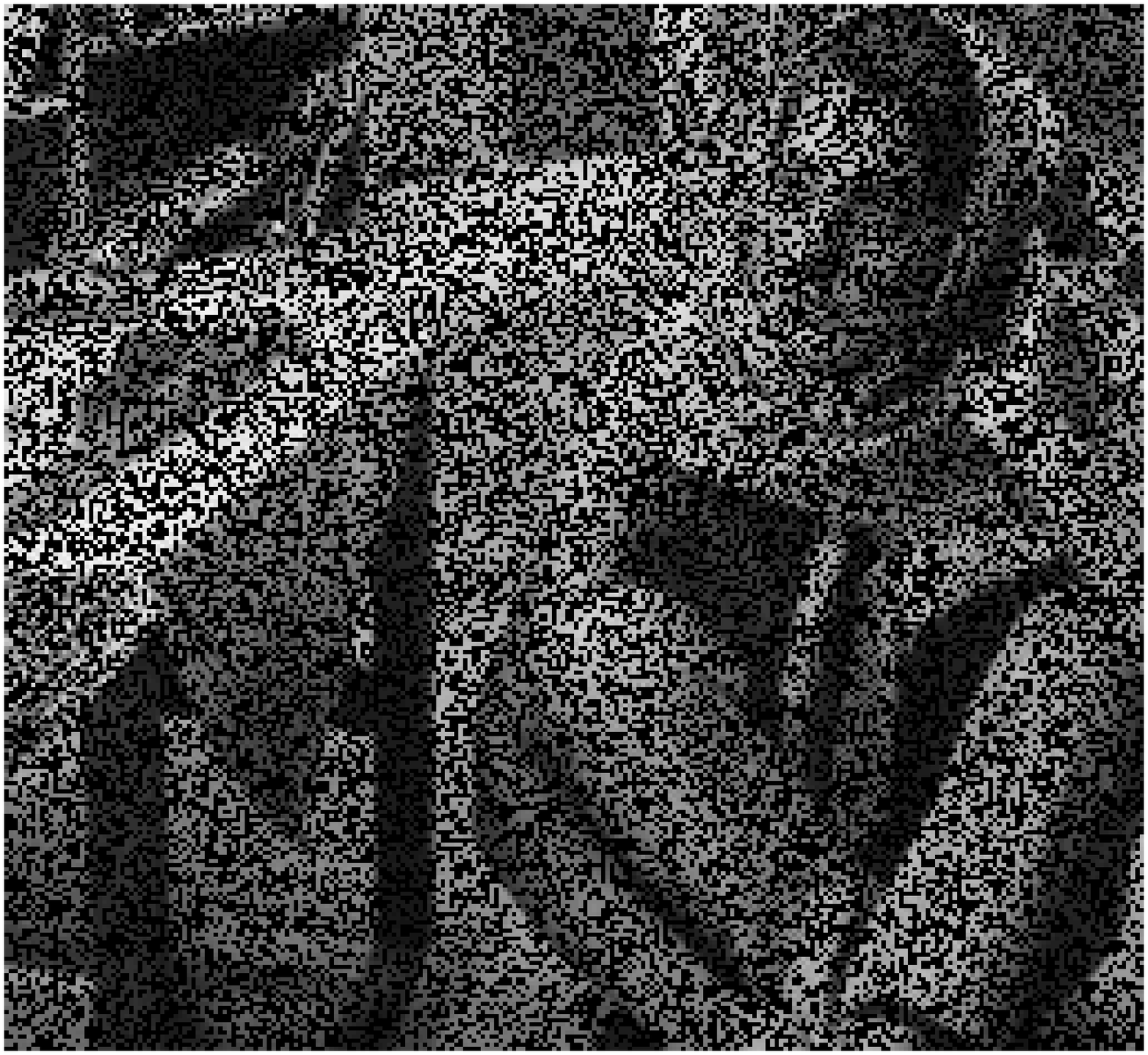}\\
(a) original image & (b) image with 50\% erasures\\
\includegraphics[width=9cm]{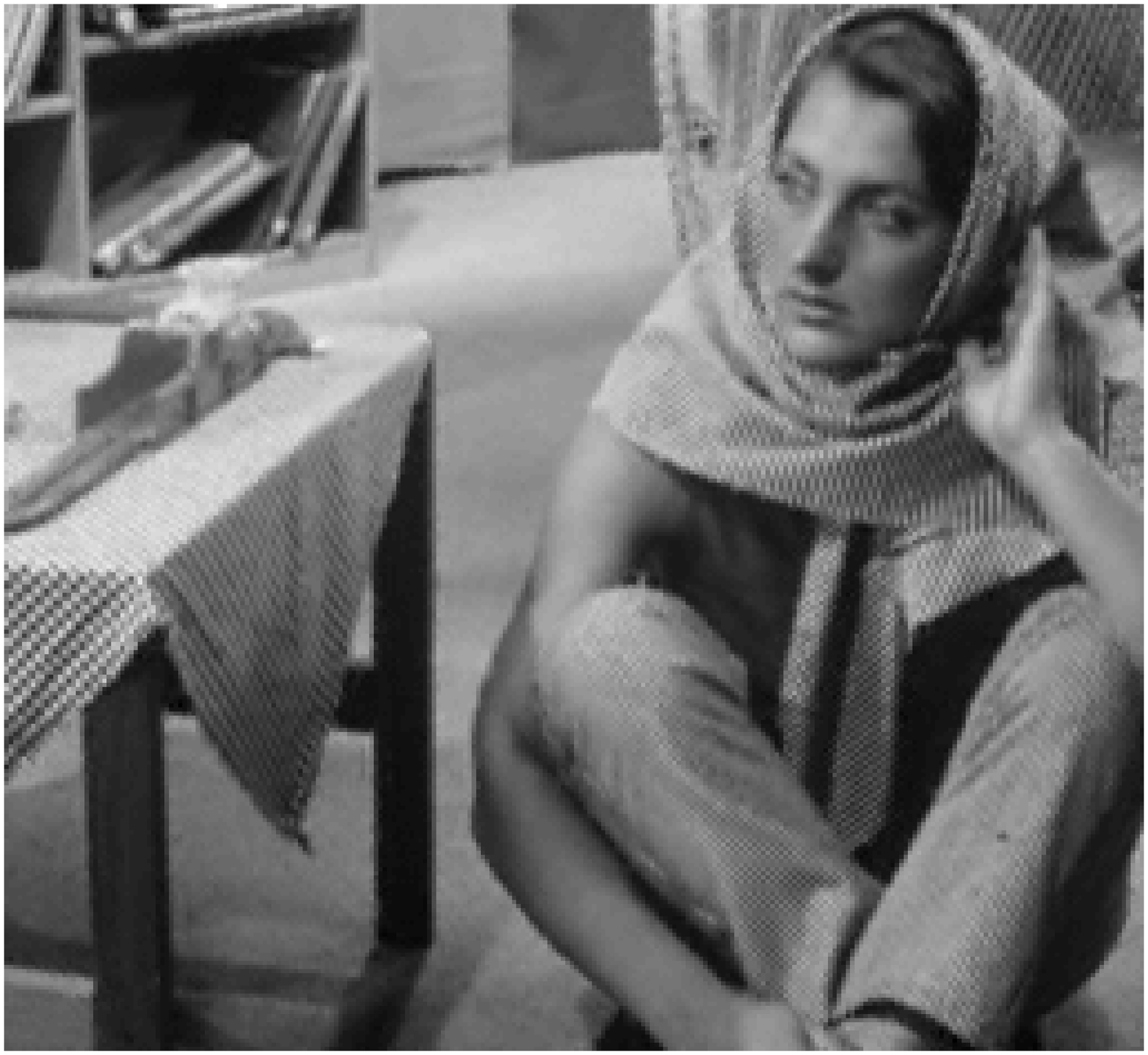} &\includegraphics[width=9cm]{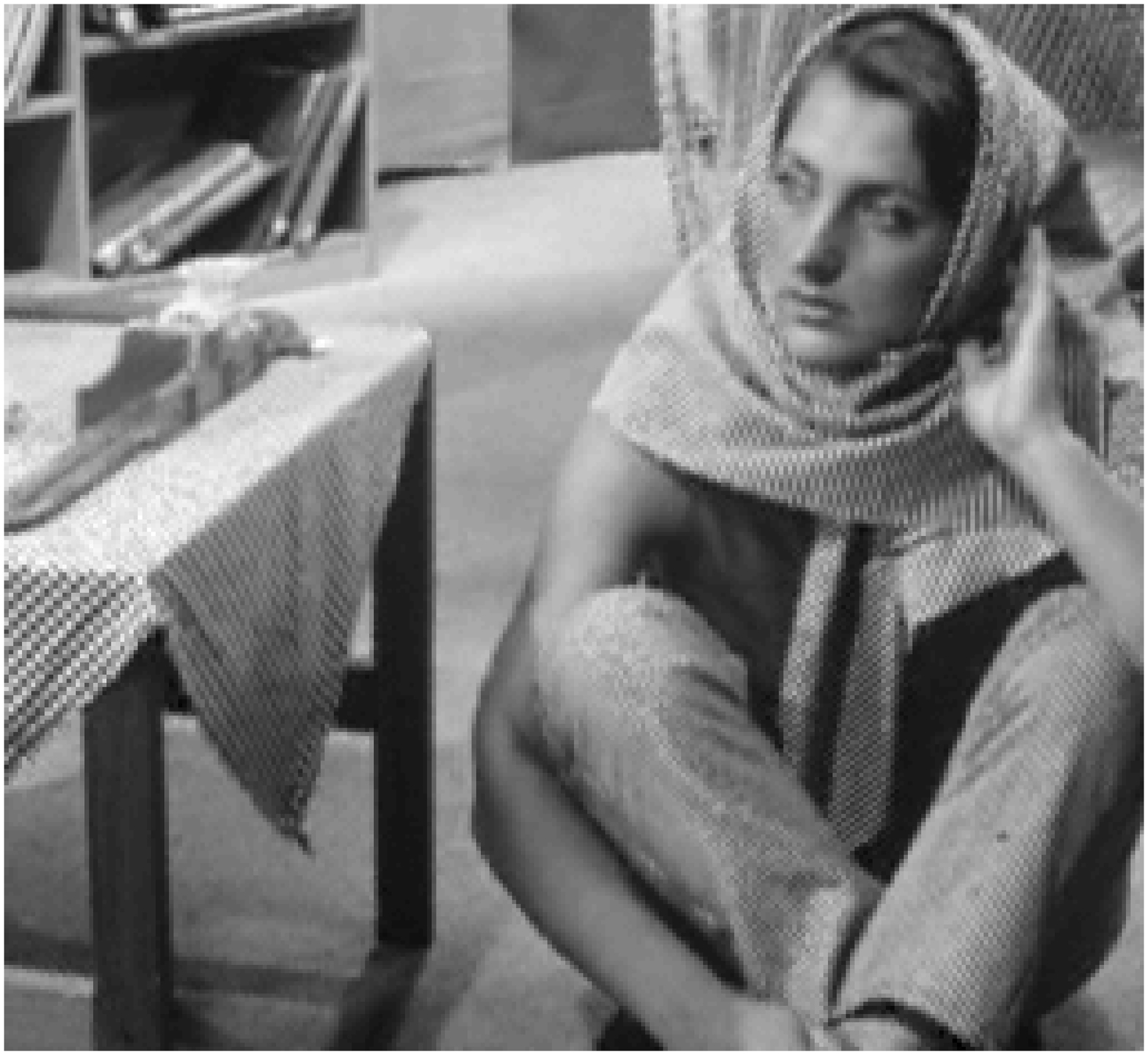} \\
(c) reconstruction with ITKrMM ($L=1$) & (d) reconstruction with wKSVD
\end{tabular}
\caption{Inpainting example: Barbara \label{fig:inp_rand5}}
\end{figure}

\begin{figure}[tbh]
\begin{tabular}{cc}
\includegraphics[width=9cm]{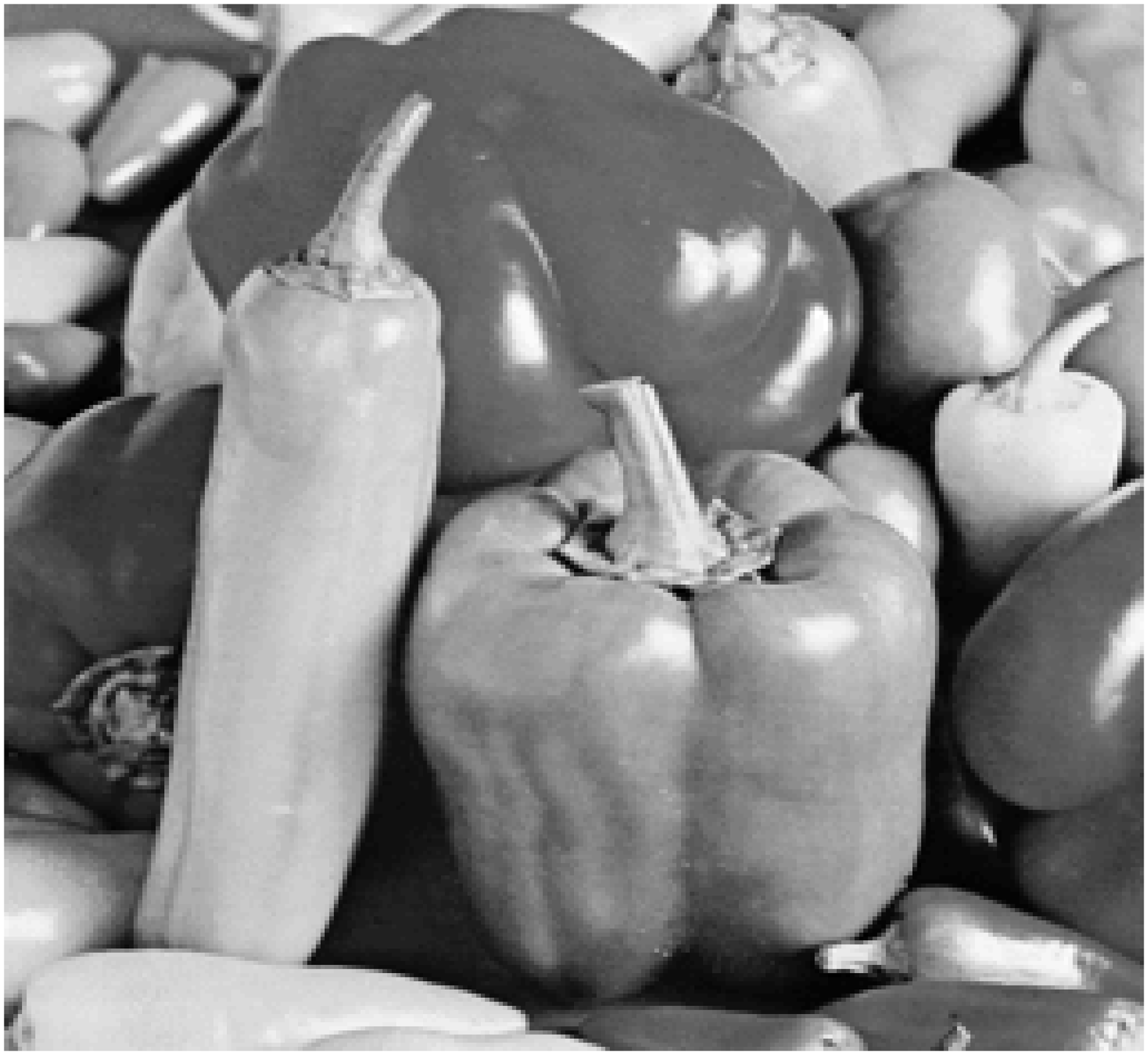}& \includegraphics[width=9cm]{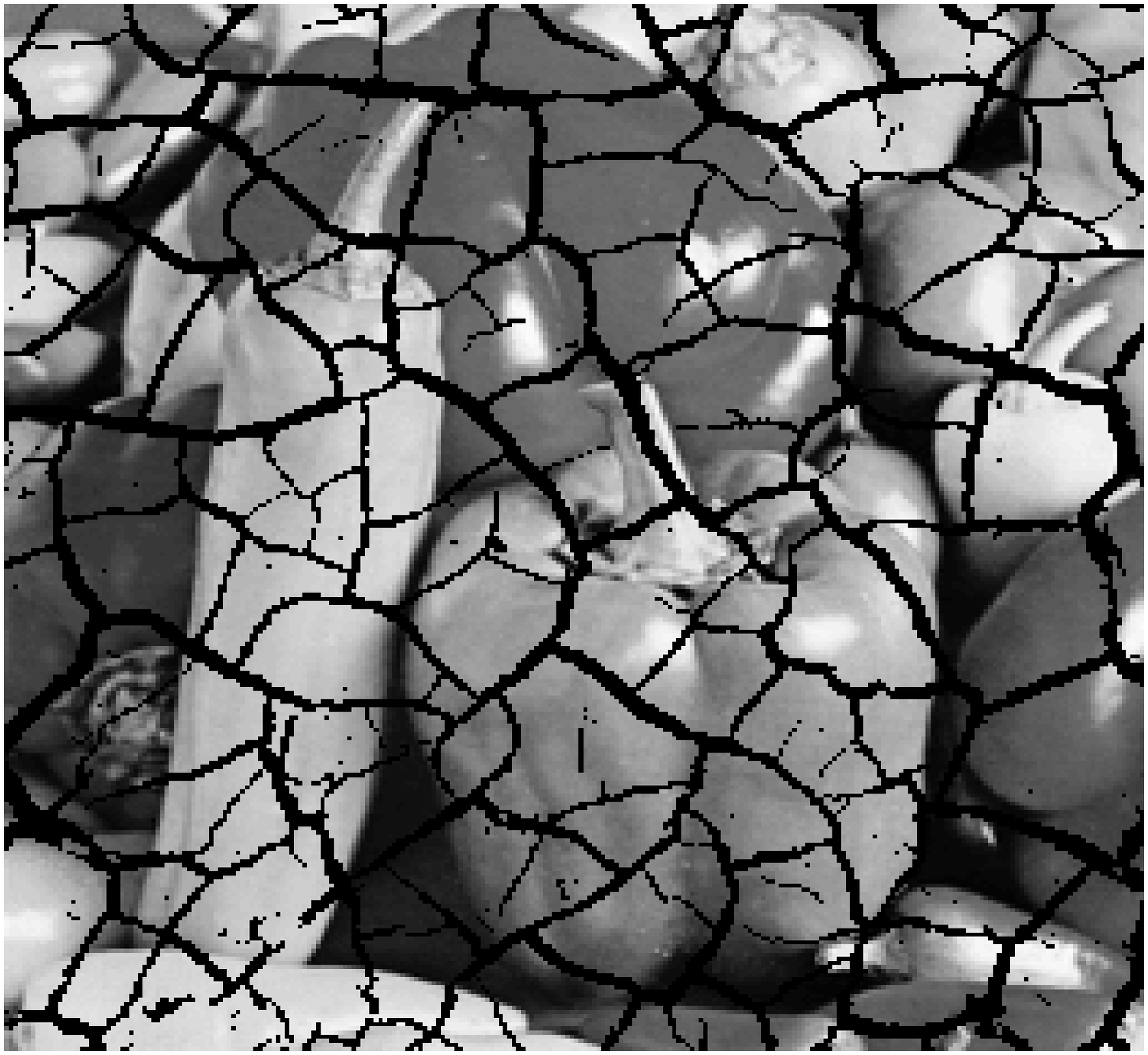}\\
(a) original image & (b) image with cracks \\
\includegraphics[width=9cm]{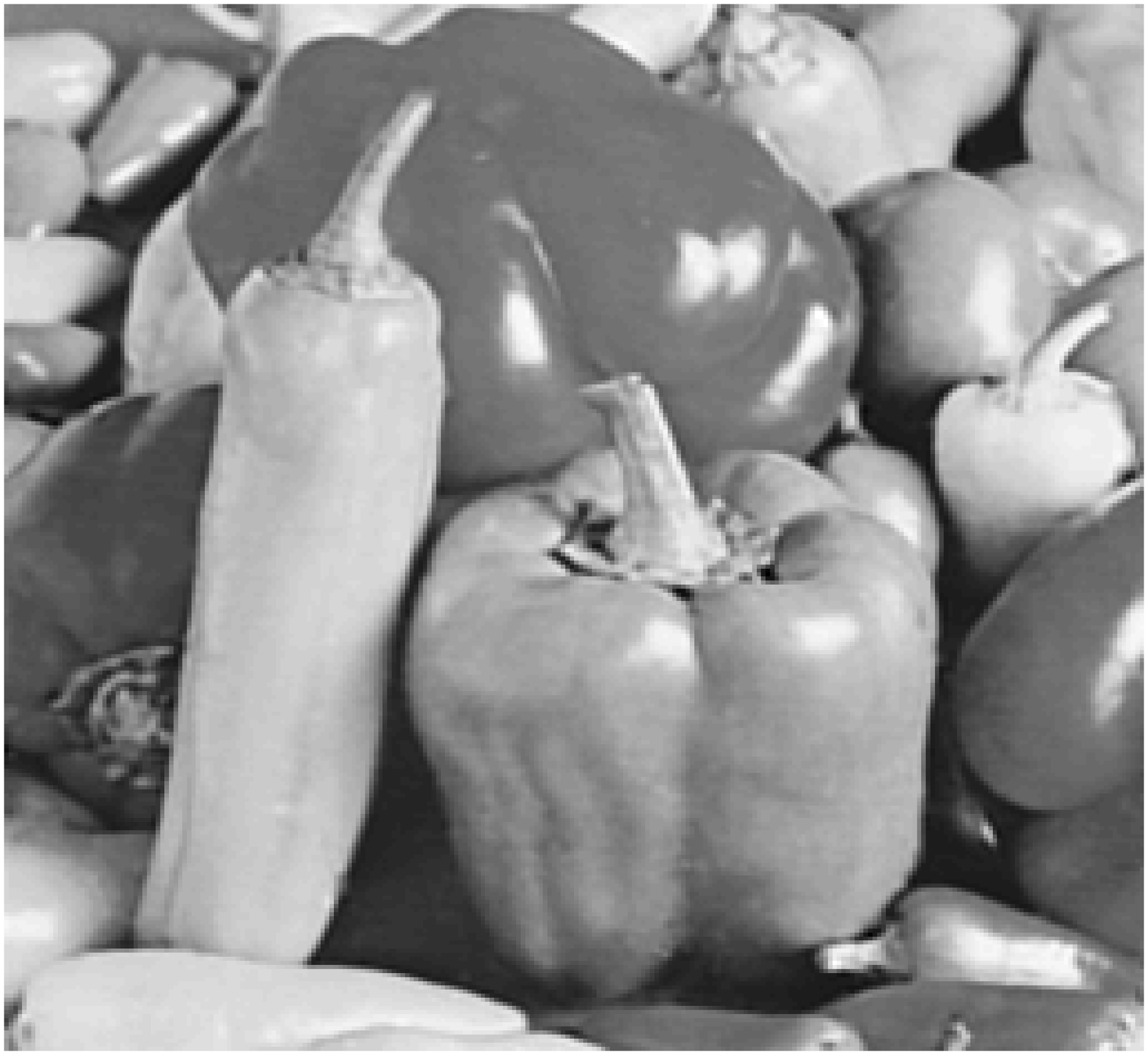} &\includegraphics[width=9cm]{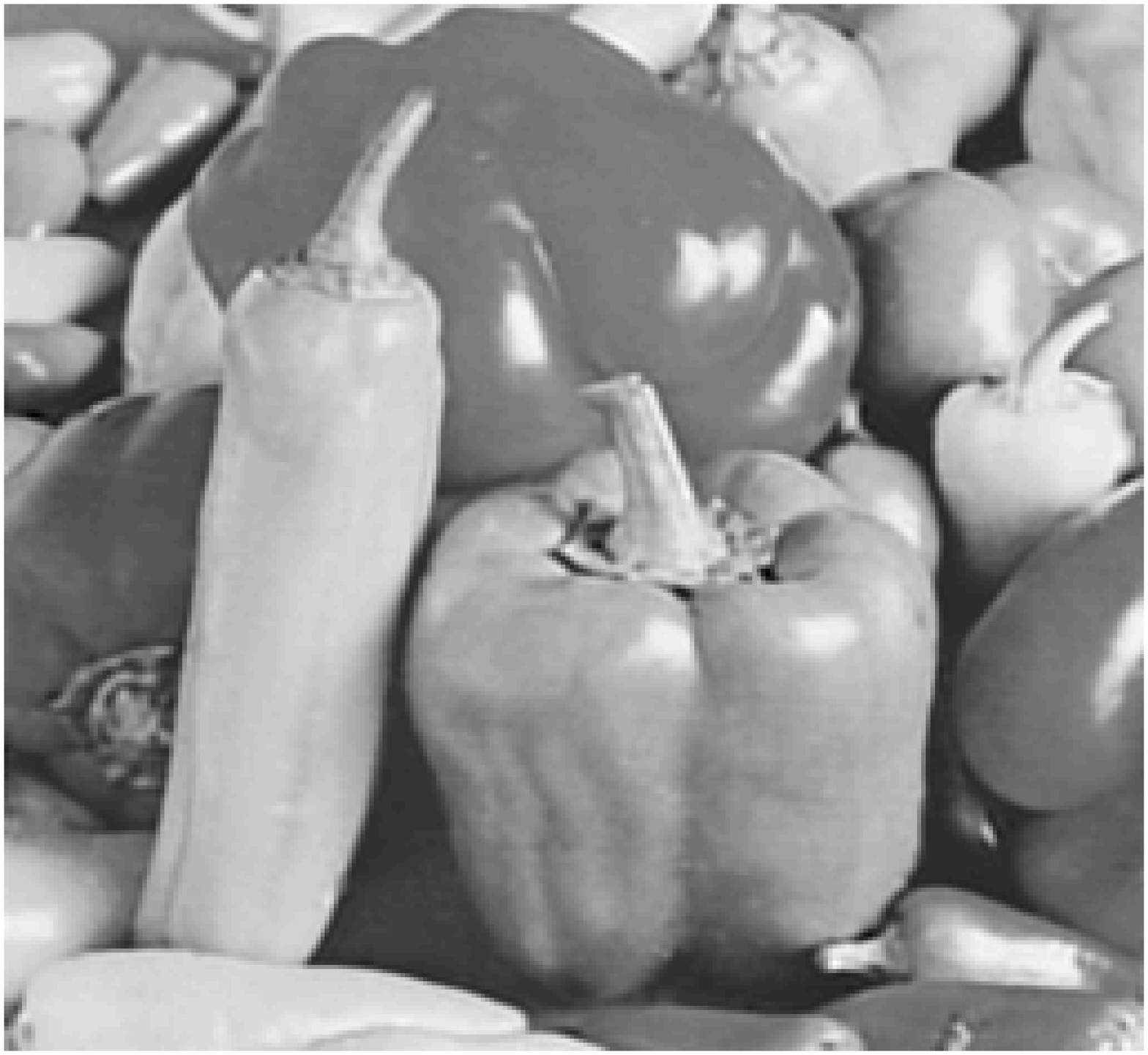} \\
(c) reconstruction with ITKrMM ($L=1$) & (d) reconstruction with wKSVD
 \end{tabular}
\caption{Inpainting example: Peppers \label{fig:inp_cracks}}
\end{figure}

%%%%%%%%%%%
To demonstrate the practical value of the ITKrMM algorithm, we here conduct an image inpainting experiment. Inpainting is the process of filling in missing information or holes in damaged signals, and our motivating task, the prediction of blood glucose levels, can be cast as inpainting problem.
Image inpainting, in particular, is used for restoration of old analog paintings, denoising of digital photos, and for removal of objects like text or date stamps from images and has become an active field of research in the mathematical and engineering communities, with a variety of specifically developed methods and approaches. Since our goal is to evaluate the ITKrMM algorithm as a robust and simple tool for dictionary learning from incomplete data, we do not compare our inpainting scheme with state-of-the-art methods for image inpainting in general, but instead compare the performance of various learned dictionaries for dictionary based inpainting. \\
Apart from comparing to the dictionaries and low-rank components of different size learned by ITKrMM, we also compare to dictionaries learned by weighted KSVD (wKSVD), a modification of KSVD, \cite{ahelbr06}, for signals affected by non-homogeneous noise or erasures. wKSVD is the most likely algorithm to fit into our setup and has already been successfully used for image inpainting, \cite{maelsa08, masael08}. 
Let us now describe our setup for dictionary based inpainting.\\
{\bf Data:} For our experiments we consider grayscale images of size $256 \times 256$ from a standard image database. In the first set of experiments, the images are corrupted by erasing each pixel iid with probability $0.3, 0.5$ or $0.7$, resulting in 30, 50 or 70\% erased pixels on average. In the second set of experiments, part of the pixels from the images are removed using a structured mask, see Figure~\ref{fig:inp_cracks}b.
We then extract all possible patches of size $p \times p$ pixels from the corrupted image as well as the corresponding mask. We set $p=8$ in case of the randomly erased pixels and $p=12$ in case of the structured mask. The vectorised corrupted patch/mask pairs are then given to the dictionary learning algorithm.\\
{\bf Dictionary \& low-rank component:} Via ITKrMM we learn dictionaries of size $K = 2d-L$ atoms, where $d=p^2$ is the dimension of the patches and $L$ denotes the size of the low-rank component. We consider $L=1$ and $L=3$. Via wKSVD we learn a dictionary of size $K=2d$ with the option of keeping the first atom always equal to the constant atom $\atom_1 \equiv c$. This corresponds to ITKrMM with $K=2d-1$ and $L=1$. The ratio 2 between patch dimension and the size of the dictionary together with the low-rank component makes the dictionary redundant and allows to capture different features, while still allowing for reasonable computational complexity. \\
{\bf Initialisation \& sparsity level}: We use the same initialisation strategies as for the synthetic experiments, ie. random vectors that are orthogonal to  the low-rank component resp. atoms that have already been learned. This means that in case $L=1$, before subtracting the low-rank component, the initial dictionaries for ITKrMM and wKSVD are the same. As sparsity level for the learning step of ITKrMM, we use $S = p-L$ to keep the comparison between different sizes of the low-rank component fair. Since within wKSVD the contribution of the constant low-rank atom counts in the sparse approximation step, we use as sparsity level $S=p$. For learning the low-rank components we use $10$ iterations per low-rank atom and all available patch/mask pairs and for learning the dictionary we use 40 iterations on all available patch/mask pairs
for both algorithms.
Note that the wKSVD version adapted to erasures was originally designed to adapt an existing patch dictionary, pre-learned on some image data base, to the image patches at hand rather than to calculate the dictionary from scratch. Still, abusing it to do just that, it works very well. \\
{\bf Dictionary based inpainting:} Dictionary based inpainting relies on the concept that every image patch $y$ is sparse in a flat patch-dictionary $\dico$ and therefore every damaged/masked patch $My$ is sparse in the damaged/masked patch dictionary $M \dico$, that is for if for $|I|\leq S$
\begin{align}
 y \approx \dico_I x_I  \quad \Rightarrow \quad My \approx M\dico_I x_I.
\end{align}
To reconstruct the original patch we simply need to recover coefficients $\tilde{x}_I \approx x_I$ by sparsely approximating $My$ in $M\dico$ and to set $\tilde y = \dico \tilde{x}_I$. The flatter the dictionary is, the more stable the atoms are to erasures and the easier it is to recover the correct coefficients.
We first reconstruct every damaged image patch via sparse approximation and then reconstruct the full image by averaging every pixel over all reconstructed patches in which it is contained.
As sparse approximation algorithm we use Orthogonal Matching Pursuit with a sparsity level of $S=20$ in case of the random erasures and $S=30$ in case of the structured mask, \cite{damazh94, parekr93}. Note that since the damaged dictionary is not normalised we need to account for this in the OMP selection step and rescale by $1/\|M\atom_k\|_2$, simarly to thresholding in the ITKrMM algorithm. Without this renormalisation less damaged atoms take precedence over better fitting ones.\footnote{In the original version of wKSVD, which uses OMP as sparse approximation algorithm, the OMP selection step seems not to have included normalisation. However, we find that including the normalisation step both for learning and inpainting  improves the final performance of the wKSVD dictionary, especially for the 70\% corruption level, where it increases the PSNR by up to 2.9dB.}
%%%%%%%%%%%
\begin{algorithm}[OMP masked] 
Given a damaged signal $My$ together with the mask $M$, a dictionary $\dico$ and a sparsity level $S$, initialise
$r=My$, $I=\emptyset$ and while $|I| < S $ do
\begin{itemize}
\item Atom selection: find $ j = \arg\max_{M\atom_k \neq 0} |\ip{r}{M\atom_k}|/\|M\atom_k\|_2$.
\item Approximation: Set $I = I \cup \{j\}$, $x_I = (M\dico_I)^\dagger My$ and $r = My - M\dico_I x_I$.
\end{itemize}
Output $x_I$.
\end{algorithm}
%%%%%%%%%%%
 {\bf Comparison/Error: }
We measure the recovery success of the different pairs of dictionaries and low-rank components by the peak signal-to-noise ratio (PSNR) 
 between the original image $Y$ and the recovered version $\tilde Y$. For $Y, \tilde Y$ both images of size $d_1\times d_2$ the PSNR is defined as
\begin{align}
 \text{PSNR in dB} = \log_{10}\left(\frac{(\max_{i,j}Y(i,j) - \min_{i,j}Y(i,j))^2 }{\frac{1}{d_1 d_2} \sum_{i,j} (Y(i,j) - \tilde Y(i,j))}\right).  
\end{align}
In case of the random erasure mask we average over 5 runs, each with a different mask and initialisation, to account for the variability of the PSNR for different mask realisations. The results of the experiment are presented in Table~\ref{table:psnr} and examples of two images, one corrupted with the 50\% mask and the other with the structured mask, together with their reconstructions and the original images can be found in Figure~\ref{fig:inp_rand5} and Figure~\ref{fig:inp_cracks}. 
%%%%
%%%%%% table PSNR
\begin{table}[tbh]
\centering
\begin{tabular}{|c|l|C{1.5cm}|C{1.5cm} | C{1.5cm}| C{1.5cm}|C{1.5cm}|C{1.5cm}|}
\toprule
Corruption & Algorithm   & Barbara & Cameram. & House & Mandrill & Peppers & Pirate\\
\midrule
\multirow{4}{*}{rand., 30\% } & {\it Noisy Image}  & {\it 11.17} & {\it 10.81}  &   {\it 10.11} &{\it 10.82}&{\it 11.18}&{\it 11.70}\\
&wKSVD  & 35.64 & 32.56 &  41.42& 30.41& 38.47 & 35.06\\
%&wKSVD (L=2) & 35.90 & 32.65 &  41.49& 30.42 & 38.66 & 35.08\\
%&wKSVD (L=3) & 36.04 & 32.65 &  41.51& 29.77& 38.70 & 35.16\\
&ITKrMM (L=1)  & 36.04 & 32.80 &  41.84 & 30.85& 39.05 & 35.54\\
%&ITKrMM (L=2)   & 36.59 & 32.99 &  42.03& 30.90 & 39.33 & 35.76 \\
&ITKrMM (L=3)   & {\bf 37.05} & {\bf 33.07} & {\bf 42.32} & {\bf 30.91} & {\bf 39.67} & {\bf 36.04} \\
\midrule
\multirow{4}{*}{rand., 50\% } &{\it Noisy Image}  & {\it 8.95} & {\it 8.59} &   {\it 7.88}&   {\it 8.60}& {\it 8.96} & {\it 9.47}\\
&wKSVD    	  & 32.96& {\bf 29.47} &  {\bf 37.99} & 27.69& 35.29 & 31.95\\
%&wKSVD (L=2) & 32.95 & 29.43 &  38.00& 27.70& 35.26 & 31.96\
%&wKSVD (L=3) & 32.94 & 29.46 &  38.06& 27.72& 35.32 & 31.99\\
&ITKrMM (L=1)  & 33.12 & 29.45 &  37.77 & 27.94 & 35.28 & 32.10\\
%&ITKrMM (L=2)   & 33.39 & 29.45&  37.90 & 27.94 & 35.25 & 32.17\\
&ITKrMM (L=3)   & {\bf 33.69} & 29.44&   37.99 & {\bf 27.95} & {\bf 35.33} & {\bf 32.22}\\
\midrule
\multirow{4}{*}{rand., 70\% } &{\it Noisy Image} & {\it 7.48}  & {\it 7.13} & {\it 6.42} & {\it 7.13}& {\it 7.50} & {\it 8.01} \\
%\cline{3-5}
&wKSVD  		 & 28.55 & 25.40 & {\bf 32.93} & 24.72 & {\bf 30.61} & {\bf 28.34}\\
%&wKSVD (L=2) & 28.50 & 25.31&  32.83 & 24.67 & 30.47 & 28.26\\
%&wKSVD (L=3) & 28.47 & 25.21 &  32.65 & 24.64 & 37.74 & 28.17\\
&ITKrMM (L=1)  &{\bf 28.91} & {\bf 25.47} & 32.39 & {\bf 24.76} & 30.37 & 28.30\\
%&ITKrMM (L=2)  & 28.80 & 25.23 & 32.19 & 24.73 & 30.03 & 27.96\\
&ITKrMM (L=3)  & 28.79 & 25.25 & 32.05 & 24.74 & 29.78 & 27.89\\
\midrule
\multirow{4}{*}{cracks, 19\% } &{\it Noisy Image} & {\it 13.07}  & {\it 12.75}  & {\it 12.01} & {\it 12.85}  & {\it 13.20}& {\it 13.67}\\
&wKSVD  		& 33.74 & {\bf 30.54} & {\bf 38.36} & 28.63 & {\bf 34.09} & 32.39\\
%&wKSVD (L=2) & 33.71 & 30.33 &  38.09& 28.67& 33.66 & 32.27\\
%&wKSVD (L=3) & 33.63 & 30.43 &  38.04& 28.65& 33.50 & 32.21\\
&ITKrMM (L=1) & 33.31 & 30.35 & 37.90 & 29.02 &  33.72 & 32.34\\
%&ITKrMM (L=2) & 33.54 & 30.33 & 37.98 & 29.14 & 32.99 & 32.34\\
&ITKrMM (L=3) & {\bf 33.86 }& 30.36 & 37.74 & {\bf 29.19} & 32.93 & {\bf 32.47}\\
\bottomrule
\end{tabular}
\caption{Comparison of the PSNR (in dB) for inpainting of images with various corruption levels based on dictionaries learned with wKSVD and ITKrMM on all available corrupted image patches. \label{table:psnr}}
\end{table}
%%%
%%% figure inpainting cracks random erasures
%%%
%%% figure inpainting cracks
%%%
Table~\ref{table:psnr} shows that when randomly erasing around $30\%$ of the pixels, inpainting with the ITKrMM dictionaries always outperforms inpainting with the wKSVD dictionaries. In case of $50\%$ and $70\%$ random erasures as well as for the structured mask, the wKSVD and ITKrMM dictionaries perform about equally, where for the more textured images (Barbara, Mandrill, Pirate) ITKrMM tends to have a slight advantage, while for the smooth images (Cameraman, House, Peppers) wKSVD is slightly better. Another trend we can observe is that for the lower corruption levels ITKrMM with a low-rank component of size $L=3$ improves over a low-rank component of size $L=1$, while for the higher corruption levels it is the other way round. We also conducted experiments with ITKrMM for $L=2$. The performance was for all settings inbetween the case $L=1$ and the case $L=3$, so we do not include them here. \\
%
% where for $70\%$ erasures ITKrMM with  in the same slightly better for 4 out of 6 images. In case of the structured mask this trend is again reversed and the ITKrMM dictionaries perform slightly better than the wKSVD dictionaries. It is also interesting to observe that for ITKrMM in the case of the $30\%$ and $50\%$ random erasures, choosing a low-rank component of size $L=3$ improves upon the low-rank component of size $L=1$, while for the 70\% erasure mask it is the other way round. For the structured mask there is no conclusive trend, but $L=1$ seems to be the safer option. We also conducted experiments with ITKrMM and $L=2$ but since the performance was for all settings between the case $L=1$ and the case $L=3$, we do not include them here. \\
%We further want to mention, as shown by the two examples of inpainted images, that inpainting with the ITKrMM dictionaries produces slightly less blurry images than inpainting with the wKSVD dictionaries - not only in cases where ITKrMM inpainting obtains a much higher PSNR value but also when it gives a similar or slightly lower PSNR value. \\
%%%%
\begin{figure}[tbh]
\begin{tabular}{cc}
\includegraphics[width=8cm]{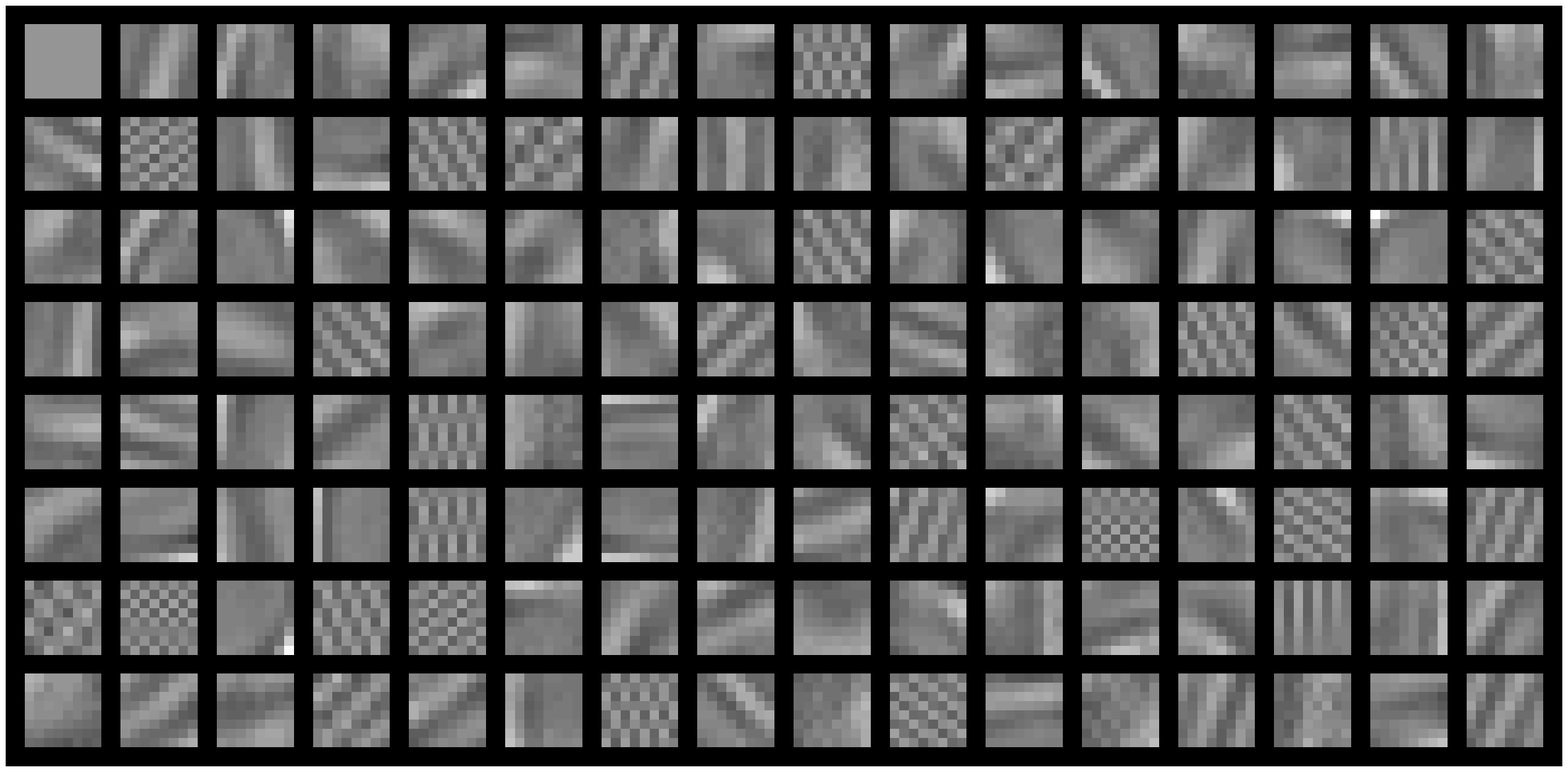} &\includegraphics[width=8cm]{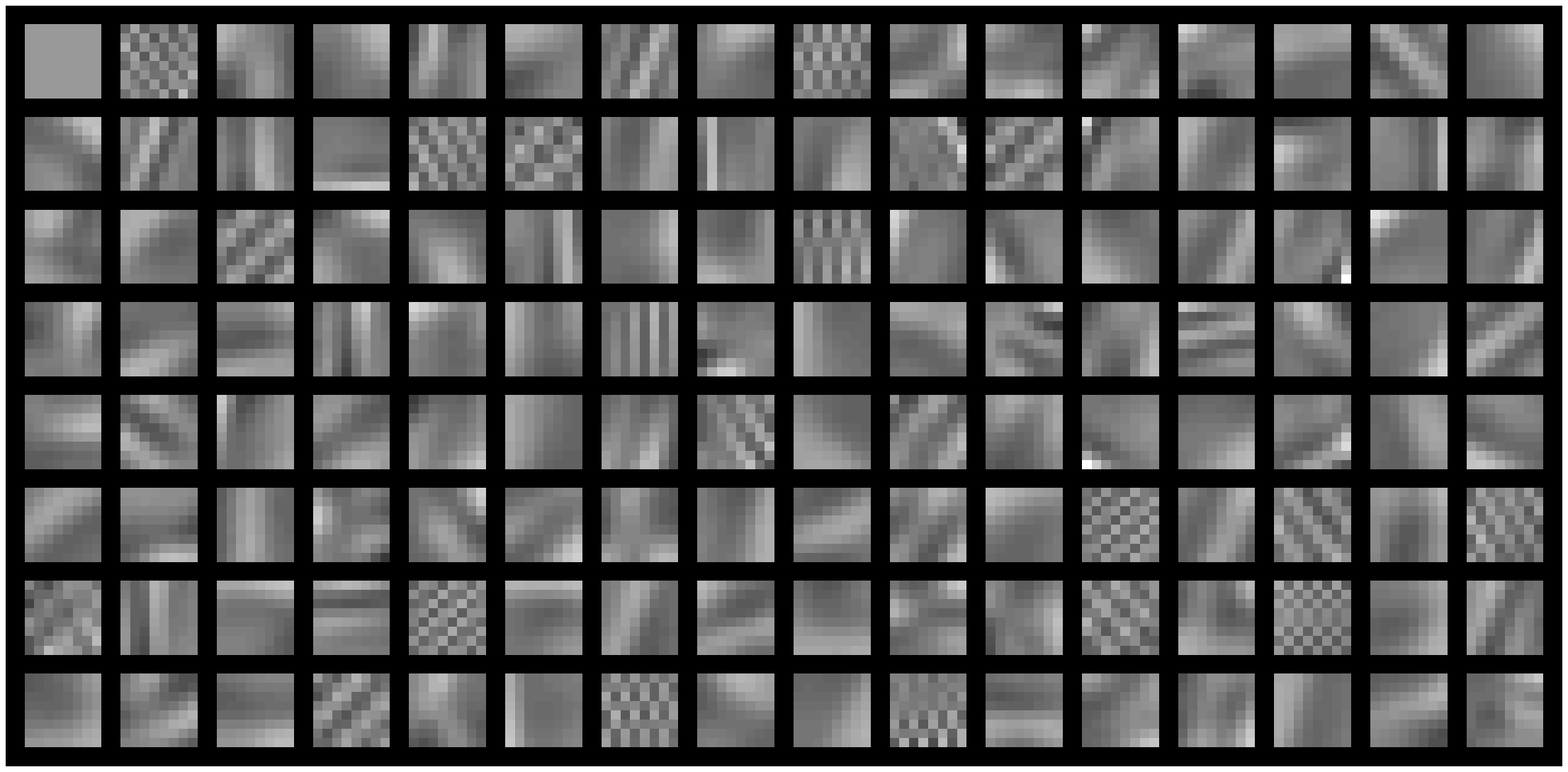} \\
(a) ITKrMM, 30\% erasures & (b) wKSVD, 30\% erasures  \\
\includegraphics[width=8cm]{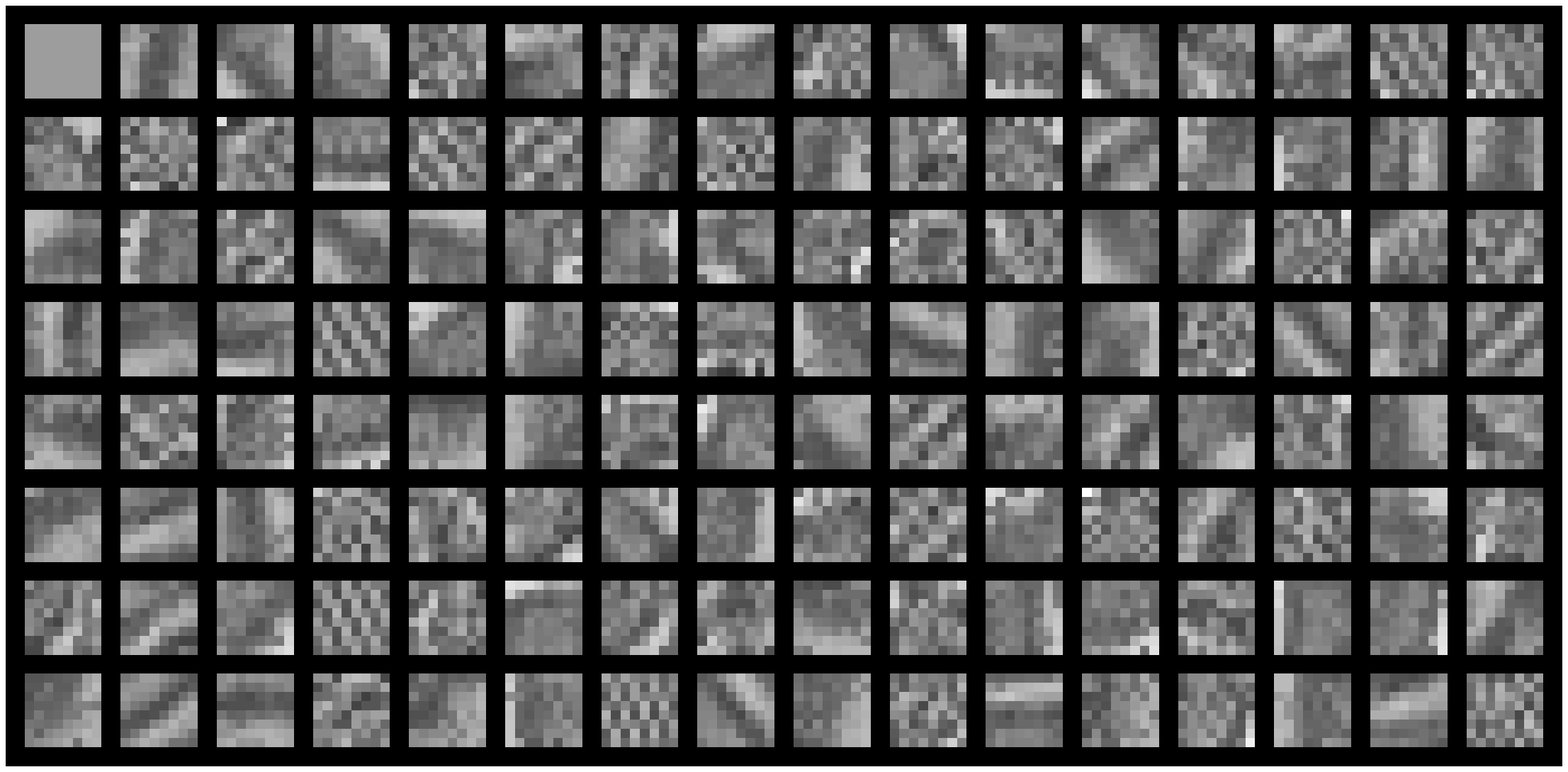} &\includegraphics[width=8cm]{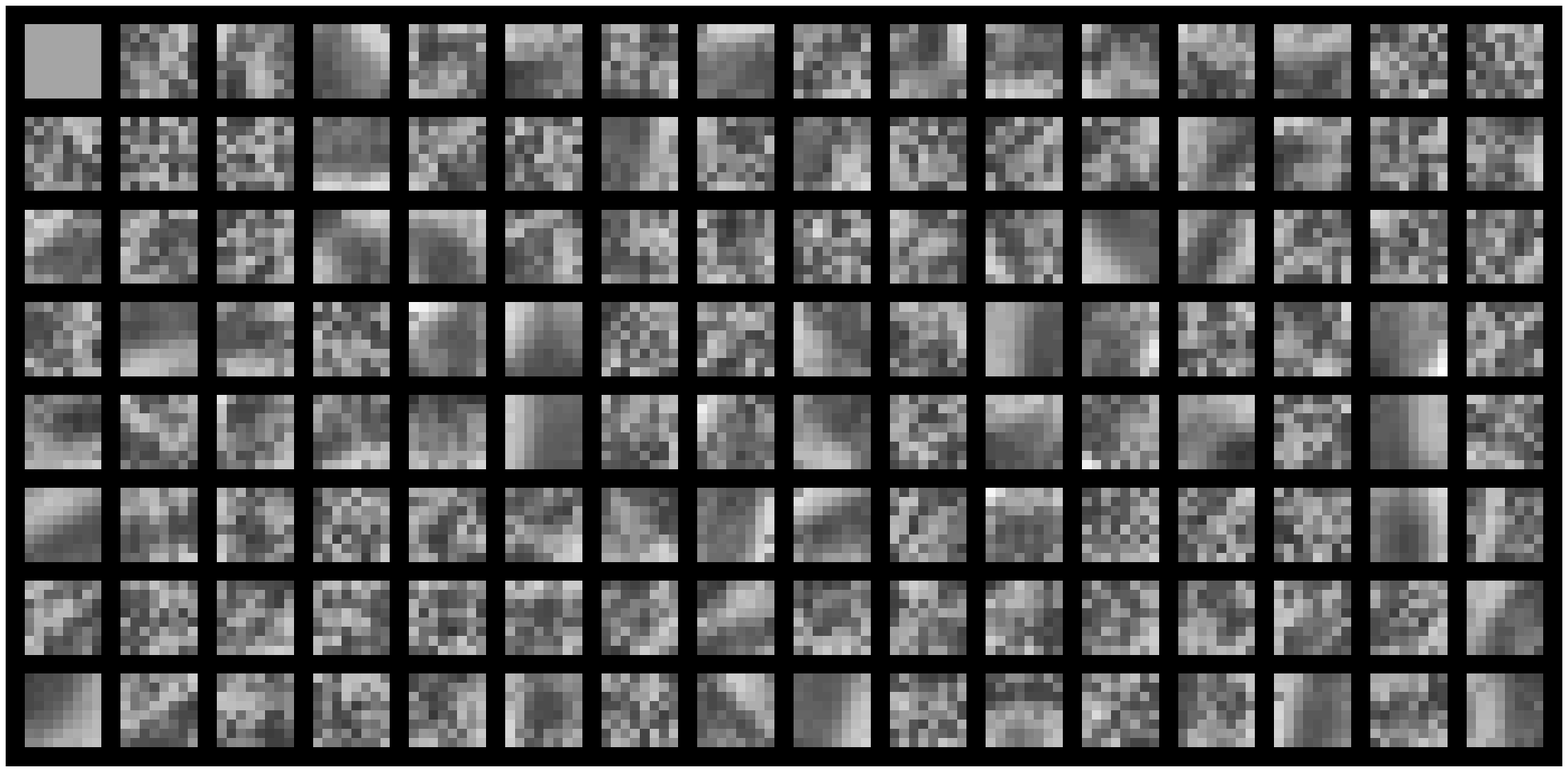} \\
(c) ITKrMM, 70\% erasures & (d)  wKSVD, 70\% erasures
\end{tabular}
\caption{Dictionaries and low-rank atom (left upper corner) learned with ITKrMM and wKSVD on all $8 \times 8$ patches of Barbara corrupted with 30\% resp. 70\% erasure rate. \label{fig:dicos_barb}}
\end{figure}
%%%%%
\begin{figure}[tbh]
\begin{tabular}{cc}
\includegraphics[width=8cm]{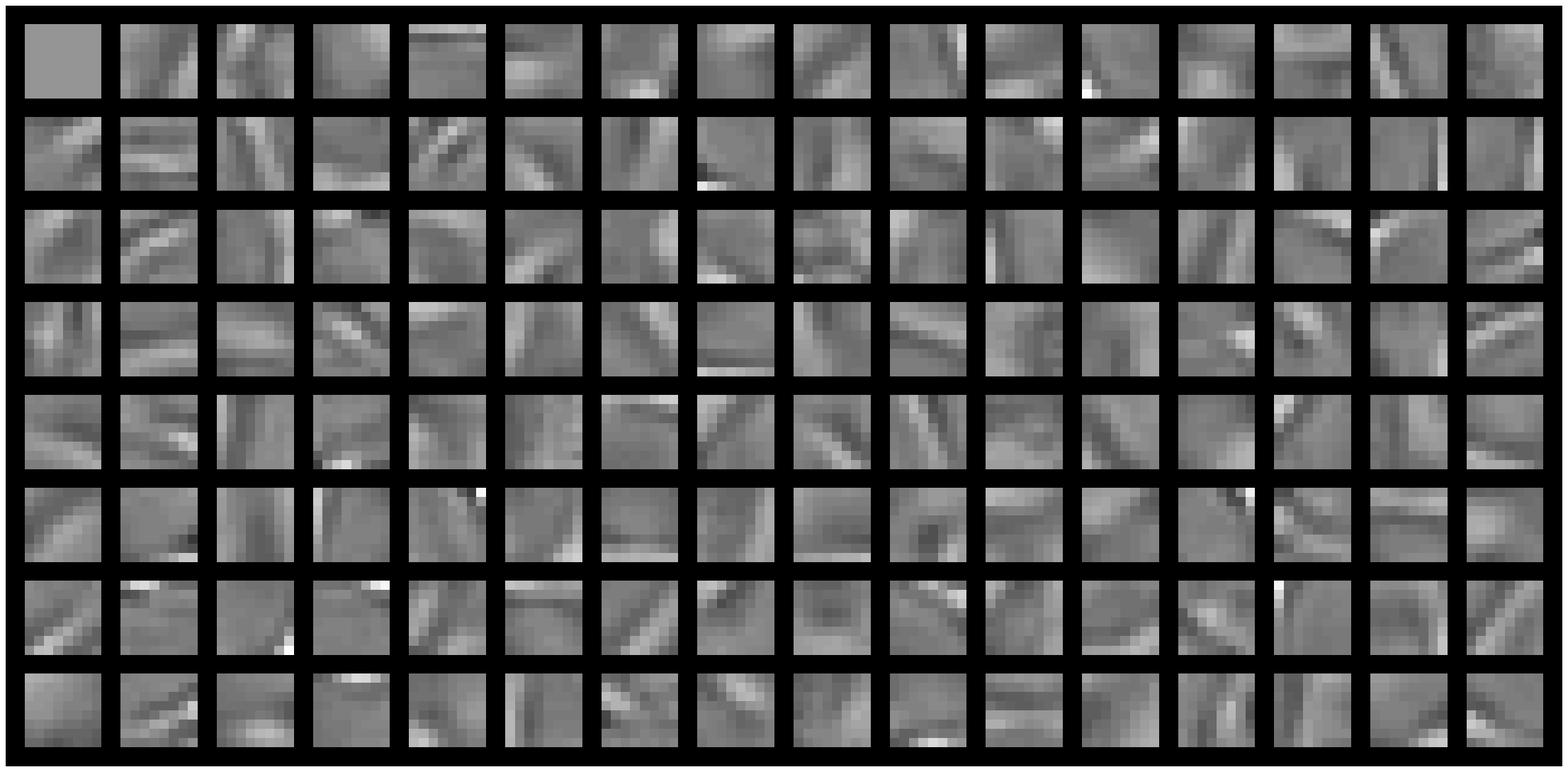} &\includegraphics[width=8cm]{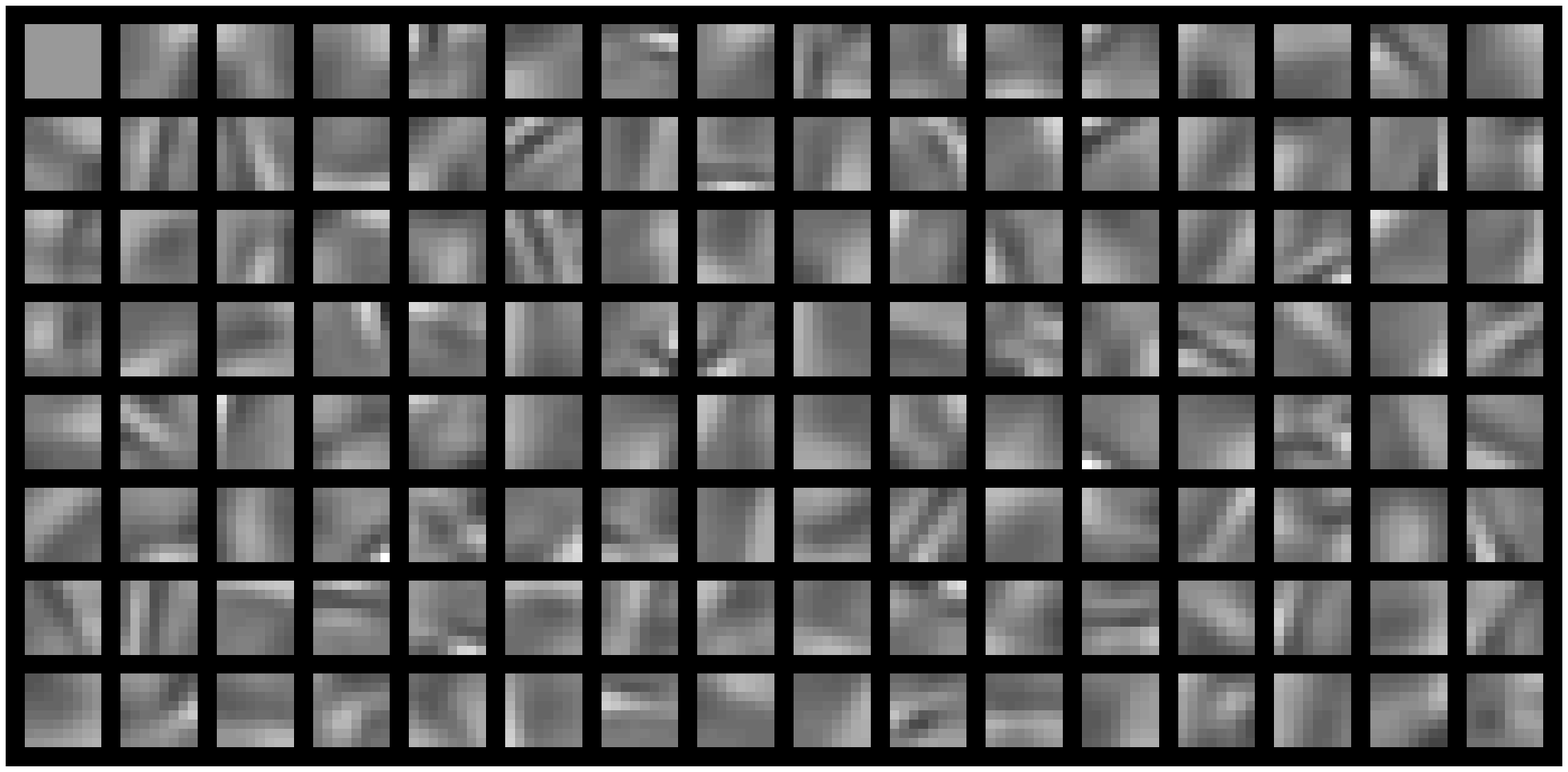} \\
(a) ITKrMM, 30\% erasures & (b) wKSVD, 30\% erasures  \\
\includegraphics[width=8cm]{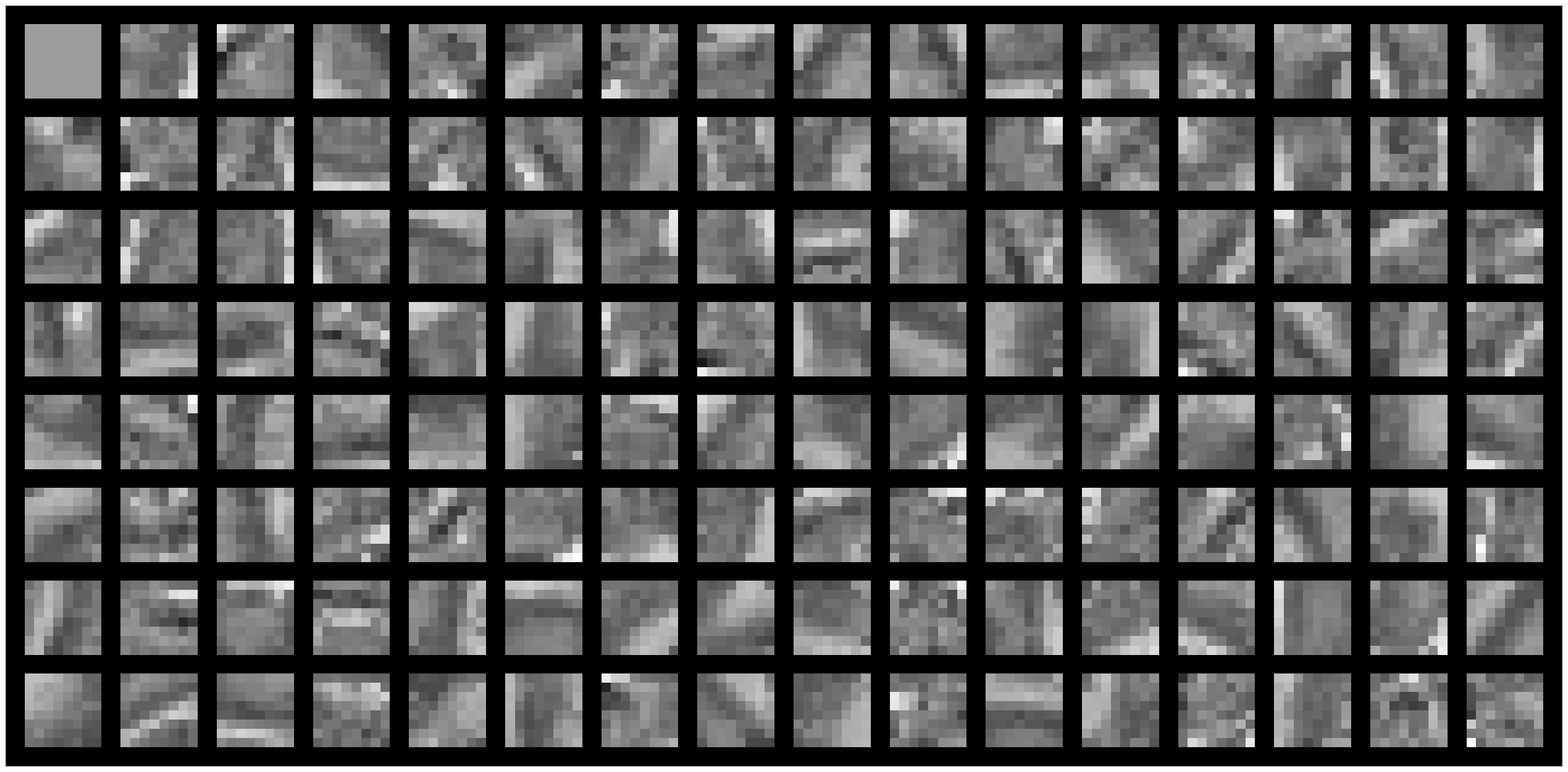} &\includegraphics[width=8cm]{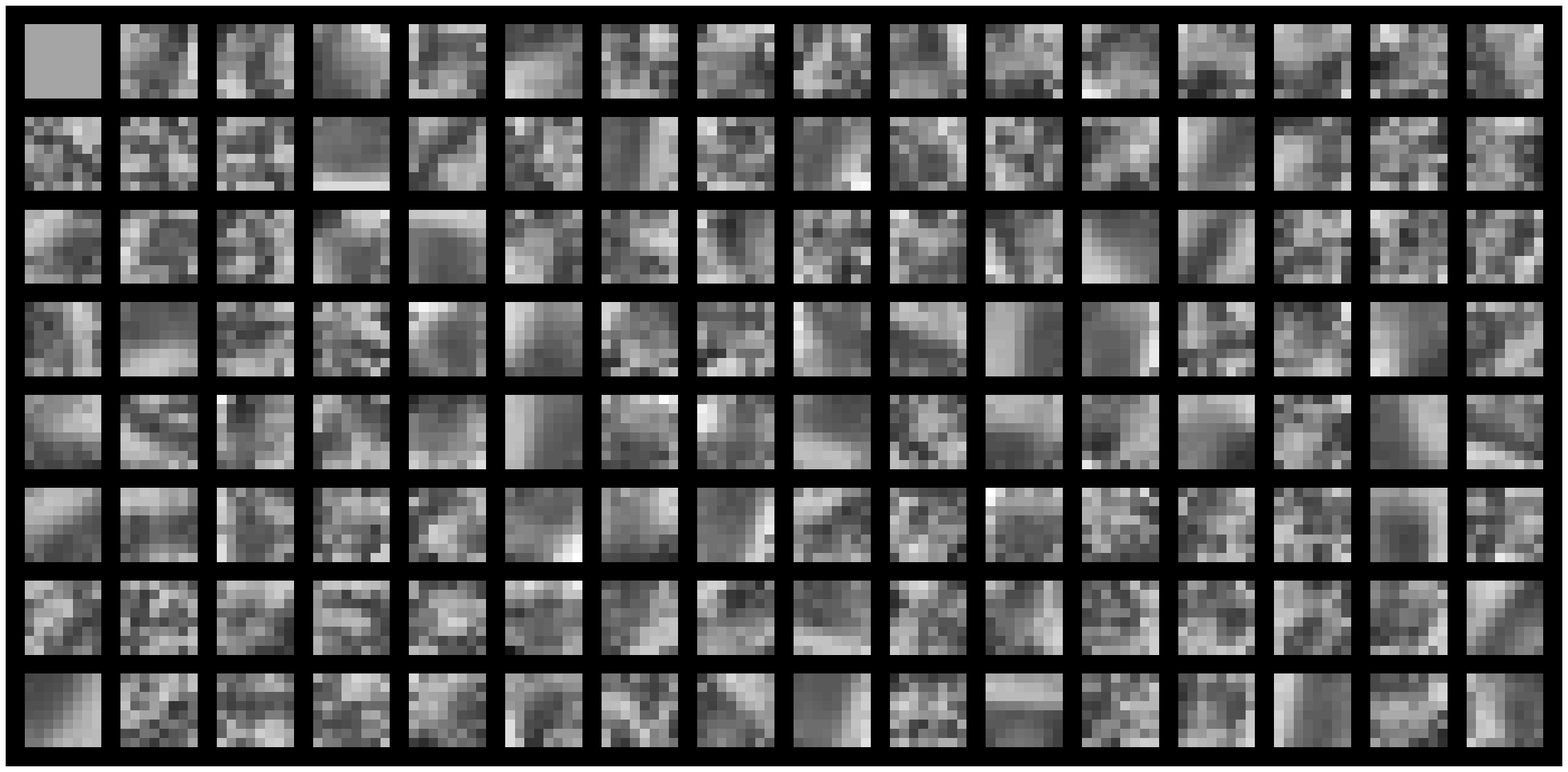} \\
(c) ITKrMM, 70\% erasures & (d)  wKSVD, 70\% erasures
\end{tabular}
\caption{Dictionaries and low-rank atom (left upper corner) learned with ITKrMM and wKSVD on all $8 \times 8$ patches of Peppers corrupted with 30\% resp. 70\% erasure rate.\label{fig:dicos_pepp}}
\end{figure}
%%%%
\begin{figure}[tbh]
\begin{tabular}{cc}
\includegraphics[width=8cm]{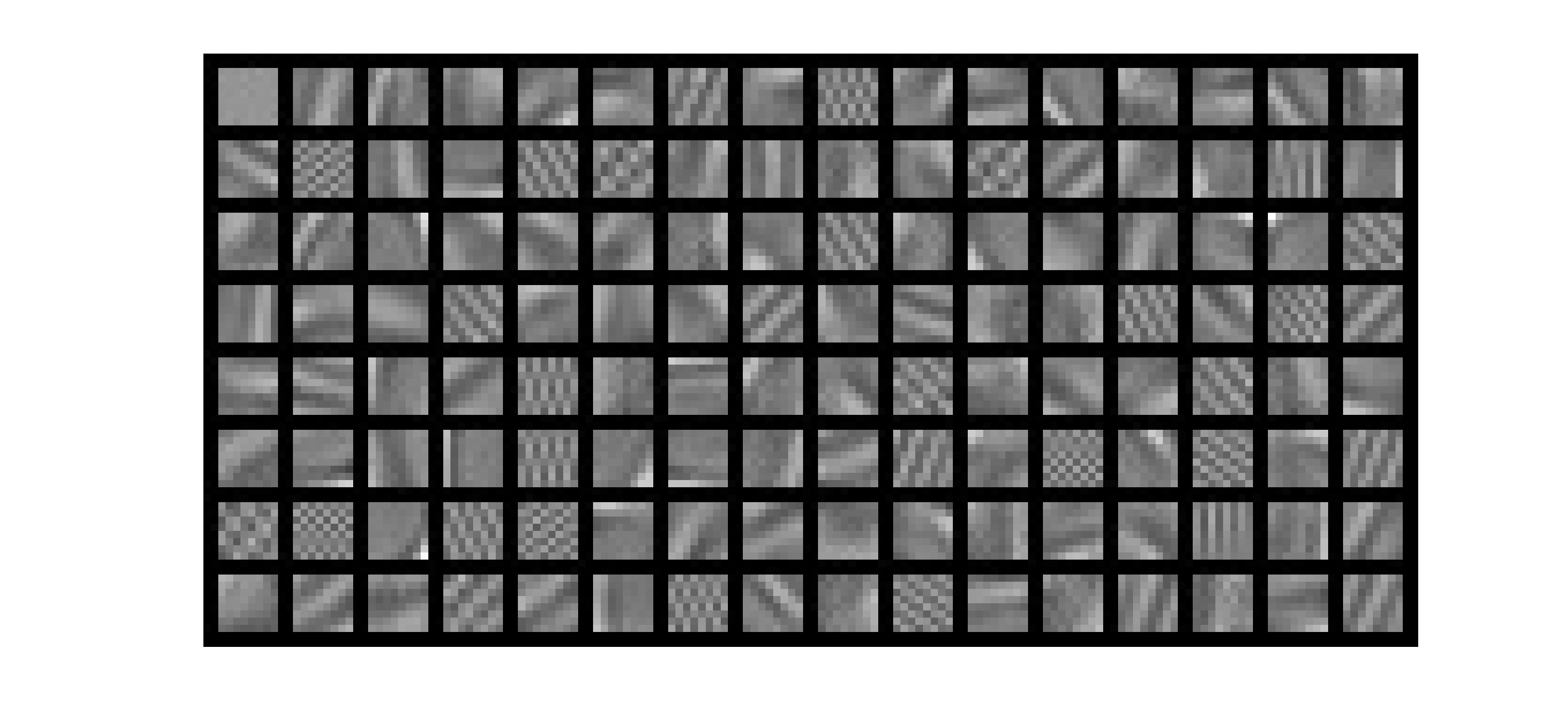} &\includegraphics[width=8cm]{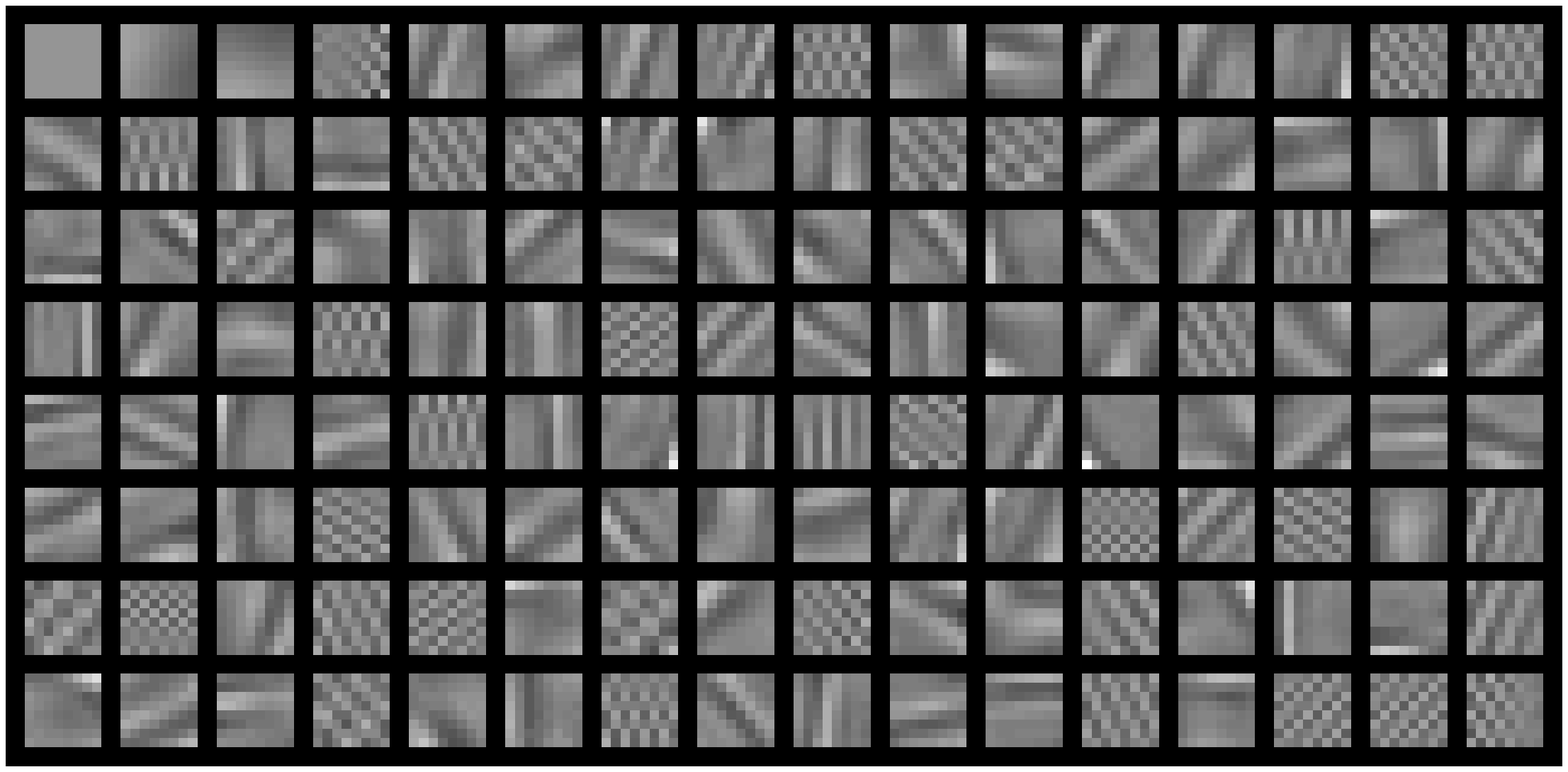} \\
(a) Barbara, $L=1$ & (b) Barbara, $L=3$  \\
\includegraphics[width=8cm]{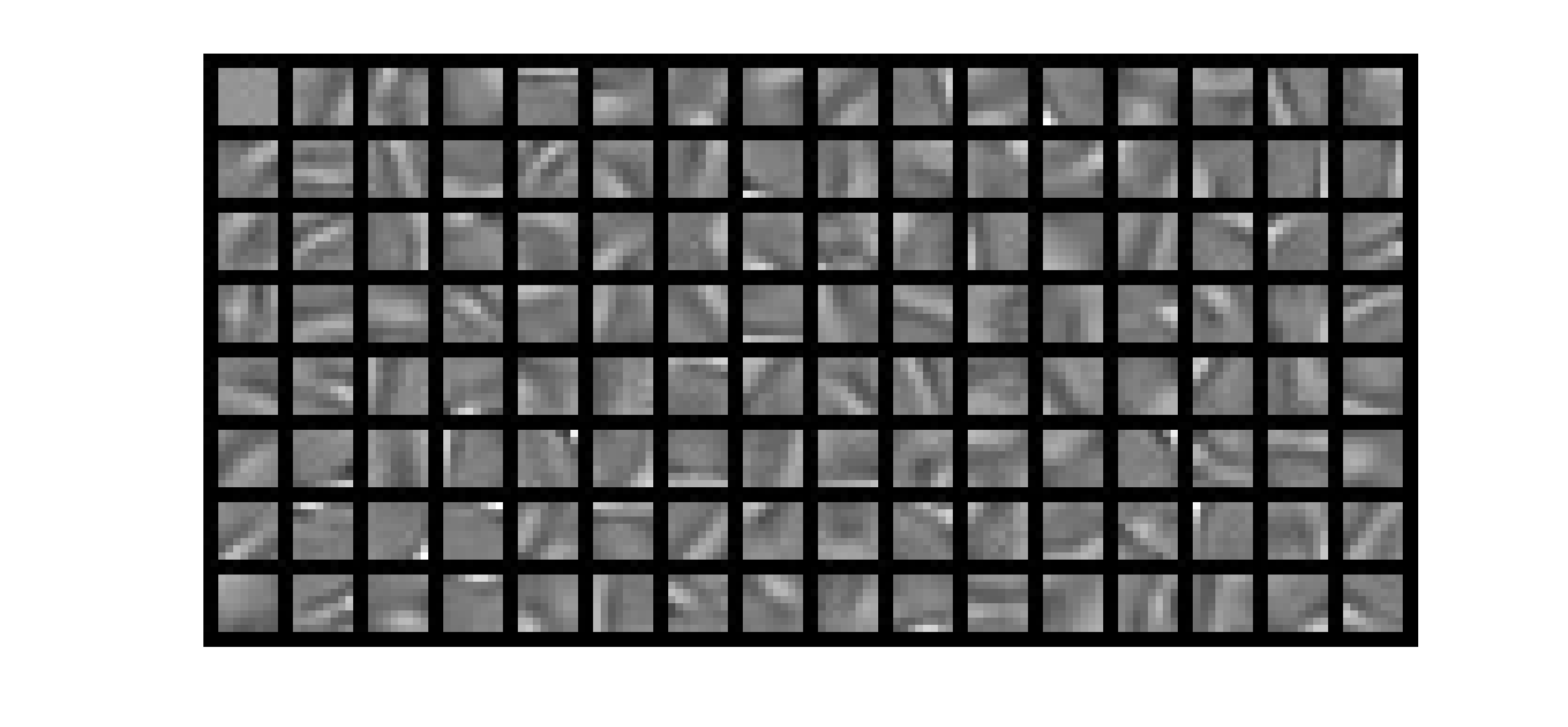} &\includegraphics[width=8cm]{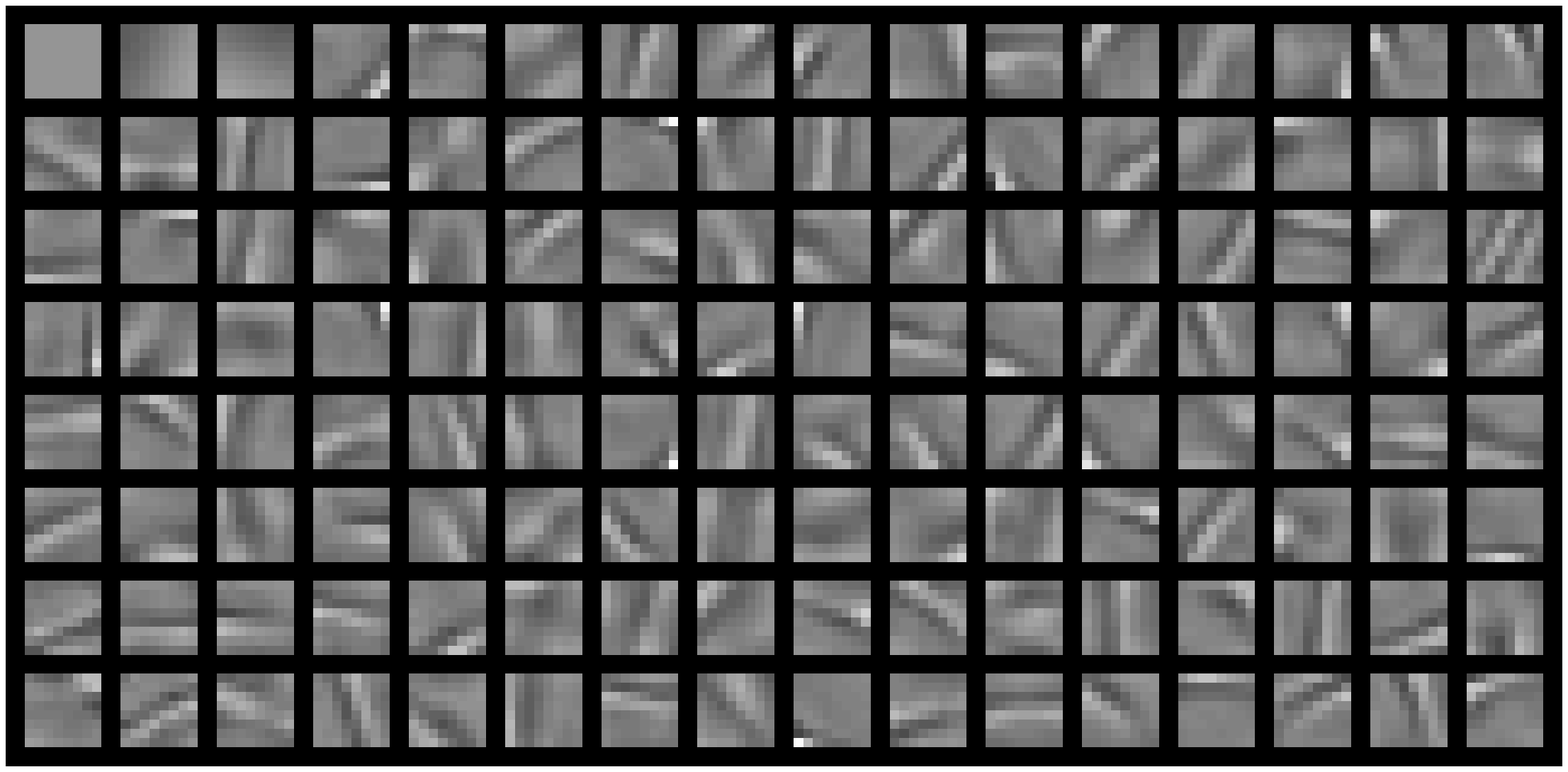} \\
(c) Peppers, $L=1$ & (d) Peppers, $L=3$
\end{tabular}
\caption{Dictionaries and low-rank atoms (left upper corner) learned with ITKrMM on all $8 \times 8$ patches of Peppers resp. Barbara corrupted with 30\% erasure rate.\label{fig:dicos_lr}}
\end{figure}
To understand why the ITKrMM dictionaries perform better for the small erasure probabilities and textured images, we have a look at the dictionaries learned with ITKrMM (L=1) and wKSVD on patches with 30\% and 70\% corruption for Barbara (Figure~\ref{fig:dicos_barb}) and Peppers (Figure~\ref{fig:dicos_pepp}). For 30\% random erasures we can see (especially on Barbara) that the dictionaries are very similar but that ITKrMM produces more textured (high frequency) atoms, while wKSVD prefers smooth (low frequency) atoms. Therefore the ITKrMM dictionary can inpaint finer details and outperforms wKSVD. Looking at the dictionaries learned on patches from 70\% corrupted imagegs, we now see different trends for the two dictionary types. For ITKrMM the 70\% erasure dictionaries are simply noisy versions of their 30\% counterparts. For wKSVD on the other hand the 70\% erasure dictionaries contain only noisy versions of the smooth atoms from their 30\% counterparts but not many recognisable noisy versions of the high frequency atoms. In the case of highly corrupted textured images this puts wKSVD at a disadvantage, while in the case of highly corrupted smooth images it prevents inpainting with noisy details. This difference in number of textured vs smooth atoms also explains the different performance of the two dictionary types for the structured mask on the textured resp. smooth images. Further, it explains why for low random erasures levels and textured images ITKrMM with $L=3$ improves over ITKrMM with $L=1$. As we can see in Figure~\ref{fig:dicos_lr}, the low-rank component consists of smooth (low frequency atoms). Because of the orthogonality constraint a larger low-rank component therefore increases the presence of textured high frequency atoms in the dictionary.\\
Finally, we want to mention that the main advantage of ITKrMM over wKSVD is the significant improvement in running time. For the random erasure setting, patch size $8\times 8$, ITKrMM with $L=1$ is about 8 times faster than wKSVD, while for the structured erasure setting, patch size $12\times12$, ITKrMM is about 12 times faster than wKSVD\footnote{as observed by running both algorithms in unoptimised form on the UIBK LEO3e computing cluster consisting of 45 nodes with 20 Intel Xeon (Haswell) computing cores each, all nodes equipped with 64GB RAM except for two nodes with 512GB RAM}. In general the speed up of ITKrMM over wKSVD becomes more pronounced the larger $S$ or $d$ resp. $K$ are.\\
After verifying that ITKrMM indeed learns meaningful and applicable dictionaries also on real data we now turn to a discussion of our results.

\section{Conclusion and Future directions}\label{sec:discussion}
%%%%%%%%%%%
Motivated by a real-life problem with corrupted data in diabetes therapy management, we here extended the iterative thresholding and K residual means (ITKrM) algorithm for dictionary learning to learning dictionaries from incomplete/masked data (ITKrMM). To account for the presence of a low-rank component in the data we further introduced a modified version of the ITKrMM algorithm to recover also the low-rank component and adapted the ITKrMM algorithm to the potential presence of such a low-rank component. 
In extensive tests on synthetic data, we demonstrated that incorporating information about the corruption (missing coordinates) dramatically improves the dictionary learning performance and that ITKrMM is able to recover dictionaries from data with up to 80\% corruption. We further showed that the algorithm also learns meaningful dictionaries on corrupted image data and provides noticeable improvements in terms of required computational resources compared to wKSVD, a state-of-the-art algorithm for dictionary learning/refinement in the presence of erasures. Moreover, when used for inpainting, the ITKrMM dictionaries perform on a par with their wKSVD counterparts and in case of textured images ITKrMM leads to notable improvements, since the high frequency atoms in the learned dictionaries are better preserved. All of the experiments reported in this paper can be reproduced with the ITKrMM Matlab toolbox, which is freely available at \url{http://homepage.uibk.ac.at/~c7021041/code/ITKrMM.zip}.
One slight disappointment is that in synthetic experiments with a random initialisation ITKrMM does not recover the full dictionary. Instead, it recovers some atoms twice and some atoms are 1:1 linear combinations of two other ground truth atoms. This phenomenon has already been observed in the case of ITKrM and there are ongoing efforts to counter it with replacement strategies, which also open up the road to adaptively choosing the sparsity level, the dictionary size and the size of the low-rank component, \cite{sc17}. Once these strategies for ITKrM are finalised they will have straighforward extensions to the case of corrupted data, that is ITKrMM. \\
Most of our active research efforts are currently devoted to extending the local convergence results for ITKrM to the case of ITKrMM with random erasures, which will hopefully be available in the near future, \cite{nascXX}. 
Another interesting future direction consists of designing faster and more efficient algorithms, for instance, by adaptive patch selection for dictionary learning from incomplete data as it was suggested in seismic data recovery \cite{yumaos16}. More generally, we are interested in extending the concept of learning dictionaries from data with erasures to more general types of corruption such as for instance blurring, where the resulting dictionaries can then be used for deblurring.

\section*{Acknowledgements}
V. Naumova acknowledges the support of project 'Function-driven Data Learning in High Dimension' (FunDaHD) funded by the Research Council of Norway
and K. Schnass is in part supported by the Austrian Science Fund (FWF) under Grant no. Y760.
In addition the computational results presented have been achieved (in part) using the HPC infrastructure LEO of the University of Innsbruck.

% and improved thanks to the reviewers' detailed and pointed comments.
%\newpage

\bibliography{/Users/karin/Desktop/latexnotes/karinbibtex}

\begin{thebibliography}{10}

\bibitem{aganjaneta13}
A.~Agarwal, A.~Anandkumar, P.~Jain, P.~Netrapalli, and R.~Tandon.
\newblock Learning sparsely used overcomplete dictionaries via alternating
  minimization.
\newblock In {\em COLT 2014 (arXiv:1310.7991)}, 2014.

\bibitem{ahelbr06}
M.~Aharon, M.~Elad, and A.M. Bruckstein.
\newblock {K}-{S}{V}{D}: An algorithm for designing overcomplete dictionaries
  for sparse representation.
\newblock {\em IEEE Transactions on Signal Processing.}, 54(11):4311--4322,
  November 2006.

\bibitem{argemamo15}
S.~Arora, R.~Ge, T.~Ma, and A.~Moitra.
\newblock Simple, efficient, and neural algorithms for sparse coding.
\newblock In {\em COLT 2015 (arXiv:1503.00778)}, 2015.

\bibitem{argemo13}
S.~Arora, R.~Ge, and A.~Moitra.
\newblock New algorithms for learning incoherent and overcomplete dictionaries.
\newblock In {\em COLT 2014 (arXiv:1308.6273)}, 2014.

\bibitem{bakest14}
B.~Barak, J.A. Kelner, and D.~Steurer.
\newblock Dictionary learning and tensor decomposition via the sum-of-squares
  method.
\newblock In {\em STOC 2015 (arXiv:1407.1543)}, 2015.

\bibitem{bestfa13}
S.~Beckouche, J.L. Starck, and M.~Fadili.
\newblock Astronomical image denoising using dictionary learning.
\newblock {\em Astronomy and Astrophysics}, 556(A132), 2013.

\bibitem{calimawr11}
E.~Cand\`es, X.~Li, Y.~Ma, and J.~Wright.
\newblock Robust principle component analysis?
\newblock {\em Journal of the {ACM}}, 58(3), 2011.

\bibitem{carota06}
{E}. {C}and{\`e}s, {J}. {R}omberg, and {T}. {T}ao.
\newblock {R}obust uncertainty principles: exact signal reconstruction from
  highly incomplete frequency information.
\newblock {\em {IEEE} {T}ransactions on {I}nformation {T}heory},
  52(2):489--509, 2006.

\bibitem{damazh94}
G.~M. Davis, S.~Mallat, and Z.~Zhang.
\newblock Adaptive time-frequency decompositions with matching pursuits.
\newblock {\em SPIE Optical Engineering}, 33(7):2183--2191, July 1994.

\bibitem{do06cs}
D.L. Donoho.
\newblock {C}ompressed sensing.
\newblock {\em {IEEE} {T}ransactions on {I}nformation {T}heory},
  52(4):1289--1306, 2006.

\bibitem{doelte06}
D.L. Donoho, M.~Elad, and V.N. Temlyakov.
\newblock Stable recovery of sparse overcomplete representations in the
  presence of noise.
\newblock {\em {IEEE} {T}ransactions on {I}nformation {T}heory}, 52(1):6--18,
  January 2006.

\bibitem{elstqudo05}
M.~Elad, J.L. Starck, P.~Querre, and D.L. Donoho.
\newblock Simultaneous cartoon and texture image inpainting using morphological
  component analysis (mca).
\newblock {\em Appl. Comput. Harmon. Anal.}, 19(3):340--358, 2005.

\bibitem{enaahu99}
K.~Engan, S.O. Aase, and J.H. Husoy.
\newblock Method of optimal directions for frame design.
\newblock In {\em {ICASSP}99}, volume~5, pages 2443--2446, 1999.

\bibitem{olsfield96}
D.J. Field and B.A. Olshausen.
\newblock Emergence of simple-cell receptive field properties by learning a
  sparse code for natural images.
\newblock {\em Nature}, 381:607--609, 1996.

\bibitem{bagrje14}
R.~Gribonval, R.~Jenatton, and F.~Bach.
\newblock Sparse and spurious: dictionary learning with noise and outliers.
\newblock {\em {I}{E}{E}{E} {T}ransactions on {I}nformation {T}heory},
  61(11):6298--6319, 2015.

\bibitem{grsc10}
R.~Gribonval and K.~Schnass.
\newblock Dictionary identifiability - sparse matrix-factorisation via
  $l_1$-minimisation.
\newblock {\em {IEEE} {T}ransactions on {I}nformation {T}heory},
  56(7):3523--3539, July 2010.

\bibitem{kreutz03}
K.~Kreutz-Delgado, J.F. Murray, B.D. Rao, K.~Engan, T.~Lee, and T.J. Sejnowski.
\newblock Dictionary learning algorithms for sparse representation.
\newblock {\em Neural Computations}, 15(2):349--396, 2003.

\bibitem{lese00}
M.~S. Lewicki and T.~J. Sejnowski.
\newblock Learning overcomplete representations.
\newblock {\em Neural Computations}, 12(2):337--365, 2000.

\bibitem{mabapo12}
J.~Mairal, F.~Bach, and J.~Ponce.
\newblock Task-driven dictionary learning.
\newblock {\em IEEE Transactions on Pattern Analysis and Machine Intelligence},
  34(4):791--804, 2012.

\bibitem{mabaposa10}
J.~Mairal, F.~Bach, J.~Ponce, and G.~Sapiro.
\newblock Online learning for matrix factorization and sparse coding.
\newblock {\em Journal of Machine Learning Research}, 11:19--60, 2010.

\bibitem{maelsa08}
J.~Mairal, M.~Elad, and G.~Sapiro.
\newblock Sparse representation for color image restoration.
\newblock {\em {IEEE} {T}ransactions on Image {P}rocessing}, 17(1):53--69,
  2008.

\bibitem{masael08}
J.~Mairal, G.~Sapiro, and M.~Elad.
\newblock Learning multiscale sparse representation for image and video
  restoration.
\newblock {\em Multiscale Model. Simul.}, 7(1):214--241, 2008.

\bibitem{nape12}
V.~Naumova and S.~Pereverzyev.
\newblock Blood glucose predictors: an overview on how recent developments help
  to unlock the problem of glucose regulation.
\newblock {\em Recent Patents on Computer Science}, 5:1--11, 2012.

\bibitem{nascXX}
V.~Naumova and K.~Schnass.
\newblock Dictionary learning from incomplete data, {P}art~{II} theory.
\newblock the near future.

\bibitem{parekr93}
Y.~Pati, R.~Rezaiifar, and P.~Krishnaprasad.
\newblock {O}rthogonal {M}atching {P}ursuit : recursive function approximation
  with application to wavelet decomposition.
\newblock In {\em Asilomar Conf. on Signals Systems and Comput.}, 1993.

\bibitem{rubrel10}
R.~Rubinstein, A.~Bruckstein, and M.~Elad.
\newblock Dictionaries for sparse representation modeling.
\newblock {\em Proceedings of the IEEE}, 98(6):1045--1057, 2010.

\bibitem{sc14}
K.~Schnass.
\newblock On the identifiability of overcomplete dictionaries via the
  minimisation principle underlying {K-SVD}.
\newblock {\em Applied Computational Harmonic Analysis}, 37(3):464--491, 2014.

\bibitem{sc15}
K.~Schnass.
\newblock Convergence radius and sample complexity of {ITKM} algorithms for
  dictionary learning.
\newblock {\em arXiv:1503.07027}, 2015.

\bibitem{sc14b}
K.~Schnass.
\newblock Local identification of overcomplete dictionaries.
\newblock {\em Journal of Machine Learning Research (arXiv:1401.6354)},
  16(Jun):1211--1242, 2015.

\bibitem{sc15imn}
K.~Schnass.
\newblock A personal introduction to theoretical dictionary learning.
\newblock {\em Internationale Mathematische Nachrichten}, 228:5--15, 2015.

\bibitem{sc17}
K.~Schnass.
\newblock Sequential dictionary learning with parameter selection.
\newblock {\em in preparation}, 2017.

\bibitem{schpa16}
M.~Schoemaker and C.~Parkin.
\newblock {CGM} - {H}ow good is good enough?
\newblock {\em in Prediction methods for blood glucose concentration. Design,
  use and evaluation}, 2015.

\bibitem{sken10}
K.~Skretting and K.~Engan.
\newblock Recursive least squares dictionary learning algorithm.
\newblock {\em {IEEE} {T}ransactions on {S}ignal {P}rocessing},
  58(4):2121--2130, April 2010.

\bibitem{spwawr12}
D.~Spielman, H.~Wang, and J.~Wright.
\newblock Exact recovery of sparsely-used dictionaries.
\newblock In {\em COLT 2012 (arXiv:1206.5882)}, 2012.

\bibitem{suquwr15}
J.~Sun, Q.~Qu, and J.~Wright.
\newblock Complete dictionary recovery over the sphere.
\newblock In {\em ICML 2015 (arXiv:1504.06785)}, 2015.

\bibitem{wrma09}
J.~Wright, A.~Yang, A.~Ganesh, S.~Sastry, and Y.~Ma.
\newblock Robust face recognition via sparse representation.
\newblock {\em IEEE Transactions on Pattern Analysis and Machine Intelligence},
  31(2), 2009.

\bibitem{yawrhuma10}
J.~Yang, J.~Wright, T.~Huang, and Y.~Ma.
\newblock Image super-resolution via sparse representation.
\newblock {\em {IEEE} {T}ransactions on Image {P}rocessing}, 19(11):2861--2873,
  2010.

\bibitem{yumaos16}
S.~Yu, J.~Ma, and S.~Osher.
\newblock Monte carlo data-driven tight frame for seismic data recovery.
\newblock {\em Geophysics}, 81(4):V327--V340, 2015.

\end{thebibliography}
\bibliographystyle{plain}
\end{document}